\documentclass[manuscript,screen,nonacm]{acmart}

\AtBeginDocument{%
  \providecommand\BibTeX{{%
    \normalfont B\kern-0.5em{\scshape i\kern-0.25em b}\kern-0.8em\TeX}}}

\usepackage{xcolor}
\usepackage{colortbl}
\usepackage{siunitx}
\usepackage{multirow}
\usepackage{makecell}

\definecolor{oddrowcolor}{gray}{0.95}
\definecolor{evenrowcolor}{gray}{0.85}

\begin{document}

\title{Physics-Informed Machine Learning On Polar Ice: A Survey}

\author{Zesheng Liu}
\affiliation{%
  \institution{Department of Computer Science and Engineering, Lehigh University}
  \city{Bethlehem}
  \state{Pennsylvania}
  \country{USA}
}
\email{zel220@lehigh.edu}

\author{YoungHyun Koo}
\affiliation{
  \institution{Department of Computer Science and Engineering, Lehigh University}
  \city{Bethlehem}
  \state{Pennsylvania}
  \country{USA}}
\email{yok223@lehigh.edu}

\author{Maryam Rahnemoonfar}
\affiliation{
  \institution{Department of Computer Science and Engineering, Department of Civil and Environmental Engineering, Lehigh University}
  \city{Bethlehem}
  \state{Pennsylvania}
  \country{USA}}
\email{maryam@lehigh.edu}
\authornote{Corresponding author}

\begin{abstract}

The mass loss of the polar ice sheets contributes considerably to ongoing sea-level rise and changing ocean circulation, leading to coastal flooding and risking the homes and livelihoods of tens of millions of people globally. To address the complex problem of ice behavior, physical models and data-driven models have been proposed in the literature. Although traditional physical models can guarantee physically meaningful results, they have limitations in producing high-resolution results. On the other hand, data-driven approaches require large amounts of high-quality and labeled data, which is rarely available in the polar regions. Hence, as a promising framework that leverages the advantages of physical models and data-driven methods, physics-informed machine learning (PIML) has been widely studied in recent years. In this paper, we review the existing algorithms of PIML, provide our own taxonomy based on the methods of combining physics and data-driven approaches, and analyze the advantages of PIML in the aspects of accuracy and efficiency. Further, our survey discusses some current challenges and highlights future opportunities, including PIML on sea ice studies, PIML with different combination methods and backbone networks, and neural operator methods.
 
\end{abstract}


\maketitle

\section{Introduction}

As the global climate has been warming due to anthropogenic CO2 emissions, Greenland and the Antarctic ice sheets have undergone significant mass loss during the last few centuries. Ice losses from Greenland and Antarctic ice sheets have reached more than 7500 Gt since 1992, contributing to 21 mm of global sea level rise \cite{Otosaka2023}. Although these trends exhibit large inter-annual variability for various regions and climate conditions, accelerated mass losses have been found in most studies \cite{Forsberg2017, Mouginot2019, zwally2011, Rignot2011} for both Greenland and the Antarctic. In general, ice dynamics is the primary driver of mass loss in Greenland \cite{Choi2021, Mouginot2019}, whereas ice melting on floating ice shelves contributes more to mass loss in the Antarctic \cite{Bell2020}. If the current CO2 emissions and ice loss trends continue, the global mean sea level will rise by 0.7-1.3 m by 2100 \cite{icerise}. Meanwhile, Arctic sea ice extent and thickness have decreased over the last few decades \cite{Notz2016, Kwok2018}, showing a loss of more than \SI{8,000}{\km\squared} of volume \cite{Schweiger2011}. As a result, the Arctic Ocean will likely be ice-free by the 2030s \cite{Francis2023}. Although the Antarctic sea ice extent showed an increasing trend from the 1970s to 2016, it has also recently experienced a dramatic reduction after 2016 \cite{eayrs2021}. Considering such changes in polar ice have significant impacts on global climate, it is essential to accurately predict polar ice behavior.

Based on the fundamental physical laws stacked for the last few decades, many numerical models have been developed to explain the physical behavior of ice sheets and sea ice. These physical models solve the continuous equations on numerical grids by considering complex interactions between the atmosphere, ocean, snow/ice surface, and bed topography. However, solving these physical models is computationally expensive due to the complexity of these variables. Hence, physical models have limitations in producing high-resolution results as the complexity and computational cost increase exponentially with higher resolution. Additionally, the high sensitivity of these models to the initial and boundary conditions and physical assumptions can lead to substantial disagreement with actual observations.

As computing techniques are dramatically enhanced, machine learning techniques have reached great success in a few domains, like computer vision or natural language processing. Networks are trained based on tons of input datasets and are shown to have the ability to represent complex patterns within the dataset. As a sequence, this raises the study of using machine learning techniques as alternatives for complex and computationally expensive physical models.

However, an important fact is that those complex machine learning algorithms are highly data-driven and usually considered black boxes. It is hard to build solid explanations for each step of the machine learning algorithm. Moreover, with the enhancement of experiment techniques, it has become easier to collect multi-fidelity observational data with different spatial and temporal resolutions. While machine learning can serve as an automatic tool to analyze and extract different patterns from these observational data, it has an implicit problem in generalization: a model can have solid performance on known data, but when it comes to unseen data, it may make predictions that are physically inconsistent or implausible \cite{Karniadakis2021}.

In this respect, physics-informed machine learning has been widely studied in recent years, which is a promising framework that leverages both the ability of data-driven machine learning techniques to automatically extract patterns or laws from vast amounts of observational data and the ability of physical models to account for the physical laws and give out the prediction that has reasonable physical meaning. The machine learning networks are trained with underlying physical laws or governing equations known as prior knowledge at the beginning of training. Physical laws are brought as a strong theoretical constraint to the machine learning algorithms\cite{Desai2021}, and physics-informed machine learning can maintain both data-driven and physical consistency. 

Currently, there is no commonly agreed-upon terminology for the hybrid method that combines physical law and data-driven machine learning techniques. ``physics-informed'', ``physics-based'', ``knowledge-informed'', and ``science-informed'' are the widely used terms. There are also various ways to categorize how physics is introduced to machine learning. Karnidakis et al.\cite{Karniadakis2021} proposed their taxonomy based on observational biases, inductive biases and learning biases in machine learning. Kim et al.\cite{Kim2021} categorized the methods in the field of dynamical systems based on three key components: the choice architecture for deep neural networks, the choice method for knowledge representation, and the choice method for knowledge.  

In the aspect of survey papers on physics-informed machine learning, there are a few survey papers that only focus on the history, methodologies, and taxonomies of physics-informed machine learning, including Karnidakis et al.\cite{Karniadakis2021}, Pateras et al.\cite{taxonomic_survey}, and Seyyedi et al.\cite{survey_integrated_models}. Several surveys also discussed the application of physics-informed machine learning in different subdomains, such as climate\cite{survey_climate}, fluid and solid mechanics\cite{survey_subdomain_fluid}, chemical engineering\cite{survey_subdomain_chemical_engineering}, reliability and systems safety applications\cite{survey_subdomain_system}, and data anomaly\cite{survey_subdomain_data_anomaly}. To the best of our knowledge, there is no previous survey on physics-informed machine learning for polar ice. Therefore, this becomes the major motivation and contribution of our work. 

\begin{figure}[h]
  \centering
  \includegraphics[width=\linewidth]{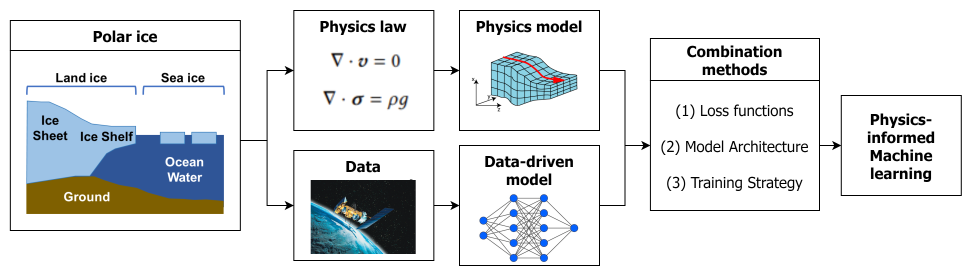}
  \caption{Roadmap of physics-driven, data-driven, and physics-informed machine learning studies for polar ice}
  \label{PIML_Intro}
  \Description[overview of polar ice study]{Polar ice can be divided into two categories depending on the targeting area: land ice and sea ice. For the modeling of these two types of ice, physics-driven and data-driven approaches have been employed. The physics-driven method uses a series of governing equations relevant to physics, but the data-driven method only relies on the observation data. Physics-informed machine learning can be achieved by integrating these two approaches; three representative integration methods are (1) adding physics-informed loss functions, (2) adjusting model architecture, and (3) using physics-informed training strategy.}
\end{figure}

In this paper, we do a comprehensive survey on the study of polar ice. We review different methods, including physical models, data-driven machine learning techniques, and physics-informed machine learning. We discuss the limitations of pure physical or data-driven methods and how physics-informed machine learning can handle problems in these methods. Inspired by Karnidakis et al.\cite{Karniadakis2021} and Kim et al.\cite{Kim2021}, We provide our own taxonomy of all the relevant physics-informed machine learning papers based on their methods to incorporate related physics. There are three major methods of combining physics and data-driven approaches: through loss function, through model architecture, and through training strategy. Fig.\ref{PIML_Intro} shows the general roadmap of our survey's topic.

Our major contributions are: (1) To the best of our knowledge, this is the first survey summarizing relevant work on physics-informed machine learning for polar ice. Although Kashinath et al. \cite{survey_climate} investigated the application of physics-informed machine learning in climate change, it is necessary to focus specifically on polar ice, considering the importance of polar ice on global climate and sea level rise. Organizing the recent development of physics-informed machine learning studies in polar ice will benefit new researchers who want to focus on relevant topics. (2) We proposed our own taxonomy to categorize those relevant papers based on their methods of combining physics with data-driven machine learning approaches. (3) Besides the physics-informed machine learning papers for polar ice, we include those physical models and pure data-driven models for polar ice to compare with. The comprehensive comparison between physics-informed machine learning and traditional pure physical or data-driven methods can clearly show the advantages and limitations of the current studies on physics-informed machine learning for polar ice and provide valuable insights on relevant research topics. (4) Based on the comprehensive comparison, we point out and discuss how to address the current challenges of physics-informed machine learning to enhance its potential for polar ice application.

This paper is organized as follows: Section \ref{ice_components} introduces the basic definition and category of polar ice. Section \ref{physics} gives a comprehensive overview of different physical laws for polar ice and their application. Section \ref{datadriven} reviews a few data-driven machine learning models used for polar ice and categorizes them based on the network architecture. Section \ref{piml_method} analyzes the fundamental concepts of physics-informed machine learning. Section \ref{case} is a case study for applying physics-informed machine learning in the domain of polar ice. Section \ref{advantages} and Section \ref{Comparision} compare the differences between physics-informed machine learning with the physical model or data-driven model, emphasizing the advantages, limitations of current implementations, and future opportunities of physics-informed machine learning on polar ice. Section \ref{conclusion} concludes the paper with a summary of the whole paper and suggestions for future outlook.

\section{Different components of ice in polar region}\label{ice_components}

Polar ice can be categorized into two kinds: Land ice and sea ice. These two kinds of polar ice play different roles in global climate and have different physical mechanisms. 

Land ice forms as the accumulation of snow or frozen fresh water that is grounded to land. The accumulation process is very slow and can usually be measured and evaluated on an annual base; therefore, the seasonality can be negligible. There are a few possible research topics for land ice, including bed properties\cite{Land_Ice_Bed_Properties}, ice thickness and internal layers\cite{Land_Ice_Internal_Layers}, land ice dynamics\cite{Land_Ice_Dynamics}, and synergistic effect of ice sheet surface and atmosphere\cite{Land_Ice_Interaction}.

On the contrary, sea ice is formed when ocean water is frozen and floating on the sea. Due to hydrostatic equilibrium, sea ice does not directly contribute to the sea level rise as it is already in the water. The formation process of sea ice is very fast, and most of the sea ice has a strong seasonal trend. In the summer, the ocean absorbs more heat, which results in a temperature increase, and then the sea ice will tend to melt (a.k.a. ice-ocean albedo feedback \cite{Judith1995_albedo_feedback}). In the winter, sea ice will reflect most of the sunlight from the ocean, and the seawater will be able to freeze and form sea ice. Possible research topics for sea ice include the mechanism and evolution of sea ice formation and melt \cite{Sea_Ice_Evolution}, sea ice dynamics \cite{Sea_Ice_Dynamics}, sea ice salinity \cite{Sea_Ice_Salinity}, and the interaction between sea ice and other earth climate systems including sunlight and atmosphere \cite{Sea_Ice_Topics}.

There are a few traditional methods to study polar ice, which include physical models and data-driven models. Table \ref{traditional} gives a summary of these methods and related research topics. In the following sections, we will briefly review these traditional methods, highlighting their achievements and limitations. All the limitations of traditional methods lead to the exploration of better methods, where physics-informed machine learning plays an essential role.

\begin{table}
\caption{An overview of the traditional methods and related research topics}
\label{traditional}
\begin{tabular}{m{1.6cm}m{3.0cm}m{4.0cm}m{4.0cm}}
\toprule
\multicolumn{2}{c}{Methods}                                 & Land ice                                       & Sea ice                                                                    \\ \midrule
\multicolumn{2}{c}{\multirow{3}{*}{Physical Models}}        & Ice flow  \cite{Glen1955,Greve_Blatter_Glen, Hindmarsh2004, Hindmarsh2004, Blatter1995,Pattyn1996, SSA1, Hutter1983, Larour2012}      & Ice concentration \cite{Holland2012,Hibler1979}
                           \\
\multicolumn{2}{c}{}                                        & Ice thickness \cite{massconservation, Morlighem2014,Morlighem2013,Morlinghem2017_bedmachine, morlighem2020_bedmachine_Ant} & Ice thickness \cite{Holland2012,Hibler1979}                                                     \\
\multicolumn{2}{c}{}                                        &            &  Ice drift \cite{Hibler1979,FATHI2020105729}          \\ \midrule
\multirow{12}{*}{\makecell{Data-driven\\Models}} & \multirow{3}{*}{\makecell{Multi-layer\\Perceptron}} & Ice mass balance \cite{Bolibar2020,Bolibar2020_reconstruction}                        & Ice concentration \cite{Kim2019_ANN}                                                                       \\
                                     &                      & Ice melting  \cite{Hu2021_surface_melting,Sellevold2021}                         &                                                                            \\
                                     &                      & Ice thickness  \cite{Clarke_2009}                       &                                                                            \\ \cmidrule{2-4} 
                                     & \multirow{3}{*}{\makecell{Convolutional\\Neural  Network}} & Ice front and movements \cite{ronneberger2015unet,Mohajerani2019,Baumhoer2019,Zhang2019,Loebel2022,Zhang2021}                & Ice concentration \cite{Andersson2021,Grigoryev2022,Kim2020_SIC_CNN,Fritzner2020,Ren2022_SIC_CNN,koo2023multitask}                                        \\
                                     &                      & Ice layer and thickness \cite{varshney2020_deep_ice_layer, Varshney2021,yari_2020_multi_scale,ibikunle_2020,rahnemoonfar_yari_paden_koenig_ibikunle_2021, yari_airborne_snow_radar, ibikunle_snow_radar_echogram_2023,varshney2023skipwavenet,icelayers2021icme, varshney_2021_regression_networks}          & Ice thickness \cite{Liang2023}                                                                          \\
                                     &                      & Ice flow emulator \cite{jouvet_cordonnier_kim_lüthi_vieli_aschwanden_2022}                              &                                                                            \\ \cmidrule{2-4} 
                                     & \multirow{2}{*}{\makecell{Recurrent\\Neural  Network}} & Ice flow \cite{Zhang2023} & Ice concentration \cite{Chi2017_LSTM,Zheng2022, Liu_seaice_ConvLSTM, Feng2023_ConvLSTM, Liu2021_LSTM_SIC, Choi2019_GRU}                                           \\
                                     &                      & Sea level rise  \cite{Katwyk2023}     &  Ice drift \cite{Petrou_2017_RNN_seaicemotion}                                                                        \\ \cmidrule{2-4} 
                                     & \multirow{3}{*}{\makecell{Graph\\Neural Network}} & Sea level rise \cite{Lin2023_GNN}     & Causality discovery \cite{Huang2021_GNN} \\
                                     &                      & Snow accumulation \cite{zalatan_icip,Zalatan2023,Zalatan_igarss}  &                                                                            \\ 
                                     &                      & Ice flow \cite{rahnemoonfar2024graph}
                                     &                                                                            \\ \cmidrule{2-4} 
                                     & \multirow{2}{*}{\makecell{Generative\\Adversarial Networks}} & Ice layer and thickness \cite{Rahnemoonfar2019_GAN,Rahnemoonfar2020_GAN}                & Ice concentration \cite{Kim2023_GAN}                           \\
                                     &                      & Bed topography \cite{Leong2020_DeepBedMap, multibranch}               &                                                                            \\ \bottomrule
\end{tabular}
\end{table}

\section{Physical model for polar ice}\label{physics}


In this section, we will cover various underlying physical laws in both land ice and sea ice modeling. Relevant papers are categorized by their study objective and underlying physics. Table \ref{term_physics} summarizes those common terminologies used in the following sections.
    
\begin{table}  
\caption{Termologies commonly used for underlying physical laws}
\label{term_physics}
\centering
\scalebox{0.8}{
\begin{tabular}{cc}
\toprule
Symbol & Defination\\
\midrule
$\boldsymbol{v}$ & 3-D velocity of ice sheets\\
$(u, v, w)$ & $x$, $y$, $z$ components of the velocity vector $\boldsymbol{v}$\\
$\boldsymbol{u}$ & 2-D velocity of ice\\
$\sigma$ & Stress tensor\\
$\rho$ & Ice Density\\
$\mu$ & Effective ice viscosity\\
$p$ & Pressure\\
$g$ & Gravity acceleration\\
$s$ & Surface elevation\\
$H$ & Ice thickness\\

\bottomrule
\end{tabular}}
\end{table}

\subsection{Land ice}

In physical models for ice sheets (land ice), the ice bodies have been commonly regarded as incompressible, viscous, and non-Newton fluid that follows the Stokes equation \citep{Glen1955}. Therefore, most physical approaches produce ice thickness and ice dynamics based on the mass conservation and momentum conservation equations. The fundamental mass conservation and momentum conservation equations of incompressible ice can be expressed by Glen's flow law \cite{Glen1955, Greve_Blatter_Glen}:

\begin{equation}\label{mass_conservation}
    \mathbf{\nabla} \cdot \boldsymbol{v} = 0
\end{equation}
\begin{equation}\label{momentum_conservation}
    \mathbf{\nabla} \cdot \boldsymbol{\sigma} = \rho g
\end{equation}
where $\boldsymbol{v}$ is three-dimensional velocity, $\sigma$ is stress tensor, $\rho$ is ice density, and $g$ is acceleration due to gravity. 

\subsubsection{Full-Stokes Equation}

Equation \ref{momentum_conservation} can be rewritten in terms of velocity and pressure:

\begin{equation}\label{FS1}
    \frac{\partial}{\partial x}\left(2\mu\frac{\partial u}{\partial x}\right) + \frac{\partial}{\partial y}\left(\mu\frac{\partial u}{\partial y}+\mu\frac{\partial v}{\partial x}\right) + \frac{\partial}{\partial z}\left(\mu\frac{\partial u}{\partial z}+\mu\frac{\partial w}{\partial x}\right) - \frac{\partial p}{\partial x} = 0
\end{equation}
\begin{equation}\label{FS2}
    \frac{\partial}{\partial x}\left(\mu\frac{\partial u}{\partial y} + \mu\frac{\partial v}{\partial x}\right) + \frac{\partial}{\partial y}\left(2\mu\frac{\partial v}{\partial y}\right) + \frac{\partial}{\partial z}\left(\mu\frac{\partial v}{\partial z}+\mu\frac{\partial w}{\partial y}\right) - \frac{\partial p}{\partial y} = 0
\end{equation}
\begin{equation}\label{FS3}
    \frac{\partial}{\partial x}\left(\mu\frac{\partial u}{\partial z} + \mu\frac{\partial w}{\partial x}\right) + \frac{\partial}{\partial y}\left(\mu\frac{\partial v}{\partial z}+\mu\frac{\partial w}{\partial y}\right) + \frac{\partial}{\partial z}\left(2\mu\frac{\partial w}{\partial z}\right) - \frac{\partial p}{\partial y}-\rho g = 0
\end{equation}
\begin{equation}\label{FS4}
    \frac{\partial u}{\partial x}+\frac{\partial v}{\partial y} + \frac{\partial w}{\partial z} = 0
\end{equation}

where $(u, v, w)$ are the $x$, $y$, and $z$ components of the velocity vector $\boldsymbol{v}$ in the Cartesian coordinate system ($z$ is the vertical direction), $p$ is the pressure, and $\mu$ is the effective ice viscosity. Equations \ref{FS1}, \ref{FS2}, \ref{FS3}, and \ref{FS4} represent the Full-Stokes (FS) system, and solving ice flow in this FS system means solving four unknown variables $(u, v, w, p)$ in these equations. Although this FS model provides accurate solutions, it is computationally expensive, which makes it challenging to apply this model to continental-scale ice flows with a high resolution. Therefore, several simplified formats of the FS model have been developed based on their own assumptions for specific ice flow conditions \cite{Hindmarsh2004}.

\subsubsection{Blatter-Pattyn approximation (BP)}

 First, the Blatter-Pattyn approximation (BP) \citep{Blatter1995, Pattyn1996} provides valid and efficient solutions in the majority of an ice sheet, both longitudinal stresses of fast-flowing ice streams and vertical shear stresses of slow ice \citep{Pattyn1996}. In this model, assuming that the vertical component of the momentum balance is approximated as hydrostatic, Equaion \ref{FS3} can be reduced to:
\begin{equation}
    \frac{\partial}{\partial z}\left(2\mu\frac{\partial w}{\partial z}\right) - \frac{\partial p}{\partial z} - \rho g = 0
\end{equation}
Additionally, this model assumes that horizontal gradients of the vertical velocity are negligible compared to the vertical gradient of the horizontal velocity (i.e. $\partial w/\partial x \ll \partial u / \partial z $ and $\partial w / \partial y \ll \partial v / \partial z$). Consequently, Equations \ref{FS1} and \ref{FS2} are reduced to:

\begin{equation}\label{BP1}
    \frac{\partial}{\partial x}\left(4\mu\frac{\partial u}{\partial x}+2\mu\frac{\partial v}{\partial y}\right) + \frac{\partial}{\partial y}\left(\mu\frac{\partial u}{\partial y}+\mu\frac{\partial v}{\partial x}\right)+\frac{\partial}{\partial z}\left(\mu\frac{\partial u}{\partial z}\right) = \rho g \frac{\partial s}{\partial x}
\end{equation}
\begin{equation}\label{BP2}
    \frac{\partial}{\partial x}\left(\mu\frac{\partial u}{\partial y}+\mu\frac{\partial v}{\partial x}\right) + \frac{\partial}{\partial y}\left(4\mu\frac{\partial v}{\partial y}+2\mu\frac{\partial u}{\partial x}\right)+\frac{\partial}{\partial z}\left(\mu\frac{\partial v}{\partial z}\right) = \rho g \frac{\partial s}{\partial y}
\end{equation}
where $s$ is surface elevation. These two equations lead the FS system with four unknown variables to the closed problem for the two horizontal velocity components (i.e., $u$ and $v$).

\subsubsection{Shallow Shelf Approximation (SSA)}

Another simplification of FS equations is Shallow Shelf Approximation (SSA) \cite{SSA1}. The SSA provides two-dimensional solutions of depth-averaged speed by assuming that horizontal velocity is depth-independent and vertical shear stresses are negligible. Therefore, under this approximation, Equations \ref{BP1} and \ref{BP2} are independent of $z$ and simplified further as follows:
\begin{equation}\label{SSA1}
    \frac{\partial}{\partial x}\left(4H\mu\frac{\partial u}{\partial x}+2H\mu\frac{\partial v}{\partial y}\right)+\frac{\partial}{\partial y}\left(H\mu\frac{\partial u}{\partial y}+H\mu\frac{\partial v}{\partial x}\right) = \rho g H \frac{\partial s}{\partial x}
\end{equation}
\begin{equation}\label{SSA2}
    \frac{\partial}{\partial y}\left(4H\mu\frac{\partial v}{\partial y}+2H\mu\frac{\partial u}{\partial x}\right)+\frac{\partial}{\partial x}\left(H\mu\frac{\partial u}{\partial y}+H\mu\frac{\partial v}{\partial x}\right) = \rho g H \frac{\partial s}{\partial y}
\end{equation}
where $H$ is the local ice thickness. The SSA can be an appropriate choice for modeling floating ice shelves and ice streams because sliding is mostly responsible for their movements. However, SSA cannot be successful near grounding lines, ice stream margins, or complex flows near an ice divide.

\subsubsection{Shallow Ice Approximation (SIA)}

The most simplified way to describe the ice sheet dynamics is the Shallow Ice Approximation (SIA), a zeroth-order approximation to the Stokes equations \citep{Hutter1983}. The SIA assumes that ice sheet dynamics is mostly driven by basal sheer stress by balancing the basal shear stress and gravitational driving stress of grounded ice. Under the SIA system, the Equation \ref{FS1}, \ref{FS2}, \ref{FS3}, and \ref{FS4} can be simplified into:

\begin{equation}\label{SIA1}
    \frac{\partial}{\partial z}\left(\mu\frac{\partial u}{\partial z}\right) = \rho g \frac{\partial s}{\partial x}
\end{equation}
\begin{equation}\label{SIA2}
    \frac{\partial}{\partial z}\left(\mu\frac{\partial v}{\partial z}\right) = \rho g \frac{\partial s}{\partial y}
\end{equation}
where $s$ denotes surface elevation. Although the SIA is computationally efficient and practical due to its simplicity, its too-simplified mechanical assumptions limit its applicability to non-ice streaming regions of ice sheets and valley glaciers where ice flow is dominated by vertical shearing. Hence, the SIA cannot properly represent short-term projections of ice dynamics in grounding lines, ice shelves, or ice streams \cite{Larour2012}.

\subsubsection{Temporal evolution of ice thickness}

In addition to the dynamics of ice sheets governed by Stokes equation, we can express the temporal evolution of ice thickness in ice sheets using the mass conservation law \cite{massconservation}:

\begin{equation}\label{mass_conservation_thickness}
    \frac{\partial H}{\partial t} + \nabla\cdot H \bar{\boldsymbol{u}}=\dot{M_s}-\dot{M_b}
\end{equation}
where $H$ is ice thickness and $\bar{\boldsymbol{u}}=(\bar{u},\bar{v})$ is the depth-averaged horizontal velocity, $\dot{M_s}$ is the surface mass balance (positive for accumulation and negative for ablation), and $\dot{M_b}$ is the basal melting rate (positive for melting and negative for freezing). \cite{massconservation} applied this equation to ice thickness measured by the airborne radar-sounding profiles and ice velocity derived from satellite synthetic‐aperture radar interferometry (InSAR) images. This mass conservation approach has been used to estimate the bed topography of many ice sheets \cite{Morlighem2014, Morlighem2013}, and further used to generate a high-resolution bed topography/bathymetric map of the entire Greenland and Antarctic ice sheets (BedMachine) \cite{Morlinghem2017_bedmachine, morlighem2020_bedmachine_Ant}.

\subsection{Sea ice}

Similar to land ice, mass balance and momentum balance equations underpin the physical models of sea ice by assuming it as elastic-viscous-plastic (EVP) material. However, the momentum regarding sea ice movement could be more complex than land ice because it is highly affected by the thermodynamic and dynamic behaviors of the atmosphere and ocean as floating on the ocean.

The evolution of sea ice concentration ($A$) and sea ice thickness ($H$) can be expressed by the following mass balance equations \cite{Holland2012, Hibler1979}:
\begin{equation}\label{SIC_massbalance}
\frac{\partial A}{\partial t} + \nabla\cdot(\boldsymbol{u}A)=f_A-r_A
\end{equation}
\begin{equation}\label{SIT_massbalance}
\frac{\partial H}{\partial t} + \nabla\cdot(\boldsymbol{u}H)=f_H-r_H
\end{equation}
where $\boldsymbol{u}$ is the sea ice motion vector, $f_A$ and $f_H$ are the changes in $A$ and $H$, respectively, from freezing or melting (thermodynamic mechanisms), and $r_A$ and $r_H$ are the changes in $A$ and $H$ from mass-conserving mechanical ice redistribution processes (e.g., ridging or rafting).

Additionally, numerous physical sea ice models have proposed the following mathematical equation to explain sea ice dynamics based on the momentum balance \cite{Hibler1979}:

\begin{equation}\label{eq:3}
m\frac{D \boldsymbol{u}}{D t} = -mf\boldsymbol{k}\times\boldsymbol{u} + \tau_{ai} + \tau_{wi} + \boldsymbol{F} -mg \nabla H_s
\end{equation}
where $D/Dt=\partial/\partial t + \boldsymbol{u} \cdot \nabla$ is the substantial time derivative, $m$ is the ice mass per unite area, $\boldsymbol{k}$ is a unit vector normal to the surface, $\boldsymbol{u}$ is the ice velocity, $f$ is the Coriolis parameter, $\tau_{ai}$ and $\tau_{wi}$ are the forces due to air and water stresses, $H_s$ is the elevation of the sea surface, $g$ is the gravity acceleration, and $\boldsymbol{F}$ is the force due to variations in internal ice stress. As many previous studies have already suggested, wind and ocean forcings have primary impacts on SID and dynamics. Fathi et al. use physics-informed deep neural networks on 4D-Flow MRI images to both generate high-resolution images and denoise\cite{FATHI2020105729}.

\subsection{Limitation of physical models}
Although scientists have proposed various physical models to explain the thermodynamics and dynamics of polar sea ice in the regime of land and sea, those physical models still have limitations and sources of uncertainties. First, considering the high complexity of ice behavior affected by the atmosphere, ocean, and topographic conditions, it is nearly impossible to represent every detail of this physical process. Although well-known and obvious physical laws, such as mass and momentum conservation, underlie physical models, we still do not fully understand every detailed process causing mass and momentum changes in ice bodies. The lack of detail in physical processes can be a source of uncertainties, particularly in boundary conditions (e.g., grounding lines of ice sheets, calving front of ice sheets, basal friction at the bedrock, sea ice marginal zone). Second, physical models highly depend on the parameterization of physical variables (e.g., ice density, ice viscosity). Since the true values of these properties are not available in most cases, physical models assume the most reasonable values for the parameterization. However, considering these values can change spatially and seasonally and bring significant uncertainties to the models, finding the optimal parameterization values is extremely challenging. Finally, solving physical models is computationally intensive. Finding numerical solutions for partial differential equations (PDEs) requires a lot of computational resources and time, even if the models are discretized (e.g., finite element method). This computational cost would be a more challenging issue with the higher complexity of PDEs, larger areas, and longer time spans.

\section{Data-driven Machine Learning for polar ice}\label{datadriven}

Various machine learning models with different network architectures have been used for modeling polar ice. In this section, we present an overview of machine learning techniques that have been applied by categorizing them into their network architectures.

\subsection{Multi-layer perceptron}
Multi-layer perception (MLP), a feedforward network consisting of interconnected nodes (a.k.a. neurons) of multiple layers, is the simplest neural network architecture that has been used in various studies. Based on the advantage of MLP in modeling complex nonlinear relationships between input and output variables, MLP has been used for the prediction of the mass balance of land ice. Bolibar et al. \cite{Bolibar2020} used this feed-forward MLP to model glacier-scale SMB from topographical variables (e.g., longitude, latitude, slope, altitude) and climatic variables (e.g., positive degree days, mean temperature, snowfall). They trained the MLP model with annual glacier-wide mass balance values for 30 glaciers in the French Alps, and the MLP outperformed other traditional statistical regression methods. Beyond these 30 glaciers with available mass balance data, this MLP was further used to reconstruct the mass balance of all 661 glaciers in the French Alps from 1967 to 2015 \cite{Bolibar2020_reconstruction}. Hu et al. \cite{Hu2021_surface_melting} also developed an MLP framework to estimate surface melting of ice sheets from the regional climate model data, meteorological observations from automatic weather stations (AWSs), and satellite-derived surface albedo. Sellevold et al. \cite{Sellevold2021} designed an MLP to predict the annual melt of the Greenland Ice Sheet from five atmospheric variables:  near-surface temperature, geopotential heights, cloud cover, incoming solar radiation, and snowfall. Clarke et al. \cite{Clarke_2009} adopted MLP to estimate glacier ice thickness from digital elevation models (DEM) and mask of present-day ice cover. Additionally, regarding sea ice, MLP has been used to predict sea ice concentration (SIC). Kim et al. \cite{Kim2019_ANN} used MLP to predict long-term SIC using satellite-derived SIC and climate variables.

\subsection{Convolutional Neural Network}
Convolutional neural networks (CNNs) have been widely used in computer vision and image processing tasks by learning 2D features via filter (kernel) optimization. Taking advantage of its strength in image classification and semantic segmentation, many studies used CNN to extract ice fronts and their movements automatically from imagery datasets. These studies often applied fully convolutional networks (FCN), particularly U-net \cite{ronneberger2015unet}, to implement image segmentation to the optical and radar images with manually labeled ice fronts. Mohajerani et al. \cite{Mohajerani2019} used a U-net architecture to detect the calving margins of several Greenland glaciers from Landsat single-band images. Baumhoer et al. \cite{Baumhoer2019} and Zhang et al. \cite{Zhang2019} also used U-net architectures and synthetic aperture radar (SAR) images to extract the fronts of glaciers. Furthermore, instead of relying on a single image sensor, there have been several attempts to combine multiple channels or multiple image sources from different sensors to take advantage of their strengths \cite{Loebel2022, Zhang2021}. 

CNN has been actively used to process radar echogram images to determine ice thickness or ice layer depth\cite{varshney2020_deep_ice_layer,Varshney2021, Dong2022,yari_2020_multi_scale,ibikunle_2020,rahnemoonfar_yari_paden_koenig_ibikunle_2021, yari_airborne_snow_radar, ibikunle_snow_radar_echogram_2023,varshney2023skipwavenet,icelayers2021icme, varshney_2021_regression_networks}. Varshney et al. \cite{varshney2020_deep_ice_layer, Varshney2021} used three popular FCNs (U-net \cite{ronneberger2015unet}, PSPNet \cite{zhao2017PSP}, DeepLabv3+ \cite{Chen2018}) to determine the depth of internal ice layers from snow radar images. Yari et al. \cite{yari_2019_smart_tracking,yari_2020_multi_scale}, Rahnemoonfar et al. \cite{rahnemoonfar_yari_paden_koenig_ibikunle_2021} and Varshney et al.\cite{varshney2023skipwavenet} adopted different kinds of multi-scale networks to extract ice layer from radar echogram images. Ibikunle et al.\cite{ibikunle_2020} detected the internal snow layers through a fully connected network and an iterative approach. Yari et al.\cite{yari_airborne_snow_radar} proposed a physics-based and a data-driven approach to process snow radar images and show that the results of the physics-based approach have better structural similarities while the data-driven approach generates results with better textural similarities. Ibikunle et al. \cite{ibikunle_snow_radar_echogram_2023} proposed the Skip\_MLP and LSTM\_PE method that adopts an iterative ``RowBlock'' methods to avoid the effect of small training data in analyzing the radar images. Wang et al.\cite{icelayers2021icme} proposed a network for general tiered segmentation tasks and applied it to detecting internal layers from snow radar images. Varshney et al. \cite{varshney_2021_regression_networks} proposed multi-output regression networks to give out individual predictions of internal ice layers in each output. Dong et al. \cite{Dong2022} proposed a deep neural network model, named EisNet, to remove the noises and extract the features of ice–bedrock interfaces and ice layers from ice-penetrating radar images. They improved the accuracy of their model by first pre-training the model with synthetic data and transfer training with real observational data. 

The applicability of CNN is not limited to image processing but extended to develop efficient statistical emulators for ice sheet flow models \cite{jouvet_cordonnier_kim_lüthi_vieli_aschwanden_2022}. Jouvet et al. \cite{jouvet_cordonnier_kim_lüthi_vieli_aschwanden_2022} developed a CNN ice flow emulator that provides vertically integrated 2-dimensional ice flow from the given ice thickness, surface slope, and basal sliding coefficient. Their CNN emulator made it possible to reduce computational time on ice flow modeling by utilizing graphic processing units (GPU).

Sea ice prediction and modeling is another popular field of CNN application. Andersson et al. \cite{Andersson2021} proposed a U-Net architecture named IceNet for monthly sea ice forecasting with six months of lead time. Grigoryev et al. \cite{Grigoryev2022} also used U-Net architecture for daily SIC forecasts in the Barents and Kara Seas, Labrador Sea, and Laptev Sea. Kim et al. \cite{Kim2020_SIC_CNN} used CNN to predict monthly SIC from past SIC records, sea surface temperature, air temperature, albedo, and wind velocity data. Similarly, the FCN proposed by Fritzner et al. \cite{Fritzner2020} used SIC, sea surface temperature, and air temperature from the previous six days to forecast SIC conditions for the next few weeks. Ren et al. \cite{Ren2022_SIC_CNN} designed an FCN architecture called SICNet to predict weekly SIC, but the sequences of satellite-derived SIC observations were only used as inputs of the model. They added a temporal-spatial attention module in their architecture to help the SICNet model capture spatiotemporal dependencies of the input SIC sequences. CNN was also used for the retrieval of daily sea ice thickness (SIT) by calculating the statistical relationship between thermodynamic parameters (i.e., sensible heat flux, latent heat flux, air/surface temperature) and SIT, with attaching self-attention blocks \cite{Liang2023}. Koo et al.\cite{koo2023multitask} proposed a multi-task FCN architecture called HIS-Unet that can predict both SIC and sea ice drift.

\subsection{Recurrent Neural Network}

Whereas CNN has significant strengths in finding spatial patterns in 2D images, recurrent neural networks (RNNs) are superior to other networks in processing sequential or temporal datasets. Based on this strength, RNN has been used in processing sequential data or predicting ice conditions. In particular, instead of the traditional RNN architecture that has issues with vanishing and exploding gradients on long input sequences, many studies use the long short-term memory (LSTM) architecture, which mitigates the issue that the original RNN has by implementing gated memory cells.

Zhang et al. \cite{Zhang2023} proposed the Convolutional Gate Recurrent Unit (ConvGRU) neural network model for the spatiotemporal prediction of movements of mountain glaciers in the Tibet Plateau. They proposed a U-net shape encoder and decoder architecture, where three ConvGRUs are inserted between the encoder and decoder. The integration of GRU and convolution networks allows for obtaining both time-series features and spatial features. Katwyk et al. \cite{Katwyk2023} used LSTM to approximate the sea level rise projections from the Ice Sheet Model Intercomparison for CMIP6 (ISMIP6). As an intercomparison of ice sheet models endorsed by the Climate Model Intercomparison Project-phase 6 (CMIP6), ISMIP6 simulations require extremely high complexity and computational cost. Therefore, they developed LSTM as emulators that approximate ice sheet models and sea level change, and this LSTM emulator shows 17 times faster computation time and higher accuracy than the traditional Gaussian Process architecture.

In the domain of sea ice, RNNs and LSTMs have been actively used to predict SIC \cite{Chi2017_LSTM, Zheng2022, Liu_seaice_ConvLSTM, Feng2023_ConvLSTM, Liu2021_LSTM_SIC, Choi2019_GRU} and sea ice drift \cite{Petrou_2017_RNN_seaicemotion}. Chi et al. \cite{Chi2017_LSTM} used an LSTM to predict one-monthly-ahead and one-year-ahead predictions of SIC, and their LSTM outperformed the MLP model. Since their LSTM model only uses the time series of satellite-derived SIC values as inputs, it does not require any external climatology or physical parameters. However, the model accuracy somehow deteriorated in the recent summer months of accelerated sea ice melting because the training data cannot represent such abnormal melting occurring in recent months. Zheng et al. \cite{Zheng2022} combined LSTM architecture with empirical orthogonal function (EOF) and principle component (PC) and deep MLP neural network for SIC prediction up to 100 days. Liu et al. \cite{Liu_seaice_ConvLSTM} used Convolutional LSTM (ConvLSTM) to forecast SIC in the Barents Sea from atmospheric and ocean reanalysis data (e.g., temperature, sea level pressure, wind, geopotential height, ocean heat content). This ConvLSTM architecture extracts both spatial and temporal relationships between different variables, preserving the physical consistency between predictors and SIC. Similarly, \cite{Feng2023_ConvLSTM} used ConvLSTM to conduct a 10-day prediction of SIC from 15-day advanced SIC and other ERA5 reanalysis weather data (e.g., sea surface temperature, air temperature, solar radiation, sea level pressure, wind). Liu et al. \cite{Liu2021_LSTM_SIC} also used ConvLSTM architecture to predict daily SIC, but they used only previous-day SIC values as predictors of the next-day SIC. Additionally, instead of LSTM, Choi et al. \cite{Choi2019_GRU} predicted a 15-day future SIC from the previous 120-day SIC by combining gated recurrent unit (GRU) on their neural network architecture, which works similarly to LSTM but has a simpler structure. Petrou et al. \cite{Petrou_2017_RNN_seaicemotion} designed an LSTM architecture to predict sea ice motion for the next 10 days using the image-derived sea ice motion for the previous 10 days.

\subsection{Graph Neural Networks}\label{GNN}

A graph neural network (GNN) is a deep learning method applied to graph structures where data points (nodes) are linked to each other by lines (edges). In graph architecture, the nodes of a graph are iteratively updated by exchanging information with their neighboring nodes via a message-passing framework. In the case of ice modeling, GNNs are useful for representing ice structures as graph-like structures and embedding the connection with the neighboring components. Lin et al. \cite{Lin2023_GNN} used GNN as a statistical emulator to approximate the relative sea level history induced by glacial isostatic adjustment (GIA). Since their model is applied to a global scale on a spherical Earth, they employed a graph-based spherical convolutional neural network (SCNN) algorithm instead of classical CNN, which is only available in Euclidean space. Zalatan et al. \cite{Zalatan2023, Zalatan_igarss, zalatan_icip} combined LSTM structure to graph convolutional network (GCN) architecture to determine snow accumulation from snow radar images of Greenland ice sheets. The integration of LSTM and GCN architectures allows this network to learn the temporal changes of the relationships between nodes in a graph and their spatial relationships. Rahnemoonfar et al.\cite{rahnemoonfar2024graph} use graph neural networks as efficient surrogate models of finite-element modeling on ice sheet. Regarding sea ice, Huang et al. \cite{Huang2021_GNN} examined the complex interaction between various atmospheric processes and sea ice variations by representing their casualty as GNN.

\subsection{Generative Adversarial Networks}\label{GAN}

The previous neural network architectures, including MLP, CNN, RNN, and GNN, are used to perform tasks of prediction or regression from certain data that has already been obtained. Beyond these traditional prediction models, generative models have been attempted in various polar ice studies to generate new and realistic data.

In training data-driven machine learning models for ice layer detection, it is crucial to collect large radar echogram datasets labeled properly. However, since these radar datasets are acquired during the Arctic and Antarctic fieldwork, the number of datasets and their spatial coverage can be limited. Furthermore, most of the available datasets are unlabeled, and the labeling process requires much time and effort from skilled experts. Therefore, significant attempts have been made to generate synthetic radar images as a supplement for training data-driven machine learning models. For this task, Generative Adversarial Networks (GANs) have been actively used \cite{Rahnemoonfar2019_GAN, Rahnemoonfar2020_GAN}. GAN is composed of two simultaneously trained parts: a generator and a discriminator. The generator produces realistic-appearing images, and the discriminator critic judges how the generated images are different from real images. By training these two models to compete against each other, the generator learns how to revise the misclassified images into more realistic images, and the discriminator learns how to catch the problems of generated images with respect to the ground truth. Based on CycleGAN framework \cite{Zhu2017_cycleGAN}, convolutional GANs were used to generate synthetic radar images \cite{Rahnemoonfar2019_GAN, Rahnemoonfar2020_GAN}. In the case of sea ice, Kim et al. \cite{Kim2023_GAN} proposed a deep convolutional GAN called PolarGAN for generating artificial images of Arctic SIC that satisfy user-defined geometric preferences.

GANs can also be used to reproduce the bed topography map of ice sheets called DeepBedMap \cite{Leong2020_DeepBedMap}. The generator of the DeepBedMap produces a high-resolution (250 m) Antarctica's bed elevation from a low-resolution (1000 m) bed elevation model (BEDMAP2 \cite{Bedmap2}). The generator and discriminator of DeepBedMap follow the architecture of Enhanced Super-Resolution Generative Adversarial Network (ESRGAN) \cite{wang2018_ESRGAN}, but they customized the input block to include conditional inputs of ice surface elevation, ice velocity, and snow accumulation grids. The discriminator produced the similarity scores to real images, and this score was used by both the generator and discriminator to tune the predictions toward more realistic bed elevations. Cai et al.\cite{multibranch} enhance the result of DeepBedMap by dividing the inputs into two groups and adopting a multi-branch version of DeepBedMap.

\subsection{Limitation of data-driven models}

Given that the performance of fully data-driven machine learning models is highly dependent on the size and representability of training datasets, it is essential to collect enough high-quality ground truth data. Despite many efforts to collect large datasets to improve the model fidelity, the number of ground truth data is extremely limited for the polar regions. Since it is challenging to collect field measurement data directly in polar regions, most of these data-driven studies rely on data from (1) satellite observations or (2) limited fieldwork. Although satellite data provides regular and large-scale observations, it has limitations in its resolution, precision, and quality. On the other hand, although data from fieldwork could have better resolution and precision, fieldwork is expensive, and the data from fieldwork is limited to specific seasons or regions of field campaigns. Furthermore, data-driven approaches generally require a lot of human effort and time to label enough data. Therefore, obtaining enough high-quality data is the biggest challenge for fully data-driven machine learning models.

\section{Methodology for incorporating physical law with machine learning techniques}\label{piml_method}
In Section \ref{physics} and Section \ref{datadriven}, we found that physical models can provide precise estimation, but it is computationally expensive and requires well-developed physical processes; data-driven models can effectively process large input data and learn underlying patterns, but they require high-quality ground-truth labels and may make predictions that have little physical meaning. Therefore, it raises the question: Is it possible to form a hybrid way that combines these two kinds of traditional methods? This is where physics-informed machine learning will be helpful.

Physics-informed machine learning is a promising framework that can combine physical law with machine learning algorithms, leveraging the advantages of both and handling the limitations of pure physical models and data-driven models. In this section, we will present the methodologies of physics-informed machine learning and how physics can be incorporated with machine learning.

There are a few ways to integrate physical relations with machine learning algorithms. In different papers, the authors may use different taxonomies and terminologies, while the core idea of each method will be similar. Karniadakis et al. summarized those methods of combining physics and data-driven approaches in the aspect of different kinds of bias: observational biases, inductive biases, and learning biases\cite{Karniadakis2021}. 
For machine learning algorithms, proper biases are needed to form the generalization ability\cite{needforbias}. 

Physics can be introduced through observational biases. Observational bias is the most straightforward way to introduce physics-related biases into a machine learning algorithm. If sufficient data that represents the underlying physical mechanisms are available in most of the input domains, then it is possible to train a machine learning algorithm directly on such data to learn a representation of the underlying physics. For example, for certain physical laws with unknown governing equations, if we can sample enough data points from experiments, we can achieve better data-fitting equations that represent the governing equation, and the gaps between observational data can be precisely estimated through the fitted equation. Thanks to the development of experiment techniques and measuring equipment, it is possible to collect high-fidelity observational data of different spatial and temporal scales and then use them as soft constraints for machine learning algorithms\cite{Karniadakis2021} to learn the equation. However, one major limitation of observational biases is that for over-parameterized deep learning models, a large amount of observational data is needed to ensure the models can make predictions that have physical meaning. Moreover, for large-scale physical processes, the cost can be very expensive.

Physics can be introduced through inductive biases. Inductive biases refer to incorporating prior physical knowledge in a customized machine learning model architecture. Prior physical knowledge will be expressed as rigorous mathematical formulas and serve as hard constraints in the whole algorithm. As a result, the model predictions will be guaranteed to satisfy the given formulas automatically. However, the understanding of the corresponding physics is the major limitation. A few physical knowledge don't have a commonly agreed mathematical formula, and inductive biases are only suitable for those well-known, widely agreed physical laws. Moreover, these kinds of implementation cannot be extended to complex problems due to the difficulty of scaling\cite{Karniadakis2021}. 

Physics can also be introduced through learning biases. Unlike inductive biases, which are hard constraints, learning biases serve as soft constraints by penalizing the loss function with physics-based terms. By combining a physical loss with the usual data fitting loss, a machine learning algorithm is trained to minimize the data fitting error and the physical error, and both approximately fit the observation and the underlying physical law. It also creates the flexibility to adjust the weights for different loss terms to let the model fit the observation or the physics better. The commonly used format of physical loss is the residuals of governing equations\cite{PINN}.

\begin{table}
\centering
\caption{Summary of combination methods, advantages and limitations}
\label{summarytable}
\begin{tabular}{cm{5cm}m{5cm}}
\toprule
Methods                           & Advantages                                                             & Limitations                                               \\ \midrule
Observational Biases              & The most straightforward method                                        & Expensive Cost                                            \\ \midrule
\multirow{2}{*}{Inductive Biases} & Hard constraints                                                       & Hard to Scale                                             \\
                                  & Guaranteed to satisfy                                                  & Only work for well-developed physics                      \\ \midrule
\multirow{3}{*}{Learning Biases}  & Scalable                                                               & Bad choice of weights for loss term may affect the result \\
                                  & Easy implementation in code                                            &                                                           \\
                                  & Flexibility to adjust weights between data fitting and physics fitting &                                                           \\ \bottomrule
\end{tabular}
\end{table}

As summarized in Table \ref{summarytable}, these three methods have their own advantages and limitations. Therefore, in real-world applications, it is more practical to form a hybrid method that combines different kinds of biases together. In this paper, we will use our own taxonomy to describe how physical knowledge is incorporated with machine learning algorithms, which are: combination through loss function, where learning biases are introduced; combination through model architecture, where inductive biases are introduced; and combination through training strategy, where both observational biases and learning biases are introduced. 

A typical example of physics-informed machine learning is the physics-informed neural networks (PINNs) proposed by Raissi et al.\cite{PINN}. PINNs are proposed for solving partial differential equations (PDEs) with certain boundary conditions and initial conditions. The authors used a multi-layer perceptron as the surrogate of the PDE solution. The original partial differential equations are integrated into the loss function as the equation loss, which is the original PDE equation's residual form. All the necessary derivatives are calculated based on automatic differentiation on the network output. Data fitting loss represents the fitting between the learned solution and corresponding boundary conditions or initial conditions. By minimizing both the data fitting loss and the equation loss through gradient-based optimizers, the network can generate a solution that satisfies both the equations and the known conditions. 

\section{Case Study: Physics-Informed Machine Learning on Ice}\label{case}
In this section, we will review the application of physics-informed machine learning on polar ice. Unlike the wide studies and various physics or data-driven machine learning methods on polar ice, the study of physics-informed machine learning on polar ice is still in its infancy. We will categorize these papers based on how the author incorporates physical law into machine learning algorithms, summarized in Table \ref{summary}.

\begin{table}
\caption{Summary of relevant papers of physics-informed machine learning on polar ice}
\label{summary}
\centering
\scalebox{0.8}{
\begin{tabular}{cccc}
\toprule
Paper and Authors & Related Physics & Combination Method & Network Architecture\\
\midrule
Teisberg et al.\cite{teisberg} & Mass Conservation Law & Loss Function & Multi-layer Perceptron\\

 Riel et al. \cite{slip} & Shallow Shelf Approximation& Loss function& Multi-layer Perceptron \\

Wang et al.\cite{rheologyiceshelf} & Shallow Shelf Approximation & Loss function & Multi-layer Perceptron\\

Riel et al.\cite{riel_minchew_2023_variational} & Shallow Shelf Approximation & Loss function& Multi-layer Perceptron \\

Iwasaki et al.\cite{Hardness} & Shallow Shelf Approximation & Loss function& Multi-layer Perceptron\\
\midrule

Varshney et al.\cite{varshney2021refining} & Wavelet transformation & Model Architecture & Convolutional Neural Network \\

He et al.\cite{HE2023112428} & DeepONet & Model Architecture & DeepONet \\
\midrule

Kamangir et al.\cite{kamangir} & Wavelet transformation & Training Strategy & Convolutional Neural Network \\

Varshney et al.\cite{Varshneyphysicslabels} & Atmospheric model & Training strategy & Convolutional Neural Network \\

Jouvet et al.\cite{jouvet_cordonnier_2023} & Blatter–Pattyn Approximation & Training strategy & Convolutional Neural Network \\
\bottomrule
\end{tabular}}
\end{table}

\subsection{Combination through loss function}
As stated in Section \ref{piml_method}, physical equations can be added to the loss function of machine learning algorithms as a physical loss term. Raissi et al.\cite{PINN} use the residual form of the PDEs as the physical loss term, and then this choice is widely extended to different subdomains, including the study of polar ice. Moreover, this combination method may only change the loss function and keep the rest of the model unchanged, which is beneficial in enhancing some existing data-driven methods by adding physical components.

\subsubsection{A machine learning approach to mass-conserving ice thickness interpolation}

Teisberg et al.\cite{teisberg} designed a modified physics-informed neural network to predict the ice thickness and velocities as a function of the spatial coordinates. Compared with classic numerical methods, the proposed physics-informed neural network method embeds mass conservation constraints into the machine learning algorithm. It is totally mesh-free, and all the needed derivatives are calculated via automatic differentiation\cite{teisberg}. Therefore, the model can be trained without pre-defining a desired resolution for the interpolation result\cite{teisberg}.

The network architecture is a feed-forward network with five hidden layers\cite{teisberg}. Physics is introduced as a soft constraint via the loss function. The mass conservation constraint of ice flow can be simplified as follows:
\begin{equation}
    \nabla\cdot(h\overline{\boldsymbol{u}})=\dot{a}
\end{equation}
where $h$ is ice thickness, $\overline{\boldsymbol{u}}$ is the depth-averaged ice velocity and $\dot{a}$ is the apparent mass balance\cite{teisberg}. The mass conversation loss term is defined as the mean squared residual of the mass conservation equation.

For the velocity loss term, the authors compare the method of allowing symmetric error bounds around the surface velocity measurement with the method of separating different loss terms for direction and magnitude of velocity vectors and allowing asymmetric error bounds\cite{teisberg}. For the latter design of the loss term, they also introduce a smoothing term defined as the mean squares of the derivatives of the velocity difference between surface and depth-averaged.

\begin{figure}[h]
  \centering
  \includegraphics[width=0.85\linewidth]{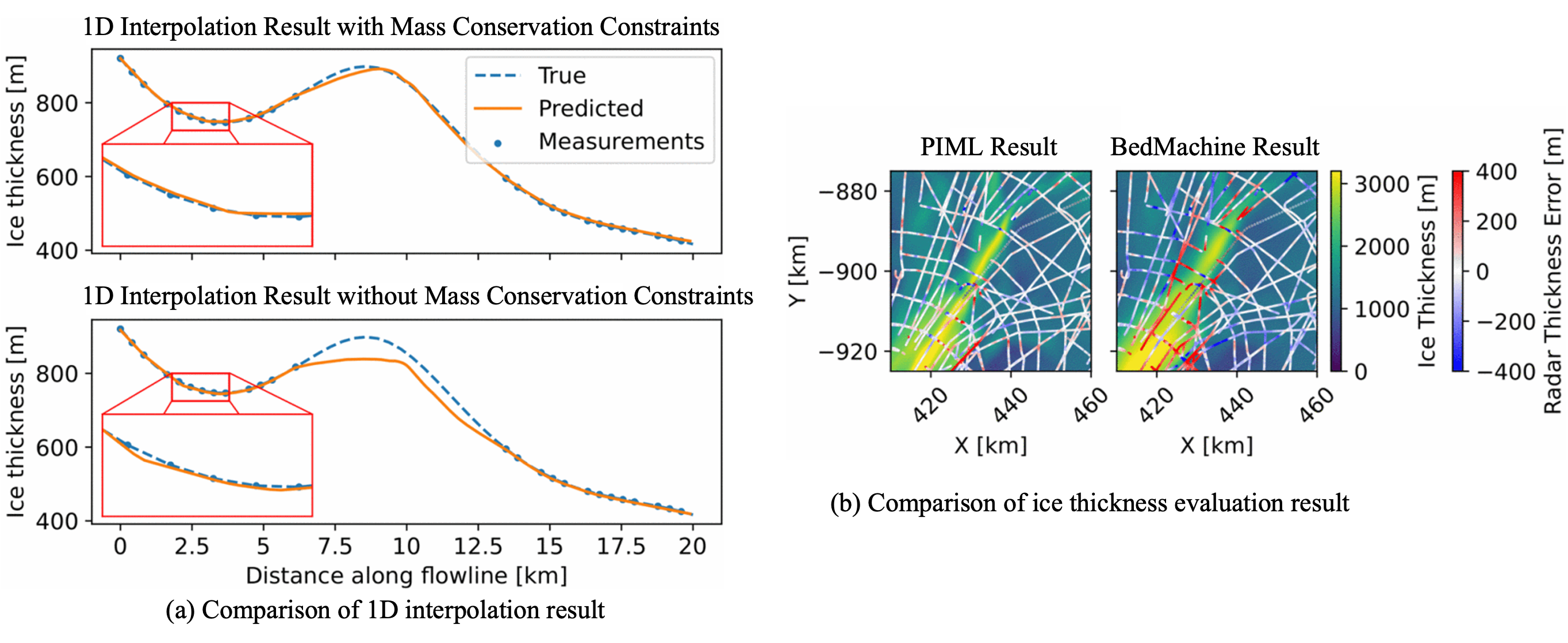}
  \caption{(a): A comparison of 1D interpolation result showing the importance of mass conservation constraint. (b): A comparison between the result of the network proposed by Teisberg et al. and BedMachine result. Overlaid lines represent the used radar data locations, with red indicating that the predicted thickness exceeds the measured radar thickness. Figure reproduced from the original paper\cite{teisberg}.}
  \Description{Left part of the figures shows two 1D interpolation results. One of them is with mass conservation constraints and the other is without mass conservation constraints. By comparison, it shows that mass conservation can make the predicted results closer to the true results. Right of the figure is a comparison between the PIML network proposed by Teisberg et al. and BedMachine result. It shows that PIML results have lower error.}
  \label{Teisberg Result}
\end{figure}

As shown in Fig.\ref{Teisberg Result}(a), introducing the mass conservation loss term contributes to improving the thickness prediction in the area with fewer radar measurements and can significantly avoid artifacts or arbitrary thickness predictions\cite{teisberg}. The proposed physics-based network has a mean squared error of 70 meters, while it is 130 meters for BedMachine Result\cite{teisberg}. This result shows the great potential of training a more efficient interpolation network by adopting PDE constraints in the loss function. This proposed physics-informed network may also serve as an exploratory result for developing new methods to capture multiple ice physical processes.

\subsubsection{Data-Driven Inference of the Mechanics of Slip Along Glacier Beds Using Physics-Informed Neural Networks: Case Study on Rutford Ice Stream, Antarctica}

Riel et al. \cite{slip} proposed a hybrid network framework combining the known physical laws for ice dynamics with a neural network representing the unknown sliding law to quantify the glacier basal drag evolution. One major part of the uncertainty of glacier dynamics is the unknown parameterization of glacier base drag and its relation to other glacier properties. Due to the fact that there is not a commonly agreed form of the sliding law, and the basal drag is not directly observable, it becomes essential to combine some observable quantities, like ice velocity and elevation, with known physical models to make the inference.

\begin{figure}[h]
  \centering
  \includegraphics[width=0.8\linewidth]{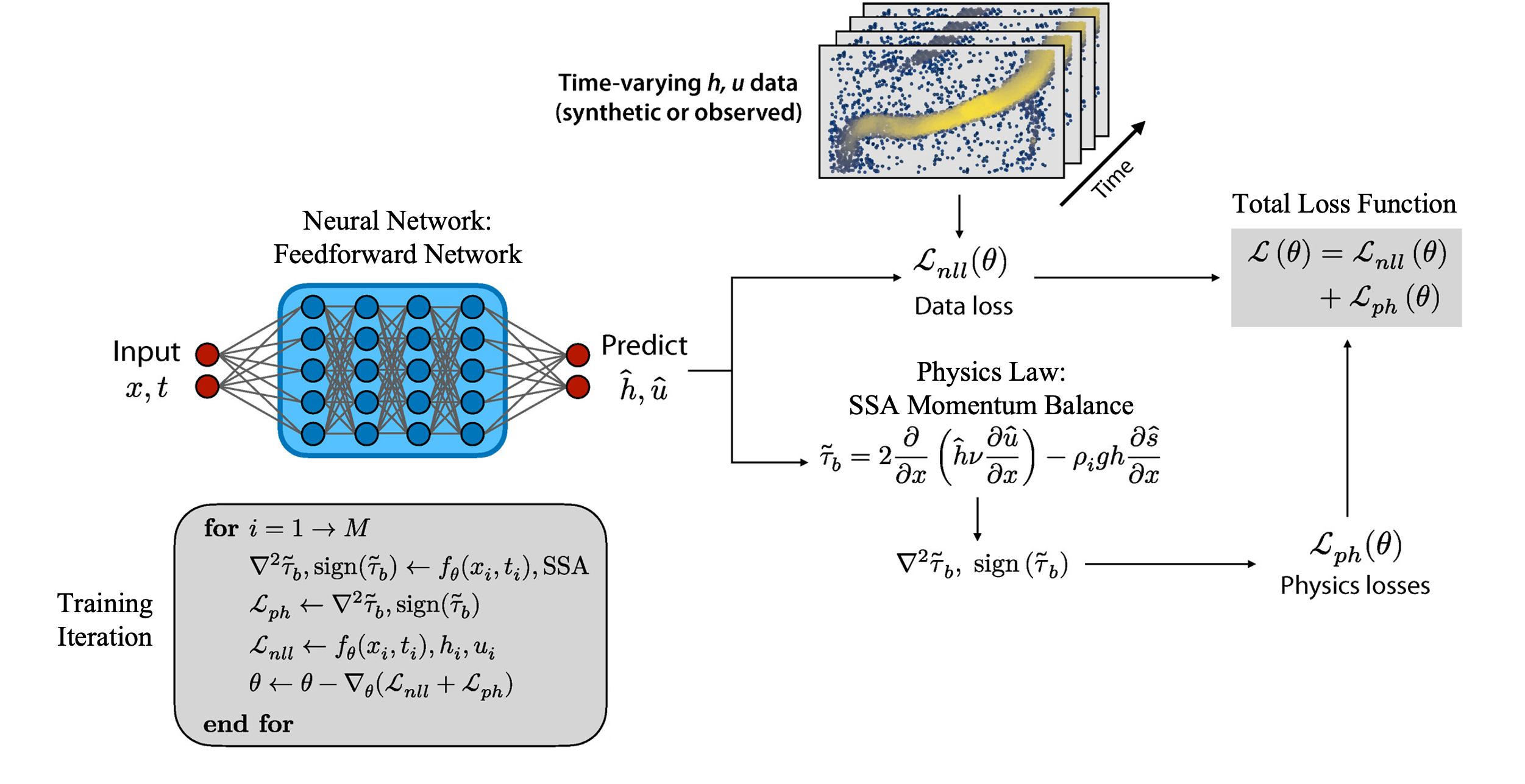}
  \caption{Network architecture diagram and loss function design for inferring the slip mechanism. Reproduced from original paper\cite{slip}.}
    \Description{In the figure, it shows that a feedforward network is used. A data loss is calculated based on network prediction and synthetic or observed grouth data. A physics loss is calculated based on SSA momentum balance and the network prediction. The total loss function is the sum of data loss and physics loss. A sample training iteration is also provided.}
  \label{slip}
\end{figure}

Derived from the SSA equations (Equation\ref{SSA1} and Equation\ref{SSA2}) for ice flow, a traditional force balance method is proposed to quantify drag variations based on surface velocity, ice geometry, and ice rheology. However, the force balance method requires the first and second-order derivatives of those observable quantities like velocity or thickness. Instead of calculating these derivatives with traditional interpolation methods, neural networks can learn a continuous hypersurface, and the derivatives can be calculated through automatic differentiation. In Fig.\ref{slip}, a fully connected network is used to represent the sliding law, predicting surface velocity and ice thickness based on spatial coordinates and time. In order to control the effect of noise in the surface observation quantities and avoid the un-physical inference of basal drag, they designed a customized physical loss term in addition to the mean square error for data fitting, which integrates the SSA momentum balance as the prior physical knowledge\cite{slip}. The physical loss term is constructed with a Laplacian smoothing term to penalize the roughness and a rectified linear unit (ReLU) term to penalize the sign of predicted drag\cite{slip}. By adopting the physical loss term, it is possible to prevent overfitting even for a small-data scenario\cite{PINN}.

The proposed method can effectively reconstruct the drag evolution without specifying a particular form of the sliding law. It can integrate existing physical knowledge with remote sensing observation and form a more accurate model. Ultimately, this proposed method shows the potential to quantify the data and modeling uncertainties under a geophysics background and can be extended to other related fields.

\subsubsection{Discovering the rheology of Antarctic Ice Shelves via physics-informed deep learning}

\begin{figure}[h]
  \centering
  \includegraphics[width=0.95\linewidth]{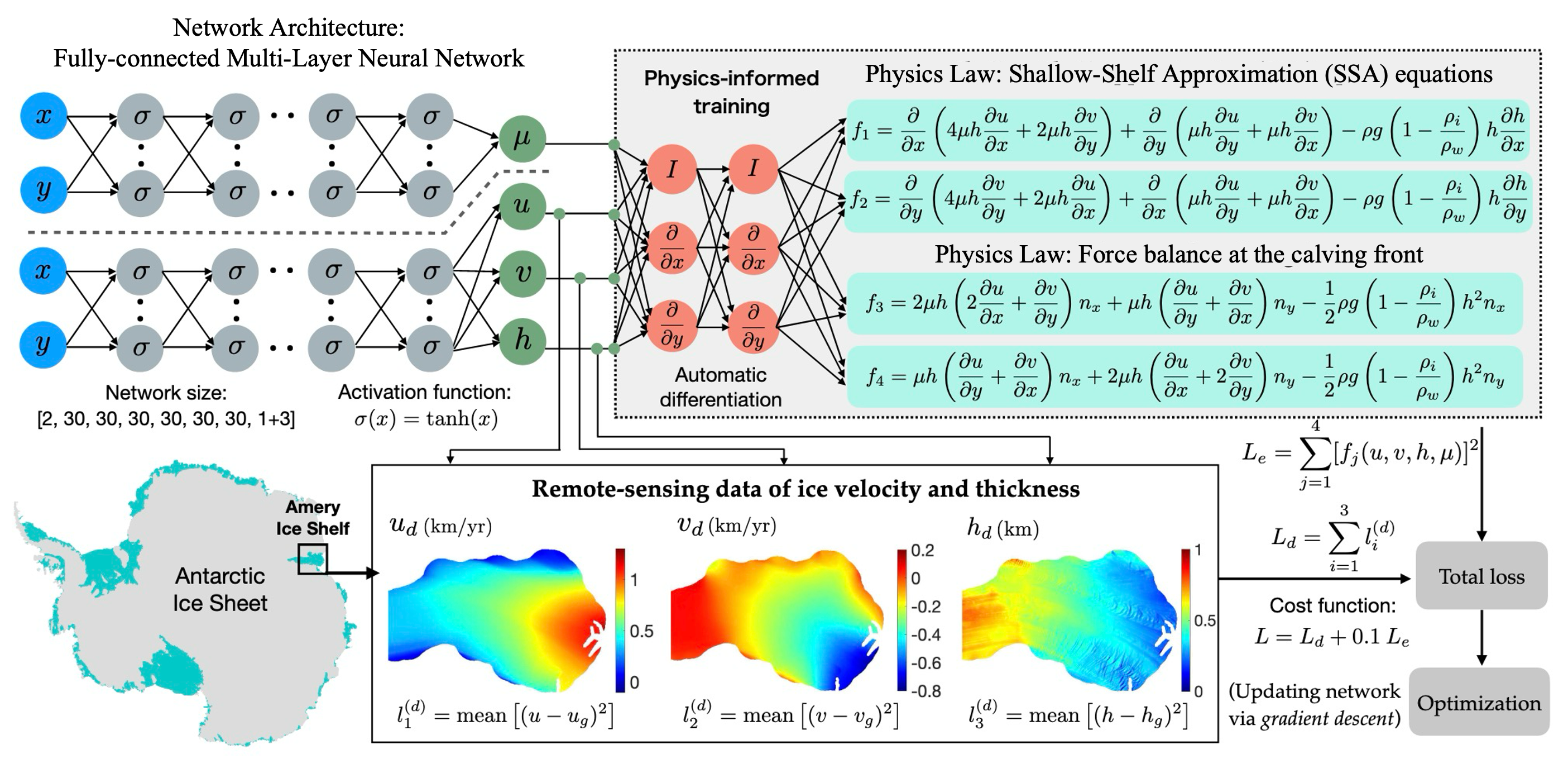}
  \caption{Network Architecture and workflow for estimating ice viscosity. Reproduced from original paper\cite{rheologyiceshelf}.}
  \Description{The figure shows that a fully-connected multi-layer neural network is used. The network prediction is used in the physics-informed training with automatic differentiation. Total loss is calculated base a data fitting loss between network prediction and remote-sensing results, and a physical loss based on network prediction, force balance and shallow-shelf approximation.}
  \label{viscosity}
\end{figure}

Wang et al. propose a method that uses physics-informed neural networks to solve the ice viscosity without assumptions of its dependence on strain rate\cite{rheologyiceshelf}. As shown in Fig.\ref{viscosity}, two fully connected networks are introduced, with one aiming at fitting known ice velocity and thickness variables and the other aiming at predicting the unknown ice viscosity. Data misfit loss is calculated based on observations and network ice thickness and velocity prediction. 

Physics is incorporated as a customized loss term in the total loss function. Wang et al. use the SSA and force balance as the governing physical equation, calculating the sum of equation residuals as the equation loss term. By minimizing both the data misfit loss and the equation loss, the network can estimate ice viscosity precisely without any prior observation data of ice viscosity\cite{rheologyiceshelf}.

\begin{figure}[h]
  \centering
  \includegraphics[width=0.95\linewidth]{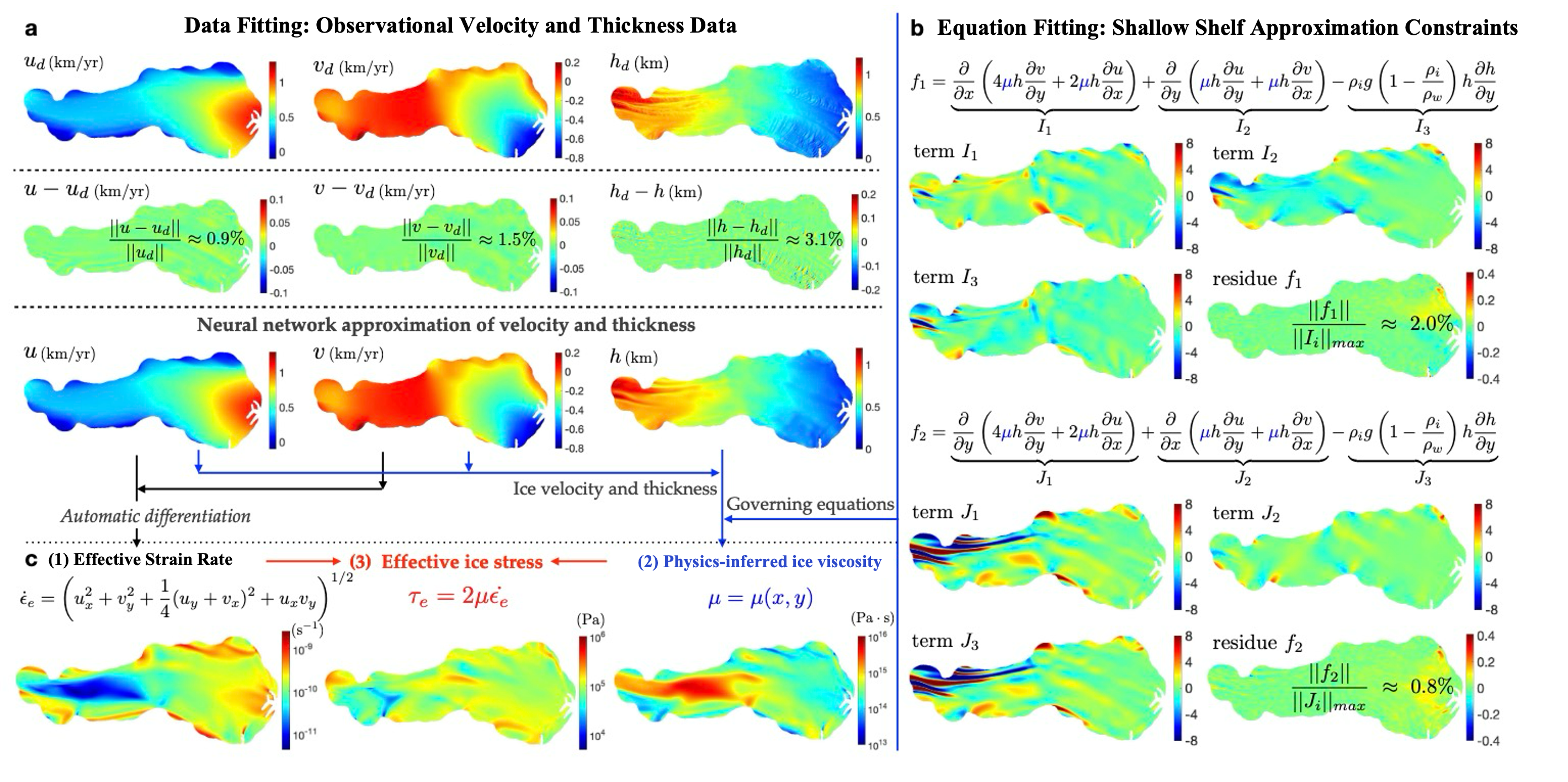}
  \caption{Network Prediction Verification. a. Comparison between observation and prediction of ice velocity and thickness. b. Verification of equation residual, showing that the PINN predictions satisfy the shallow-shelf approximation equation. c. Estimation of ice stress based on strain rate and ice viscosity. Reproduced from original paper\cite{rheologyiceshelf}.}
  \Description{In the figure, it shows a comparison between observation and prediction of ice velocity and thickness in part a. In part b, it shows the equation fitting based on shallow-shelf approximation. In part c, it shows that predictions from part a and b are used to get effective strain rate and physics-informed ice viscosity, and then to achieve effective ice stress.}
  \label{viscosityre}
\end{figure}

Fig.\ref{viscosityre} shows that physics-informed neural network proposed by Wang et al.\cite{rheologyiceshelf} can effectively give predictions of ice velocity and thickness that are close to the real observational data and can infer the ice viscosity that satisfies the underlying physical equation precisely. Combining all the network predictions makes it possible to obtain the ice stress that can be used in the ice shelf rheology study. The authors also show that the stress-strain rate relationship in the ice compression zone fits the power laws better, with an exponent varying between 1 and 3\cite{rheologyiceshelf}. The rheology is different in the ice extension zone, and the relationship is currently more uncertain.

\subsubsection{Variational inference of ice shelf rheology with physics-informed machine learning}
Riel et al. proposed a physics-informed machine learning framework to predict the distribution of ice rigidity, given a spatial domain with coordinates, and can integrate with prior physical knowledge\cite{riel_minchew_2023_variational}. Ice rigidity is the key factor that governs the resistance of ice flow. Glen's flow law is the common physical law to define the relationship between ice rigidity, stress, stress exponent, and strain rate. Due to the fact that rigidity and stress exponent are the result of multiple mechanisms that cannot be directly observed, ice rigidity is usually inferred from surface observations.

\begin{figure}[h]
  \centering
  \includegraphics[width=0.8\linewidth]{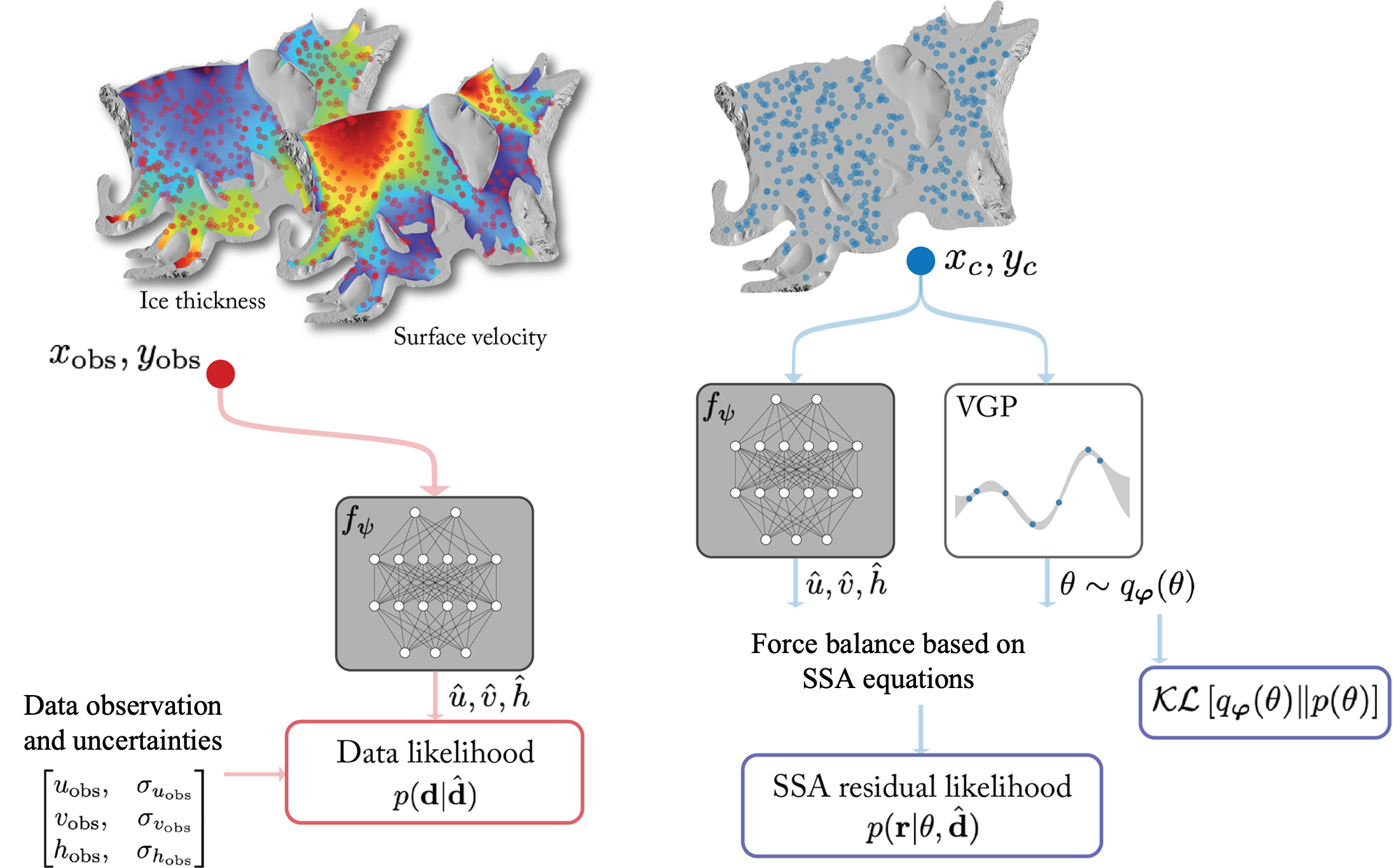}
  \caption{Framework of physics-informed variational inference for ice rigidity. Reproduced from original paper\cite{riel_minchew_2023_variational}.}
  \Description{In the left, it shows that a multi-layer perceptron is trained with data likelihood as loss function and observation data for ice thickness and surface velocity. In the left, the trained network are used together with a variational gaussian process, with KL-divergence and SSA residual likelihood as loss function, to do physics-informed variational inference for ice rigidity.}
  \label{variational}
\end{figure}

The proposed framework is a combination of neural networks to approximate surface observation continuously and variational Gaussian Processes to estimate the probabilistic distribution of ice rigidity. Through Bayes' theorem, Riel et al. define the joint posterior distribution for rigidity and reconstructed surface observation and then convert the estimation problem into a variational inference framework, where the goal is to minimize the Kullback-Leibler divergence between machine learning reconstructed distributions and the joint posterior distribution \cite{riel_minchew_2023_variational}. In Fig.\ref{variational}, a fully connected network is used to reconstruct surface observations continuously, and a variational Gaussian Processes to approximate the posterior distribution of ice rigidity. The total loss function contains a data likelihood loss (evaluated based on remote sensing surface observations), an SSA residual likelihood loss, and a Kullback-Leibler divergence term (evaluated at independent sampled coordinates).

Physics is integrated into the customized physical loss term. SSA momentum balance can be written in the residual format based on Equations \ref{SSA1} and \ref{SSA2}:
\begin{equation}
r_x = \frac{\partial}{\partial x} \left( 2\mu H \left( \frac{2 \partial u}{\partial x} + \frac{\partial v}{\partial y} \right) \right) + \frac{\partial}{\partial y} \left( \mu H \left( \frac{\partial u}{\partial y} + \frac{\partial v}{\partial x} \right) \right) - \rho g H \frac{\partial s}{\partial x}
\end{equation}
\begin{equation}
r_y = \frac{\partial}{\partial y} \left( 2\mu H \left( \frac{2 \partial v}{\partial y} + \frac{\partial u}{\partial x} \right) \right) + \frac{\partial}{\partial x} \left( \mu H \left( \frac{\partial u}{\partial y} + \frac{\partial v}{\partial x} \right) \right) - \rho g H \frac{\partial s}{\partial y}.
\end{equation}
where $u$ and $v$ are the velocity components, $H$ is ice thickness, $s$ is the surface elevation, $\mu$ is the viscosity of ice (defined as a function of ice rigidity), $\rho$ is ice density and $g$ is the gravitational acceleration. These residual terms are the basal drag for the ice stream. As the viscosity of seawater is negligible, $r_x$ and $r_y$ are nominally 0. Therefore, in the proposed variational inference framework, the distribution of ice rigidity is constructed to let $r_x$ and $r_y$ close to zero, and the likelihood of SSA residual is used as the physical loss term.

Riel et al. tested the proposed method on synthetic 1D and 2D ice shelves. They find that uncertainties achieve the lowest value when stress and strain rates are higher\cite{riel_minchew_2023_variational}. They also show that the proposed forward model can reduce bias in certain regions where the shallow shelf approximation of momentum balance is not satisfied\cite{riel_minchew_2023_variational}. Combining the variational inference framework with the traditional inversion methods makes it possible to lower the uncertainties in a further step.

\subsubsection{One-dimensional ice shelf hardness inversion: Clustering behavior and collocation resampling in physics-informed neural networks}

Iwasaki et al. investigated the capacity of the physics-informed neural network (PINN) that is used in ice shelf hardness inverse problem, finding that there exist problems when the training data has significant noise, and proposed algorithms to handle those problems\cite{Hardness}. Ice shelves are a kind of floating extensions that have an essential role in ice flow modeling\cite{Hardness}. The dynamic of ice shelves is governed by the ice hardness (B), a property that cannot be measured directly. Traditionally, ice hardness is calculated by inversion methods like control methods that are widely used in glaciology \cite{control1,control2,control3}. These methods will enforce ice thickness data and momentum equations as hard constraints and may require interpolation between observational data to ensure the same resolution as numerical models\cite{Hardness}. Therefore, these control methods rely strongly on the accuracy of input ice thickness data. However, even the best ice thickness data that are currently available can have a considerable uncertainty of 100 meters, which can slightly affect the inversion process on ice hardness estimation\cite{Morlighem}. 

In order to reduce the needed amount of training data and reduce the noise, physics-informed neural networks are used as soft loss function constraints in the inversion problem of ice hardness estimation\cite{Hardness}. The PINN network, with architecture as a simple feedforward neural network, is built to predict velocity $u(x)$, thickness $h(x)$, and hardness $B(x)$ of one-dimensional ice shelf based on noisy velocity and thickness training data. 

Physics is incorporated in the custom loss function for the PINN network. The loss function is defined as:
\begin{equation}
    J(\Theta) = \gamma E(\Theta)+(1-\gamma)D(\Theta)
\end{equation}
where $E(\Theta)$ represents equation loss, which is the error of fitting to known physical laws, $D(\Theta)$ represents data loss, which is the error of fitting to training data, and $\gamma$ is the hyperparameter used to adjust weights of different loss term, which represents the adjustment between fitting to the training data and fitting to known physical laws. For ice shelf hardness estimation, data loss is the mean squared error of predicted velocity and thickness compared to the true value. The equation loss is the residual of the modified SSA equation, which serves as the governing equation for ice hardness \cite{SSA1,SSA2}.

\begin{figure}[h]
  \centering
  \scalebox{0.6}{
  \includegraphics{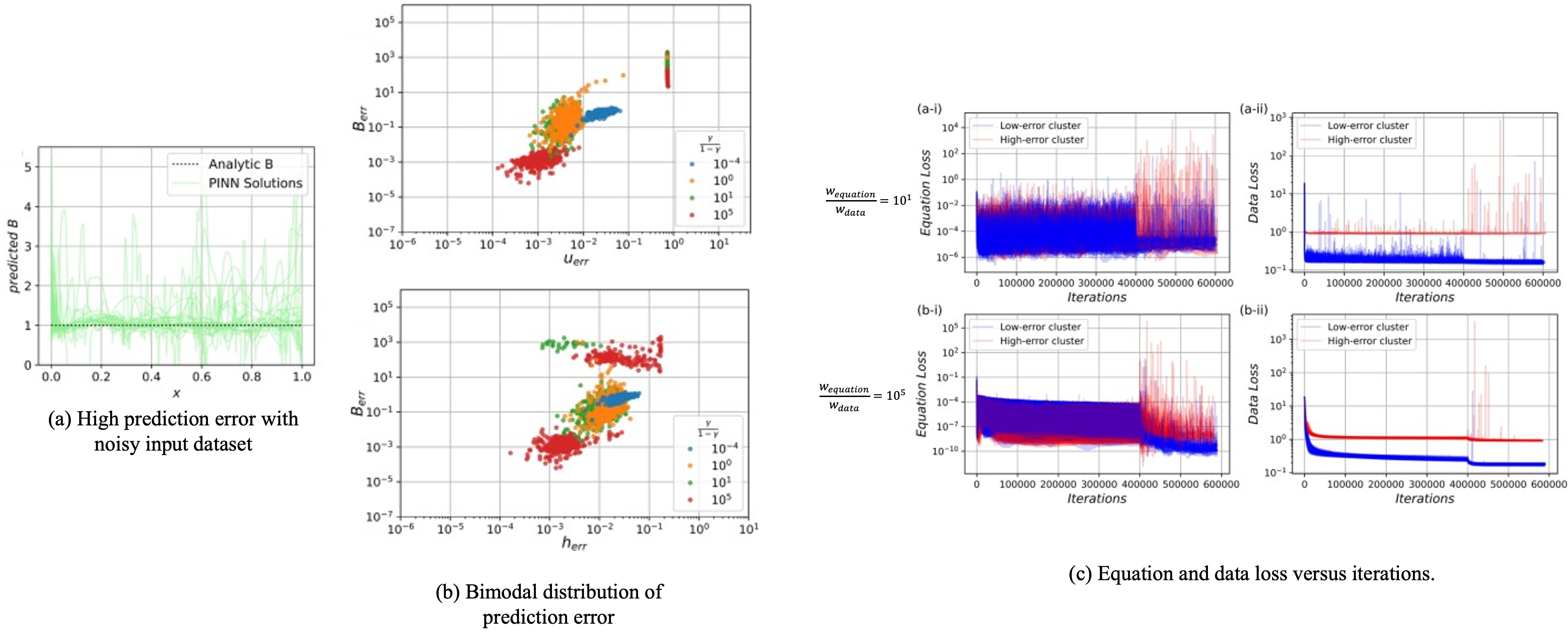}
  }
  \caption{(a): High prediction error exists when there exists high noise in the training data and weights of equation loss and data loss are equal. (b): Error for $u$, $h$, and $B$ are bimodally distributed when $\gamma/(1-\gamma)>10^0$. (c) Equation and data loss versus iterations. Loss trajectories of the first 100 iterations are overlayedReproduced from original paper\cite{Hardness}.}
  \Description{In part a of the figure, it shows that when there exists high noise in training data, and the weights of equation loss and data loss in PINN are equal, there will be high prediction error. In part b, it shows that the prediction error are bimodally distributed.In part c, it shows how equation loss and data loss are changed in the training process with different weights. Results highlights that the the two clusters for bimodal distribution form very early during the training process.}
  \label{bimodally}
\end{figure}

The authors tested the performance of PINN and the effect of $\gamma$ for different noise levels in the training data by adding Gaussian noise with different standard deviations to clean training data\cite{Hardness}. When the noise level is high, especially the same magnitude as the estimated error of radar-derived ice thickness data, an equal weight between the equation loss and data loss will result in a poor prediction of B, shown in Fig.\ref{bimodally}(a). They then tested the model prediction error for u, h, and B with different levels of noise and different values for hyperparameter $\gamma$. Increasing the value of $\gamma$ will push the network to better fit with the underlying equation while decreasing $\gamma$ meant to better fit with the training data. They found that the error for u, h, and B will be bimodally distributed in a log scale when $\gamma/(1-\gamma)>10^0$\cite{Hardness}. As shown in Fig.\ref{bimodally}(b), k-means clustering divides these errors into a low-error cluster corresponding to the desired solution and a high-error cluster poorly fitting the ground truth.

As shown in Fig.\ref{bimodally}(c), experiments show that two clusters form very early, about 50000 iterations. Additional experiments show that PINN is quite sensitive to the parameter initialization. The reason for the high-error cluster may be the situation in which the model learns a wrong sign for the velocity spatial gradient.

Iwasaki et al. also proposed a collocation resampling method to improve the performance of PINN on noisy training data, where the collocation points used to evaluate equation loss are resampled after each iteration\cite{Hardness}. Their result shows that it can significantly improve accuracy with less training time. The proposed collocation resampling method does not need to weigh the equation loss over the data fitting loss and avoid the bimodal error distribution.

\subsection{Combination through model architecture}
As discussed in Section \ref{piml_method}, physics law can be incorporated as hard constraints in the model architecture design, ensuring that the model predictions are guaranteed to satisfy the corresponding physical law. Moreover, we may add a physics-based stage before or after the feature extraction and pattern recognition stages, where the physics-based stage serves as the pre-processing or post-processing steps. In this section, we do a case study on a few relevant papers with a physics-based part in their network architecture.

\subsubsection{Reﬁning Ice Layer Tracking through Wavelet combined Neural Networks}

Varshney et al.\cite{varshney2021refining} use the physics of signal through wavelet transform to better extract ice layer boundary information from the snow radar echogram images. Their results show that noise in snow radar echogram images is the major problem for extracting better layer information. Wavelet transform is a well-used signal processing technique\cite{mallat1989}. It can capture contextual and textual information in multiple scales and reduce noise in the raw image\cite{waveletcnn}. 

\begin{figure}[h]
  \centering
  \includegraphics[width=\linewidth]{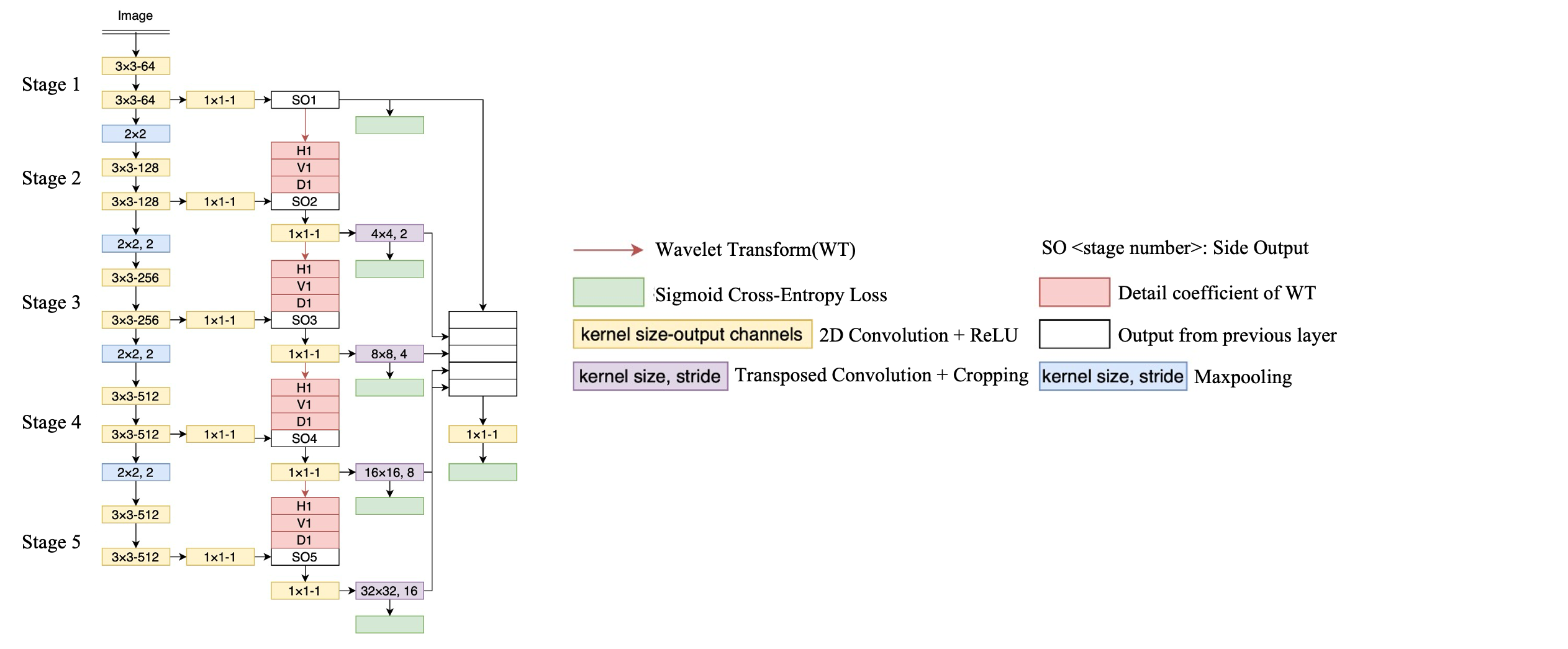}
  \caption{Wavelet combined network architecture for image denoising and ice layer boundary detection. Reproduced from original paper\cite{varshney2021refining}.}
  \Description{The figure shows the detailed network architecture of image denoising and ice layer boundary detection.}
  \label{refinearchi}
\end{figure}

As shown in Fig.\ref{refinearchi}, Varshney et al. used a VGG13 architecture as the backbone network structure, added a 1*1 convolution before each max pooling layer, and created an intermediate feature map. Then, they implemented level 1 Discrete Wavelet Transform to $i^{th}$ intermediate feature map and concatenate the detail coefficients of the DWT with the $(i+1)^{th}$ side output\cite{varshney2021refining}, forming a connected feature map. Those feature maps are sampled to the size of the input image by transposing convolution and clipped to eliminate subtle size differences. Finally, the feature maps of the five stages are cascaded and convoluted to form a ``fuse'' layer\cite{varshney2021refining}. The network is trained with deep supervision and calculates the cumulative loss function for each stage as well as the cascade output. Physics of signal is incorporated with the wavelet transformation. Adding wavelet transformation and combination to backbone CNN shows that wavelet can reduce the noise in the raw image, extract those sharp ice layer edges, and detect deeper layers.

Varshney et al. show that combining wavelet transformation and CNN makes it possible to improve identifying ice layers from snow radar images. Their results show that wavelet, especially db2 wavelet, can detect deeper and sharper ice layer edges\cite{varshney2021refining}. However, there is a tradeoff between sharper output and layer completeness. Moreover, by incorporating wavelet transforms after a convolutional layer, it is helpful to build more ``learned'' detail coefficients, as the transform depends on the changing weights of that convolutional layer\cite{varshney2021refining}. This combination of CNN and wavelet transform can be applied to multiple ice datasets from different years, quantify the ice layer changes, and make the climate model more predictable\cite{varshney2021refining}.

\subsubsection{A Hybrid Deep Neural Operator/Finite Element Method for Ice-Sheet Modeling} 

He et al. proposed a hybrid ice flow model that combines the classic finite element method with a physics-informed machine learning model to approximate existing ice sheet dynamics models with less cost\cite{HE2023112428}. Eq. \ref{mass_conservation_thickness} can be discretized as: 
\begin{equation}
\left\{
\begin{array}{l}
H^{n+1} = H^n - \Delta t \nabla \cdot (\overline{\mathbf{u}}^{n+1}) + \Delta t F^n_H \\
\overline{\mathbf{u}}^n = \mathcal{G}(\beta, H^n)
\end{array}
\right.
\label{finite}
\end{equation}
where the superscript $n$ represents the current time step, $\overline{\mathbf{u}}$ is the depth-integrated velocity, $F_H$ is the accumulation rate, and $H$ is the ice thickness.\cite{HE2023112428}. The construction of the operator $\mathcal{G}$ is computationally intense, which takes the ice thickness $H^n$ and basal friction field $\beta$ as input, and is computed either based on the MOLHO approximation\cite{MOLHO} or SSA approximation\cite{SSA1}. Instead of computing the operator $\mathcal{G}$ through another finite element method, He et al. used the DeepONet network as an approximation, which is a physics-informed neural network\cite{Lu2021}. The hybrid ice flow model contains the DeepONet approximation of operator $\mathcal{G}$ and the first discretization equation of Equation \ref{finite}.

\begin{figure}[h]
  \centering
  \includegraphics[width=0.8\linewidth]{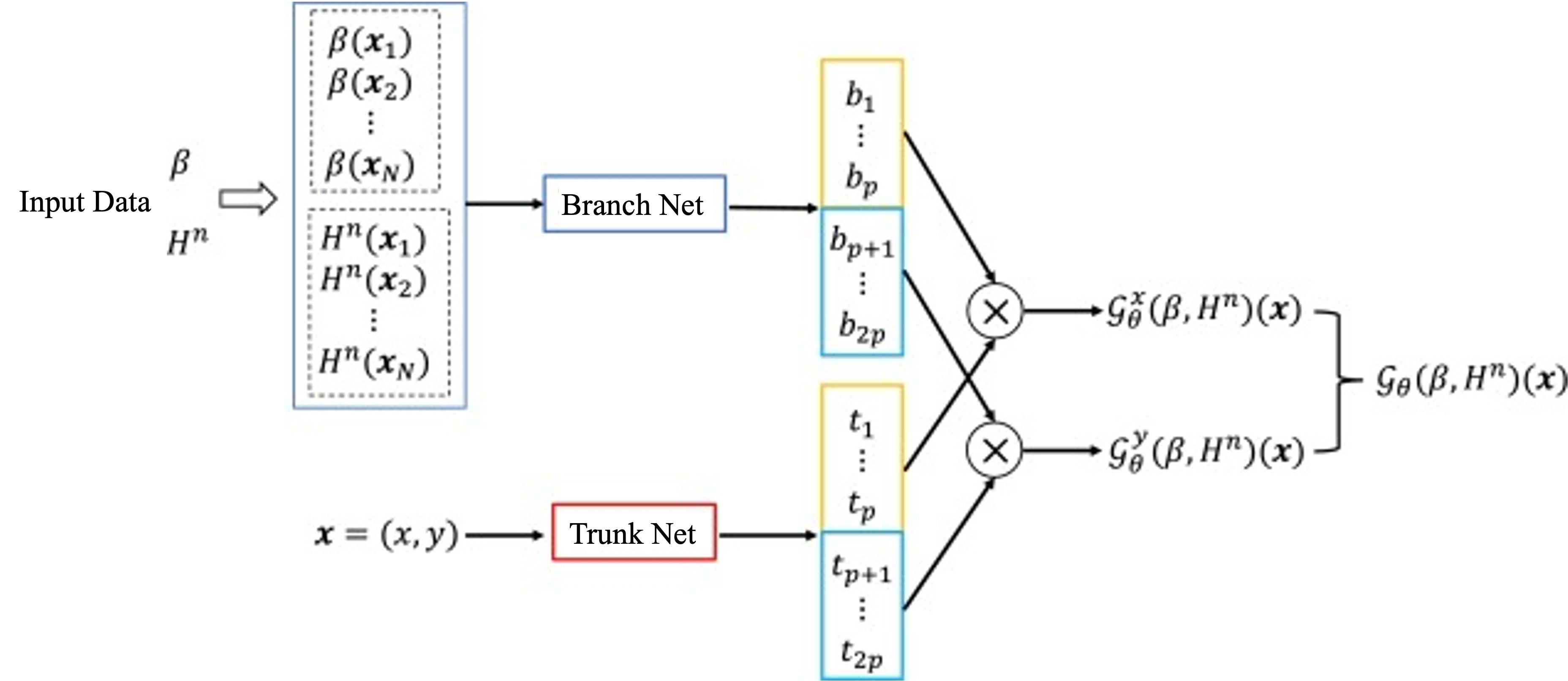}
  \caption{Schematic diagram of DeepONet used to approximate operator $\mathcal{G}$. Reproduced from original paper\cite{HE2023112428}.}
  \Description{This figure shows the schema of DeepONet to approximate the target operator, which includes a branch net and a trunk net.}
  \label{deeponet}
\end{figure}

The main idea of DeepONet is to learn operator mapping using a deep neural network\cite{Lu2021}. In Fig.\ref{deeponet}, DeepONet contains a branch net used to encode the sampled input function and return an embedding vector $\mathbf{b}$, and a trunk net used to evaluate the output function based on continuous input $\mathbf{x}$ and returns another vector $\mathbf{t}$. The value of $\mathcal{G}_{\theta}^x(\beta, H^n)(\mathbf{x})$ is calculated by the dot product of those two embedding vectors. The network will be trained by minimizing a loss function defined as a weighted mean square error. He et al. use a self-adaptive weight to bring the model with better generalization ability, which assumes the weight is a function of the space coordinates and can be optimized simultaneously with network parameters\cite{MCCLENNY2023111722selfadaptive,goswami2022physicsinformed}.
 
\begin{figure}[h]
  \centering
  \includegraphics[width=0.8\linewidth]{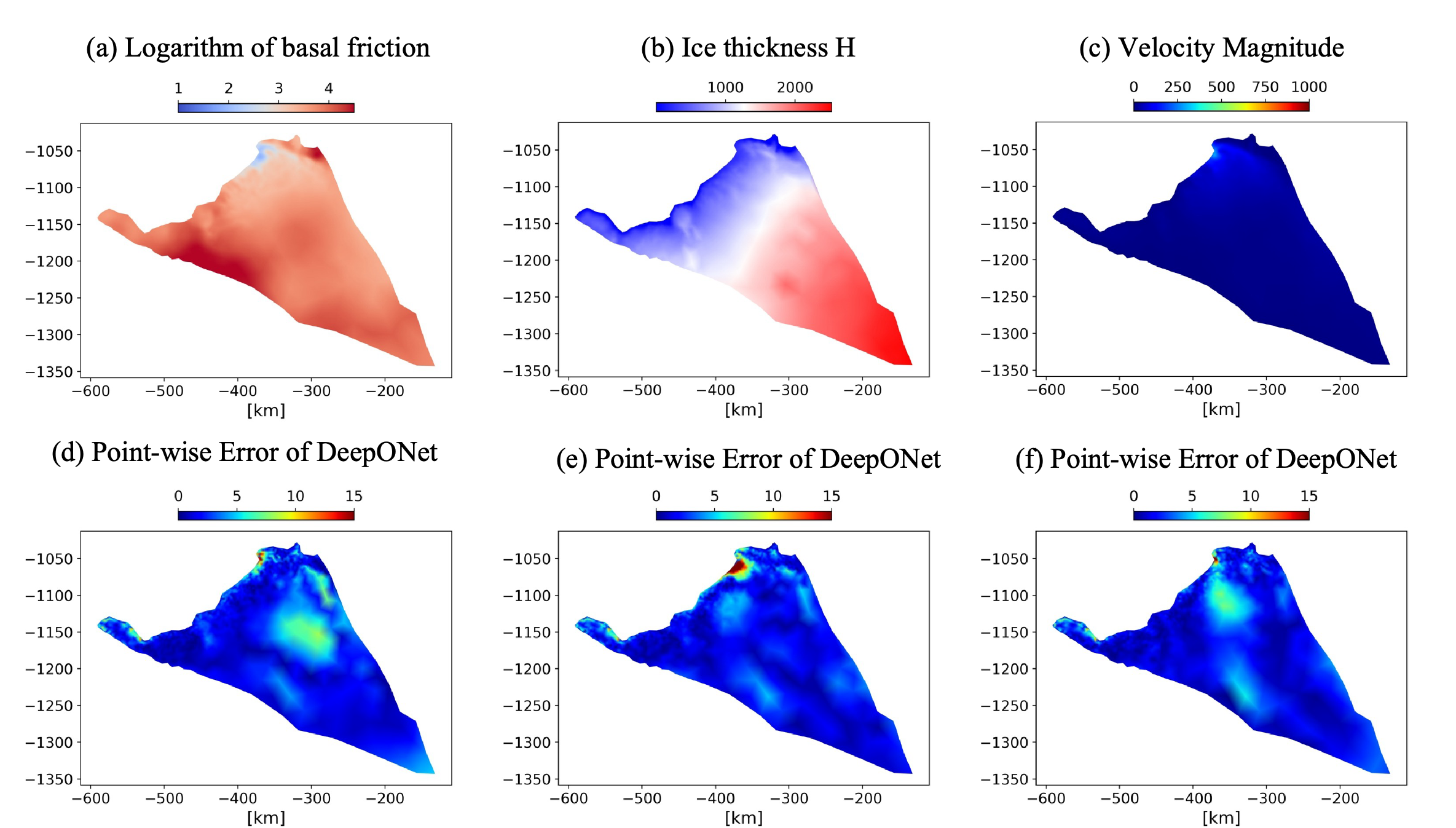}
  \caption{The result of DeepONet. (a) Logarithm of basal friction, (b) Ice thickness, (c) Magnitude of surface velocity, (d) Point-wise error for DeepONet, (e) Point-wise error for DeepONet with self-adaptive weight as $\lambda^2$, (f) Point-wise error for DeepONet with self-adaptive weight as $\lambda^4$. Reproduced from original paper\cite{HE2023112428}.}
  \Description{In this figure, (a) gives the result of DeepONet to predict the logarithm of basal friction. (b) gives the result of DeepONet to predict ice thickness. (c) shows the magnitude of velocity magnitude. (d) shows the point-wise error of DeepONet with no self-adaptive weights. (e) shows the point-wise error of DeepONet with self-adaptive weights as a square term. (f) shows the point-wise error of DeepONet with self-adaptive weights as a term with power 4.}
  \label{deeponetres}
\end{figure}

He et al. test the proposed model on the Humboldt Glacier with different settings of self-adaptive parameters. In Fig.\ref{deeponetres}, DeepONet can provide a precise prediction for the ice velocity field, while the use of self-adaptive weight can lower the prediction error in the interior and northwest region of the study domain\cite{HEDedge}. He et al. also show that the total computation time can be optimized by using DeepONet as a surrogate model for the non-linear operator. The entire computational time, including the time to allocate memory, initialization, data i/o, and model computation, can speed up five times using the DeepONet, reducing from 123.3 seconds per sample to 24.15. For solving the discretized equation Eq.\ref{finite}, the speed-up can reach 11 times\cite{HE2023112428}, reducing from 105.2 seconds per sample to 9.46. This result shows the great ability of the DeepONet surrogate model to avoid intense computation and provide a more accurate estimation for the ice velocity field.


\subsection{Combination through training strategy}
Some studies propose a physics-based training strategy on machine learning algorithms for polar ice. Typically, a deep neural network will have millions of parameters or more, and as a result, how these parameters are initialized and how the network is trained may directly affect the network's performance. In this section, we will do a case study on two relevant papers that include a unique physics-based training strategy.

\subsubsection{Deep Hybrid Wavelet Network for Ice Boundary Detection in Radar Imagery}
Kamangir et al.\cite{kamangir} combine wavelet transformation as the pre-processing stage with a neural network and propose a novel method to detect ice boundaries from radar image\cite{kamangir}. The development of deep learning techniques provides a possibility for building a more effective boundary detection method for ice layers from airborne radar images. Based on convolutional neural networks, Xie et al. proposed the Holistically-Nested Edge Detection(HED), which is a multiscale algorithm that takes side outputs as a kind of deep supervision\cite{HEDedge}. Inspired by HED, the authors show that adding wavelet transformation as a pre-processing step can improve the performance compared with other deep learning-based methods.

\begin{figure}[h]
  \centering
  \includegraphics[width=0.8\linewidth]{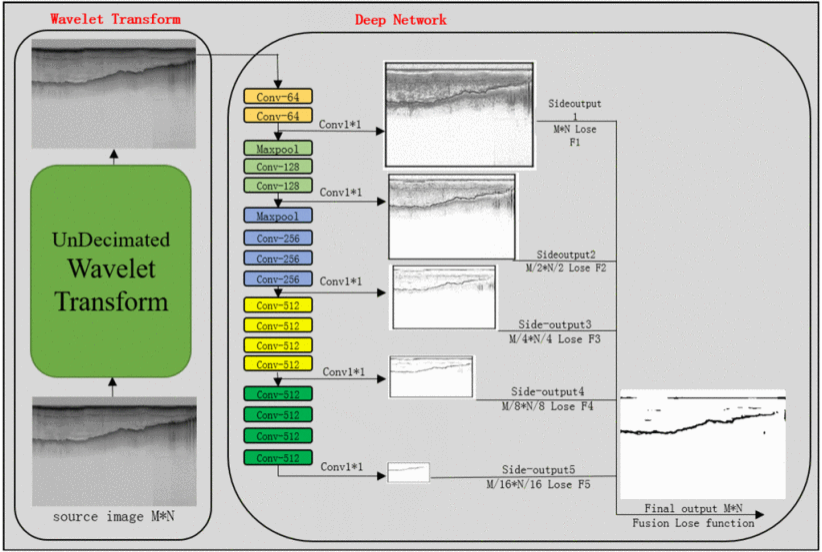}
  \caption{Network architecture for deep hybrid wavelet. Reproduced from original paper\cite{kamangir}.}
  \Description{The figure shows the network architecture of deep hybrid wavelet, where an UnDecimated Wavelet Transform is added before the deep convolutional network.}
  \label{hybridwave}
\end{figure}

Physics is incorporated by using a wavelet transformation as a preprocessing step to effectively reduce noise in radar data. Radar images suffer from heavy noise, which is the major obstacle to creating more precise boundary detection. Wavelet transformation is an efficient way to handle such noise, while the kinds of discrete wavelet transformation will cause the loss of coefficients during the transformation process. Therefore, as shown in Fig.\ref{hybridwave}, Kamangir et al. use an undecimated discrete wavelet transform as a preprocessing step to filter and denoise the raw radar image\cite{UDWT1,UDWT2}, and then use a HED-based MLP network to predict whether a pixel is a boundary point. To handle the imbalance that there are more non-edge pixels in the ground truth compared with edge pixels, Kamangir et al. implement a class-balanced cross-entropy loss function.

Kamangir et al. compared their proposed method with other traditional and deep learning-based boundary detection methods on the year 2009 images of NASA Operation IceBridge Mission Dataset. Their proposed Deep Hybrid Wavelet Network has the best Precision (0.801), Recall (0.742), and F-measure (0.71) compared with others, while a pre-trained HED network on BSDS500 serves as a benchmark and has an F-measure to be 0.73\cite{HEDbenchmark}.

\subsubsection{Learning Snow Layer Thickness Through Physics Defined Labels} 

Varshney et al.\cite{Varshneyphysicslabels} show that regression-based models pre-trained on labels simulated from a physics-based model can better learn the snow layer thickness on manual annotation of the radar images. The transfer learning process can let the network have better generalization ability to estimate englacial layer thickness. There are different methods to estimate snow layer thickness via deep learning methods. Besides building a network to do edge detection or segmentation on radar echograms, it is also possible to build a regression-based convolution neural network to predict the thickness values. Through regression-based networks, Varshney et al. show that the accuracy of layer thickness prediction is heavily affected by the lack of well-defined labels\cite{varshney_2021_regression_networks}. In order to handle this problem, physics is incorporated into the neural network by generating physics-simulated thickness and labels. Varshney et al. use the Modèle Atmosphérique Régional (MAR) physical model with a 15km-grid resolution to calculate the surface mass balance\cite{Varshneyphysicslabels} and use the Herron and Langway model to estimate the snow density in different depths\cite{herron_langway_1980}.

The authors conduct two experiments to test the effect of physics-simulated labels. First, they initialize the regression-based model with ImageNet weights and train it on manual annotation of radar echograms for 100 epochs. For the second experiment, they first train the same model for 100 epochs on physics-simulated labels from the MAR model, then use it as a pre-trained model and fine-tune it on manual annotation of radar images for another 100 epochs.

\begin{table}  
\caption{Comparison of mean absolute error achieved by different networks and different initialization methods. Results of DeepLabv3+-Poly are reported as a baseline\cite{varshney2020_deep_ice_layer}. Reproduced from the original paper\cite{Varshneyphysicslabels}.}
\centering

\scalebox{0.8}{
\begin{tabular}{ccccc}
\toprule
Network & \multicolumn{2}{c}{Training Set} & \multicolumn{2}{c}{Test Set} \\
\cmidrule{2-5}
 & ImageNet Initialization & MAR Initialization & ImageNet Initialization & MAR Initialization \\
\midrule
InceptionV3  & 1.622 & 0.990 & 3.431 & 2.834 \\
DenseNet  & 1.681 & 1.339 & 3.290 & 3.061 \\
Xception & 1.677 & 1.353 & 3.406 & 3.194 \\
MobileNetV2  & 1.451 & 1.3 & 4.435 & 3.122 \\
DeepLabv3+-Poly & 2.36 & NA & 3.590 & NA \\
\bottomrule
\end{tabular}}
\label{regresult}
\end{table}

The authors calculate the mean absolute error for two experiments, shown in Table \ref{regresult}. Results show that pre-training the network with physics-simulated labels based on the MAR model and then fine-tuning on manual annotations of radar echogram can achieve a lower mean absolute error. The reason is that by pre-training on physics simulated labels, the model can learn the gap between the simulated layers as prior knowledge, and when it comes to fine-tuning on manual annotations, the model can build a better initial estimation for layer thickness based on the prior knowledge. Moreover, training neural networks with prior physical knowledge can help the model achieve better generalization ability.

\subsubsection{Ice-flow model emulator based on physics-informed deep learning}
Jouvet et al. proposed a novel method to efficiently simulate the first-order approximation model of ice fluid dynamics on GPUs and achieve high-fidelity results. In ice models, ice fluid dynamic is governed by the 3D non-linear Glen-Stokes equations\cite{Greve_Blatter_Glen}. In order to reduce the intensive computation, the ice-flow equation is usually simplified by neglecting higher-order term to form the first-order approximation model (FOA) and then dropping the first-order term to form the SIA \cite{FOAmodel,Hutter1983}. GPUs can significantly improve the computation efficiency of large numeric models. However, for traditional solvers for ice fluid dynamics, like the SIA model, it is challenging to implement a parallelization version on GPU. 

Deep learning-based models also show great potential in reducing the high computational cost with GPU acceleration while keeping the same level of accuracy\cite{jouvet_cordonnier_kim_lüthi_vieli_aschwanden_2022,HE2023112428,brinkerhoff_aschwanden_fahnestock_2021}. Jouvet et al. proposed that the Glen-Stokes model can be emulated by a simple CNN, which is much faster and can obtain high-fidelity results\cite{jouvet_cordonnier_kim_lüthi_vieli_aschwanden_2022}. However, the dependency of the instructor model makes it difficult to train these models (i.e., an emulator) and limits the generalization ability.

Physics-informed neural networks (PINNs) have shown to be a powerful method for enforcing networks to learn from underlying physical laws. Basic PINNs are trained to minimize the residual of the underlying physical equation, while variational PINNs use the variational form of the problem as loss function\cite{basicpinn,variationalpinn}. Compared with basic PINNs, variational PINNs can avoid calculating high-order derivatives for residuals and have a connection with traditional numerical methods\cite{jouvet_cordonnier_2023}. Inspired by variational PINNs, the authors use a convolution neural network ice-flow emulator and design a novel training strategy. Experiment results show that their proposed method can remove the dependence of the instructor model and obtain a more generic emulator\cite{jouvet_cordonnier_2023}.

\begin{figure}[h]
  \centering
  \scalebox{0.9}{
  \includegraphics[width=0.8\linewidth]{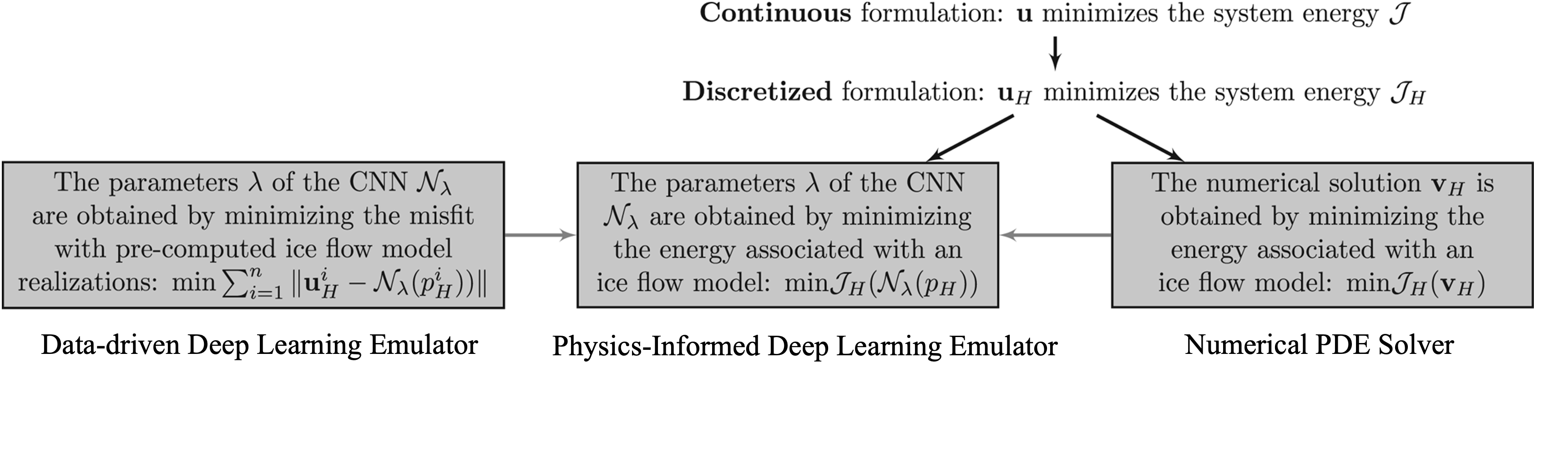}
  }
  \caption{The concept of physics-informed deep learning emulator on ice flow. It can be understood as the combination of a data-driven method and a traditional numerical method. Reproduced from original 
  paper\cite{jouvet_cordonnier_2023}.}
  \Description{The figure is a schema of the physics-informed deep learning approach on the ice flow emulator. The continuous formulation of the problem that aims to minimize the system energy can be discretized. Then, a traditional numerical PDE solver can directly achieve the solution. Data-driven deep learning emulator can simulate the ice flow. The desired physics-informed deep learning emulator can be seen as a combination of a numerical PDE solver and a data-driven deep learning emulator.}
  \label{emulator}
\end{figure}

Shown in Fig.\ref{emulator}, their proposed method can be viewed as a combination of data-driven method and traditional numerical method\cite{jouvet_cordonnier_2023}. Compared with traditional PINNs, the proposed method by Jouvet et al. uses model parameters instead of sampling point coordinates as the input. Instead of using the residual of the underlying physical equations as the loss function terms, Jouvet et al. adopt the Variational PINN training strategy and use the associated energy of the FOA model as the loss function\cite{jouvet_cordonnier_2023}.

Jouvet et al. show that the proposed PINN emulator can have high-fidelity results and be more computationally efficient. GPU acceleration can significantly improve the model's computational performance. The proposed method, which belongs to the deep Ritz method, does not require any data from an external model\cite{jouvet_cordonnier_2023}. The proposed emulator can enforce the ice-flow physical laws in training by minimizing the associated energy with the physical equations.

\begin{figure}[h]
  \centering
  \scalebox{0.9}{
  \includegraphics[width=0.8\linewidth]{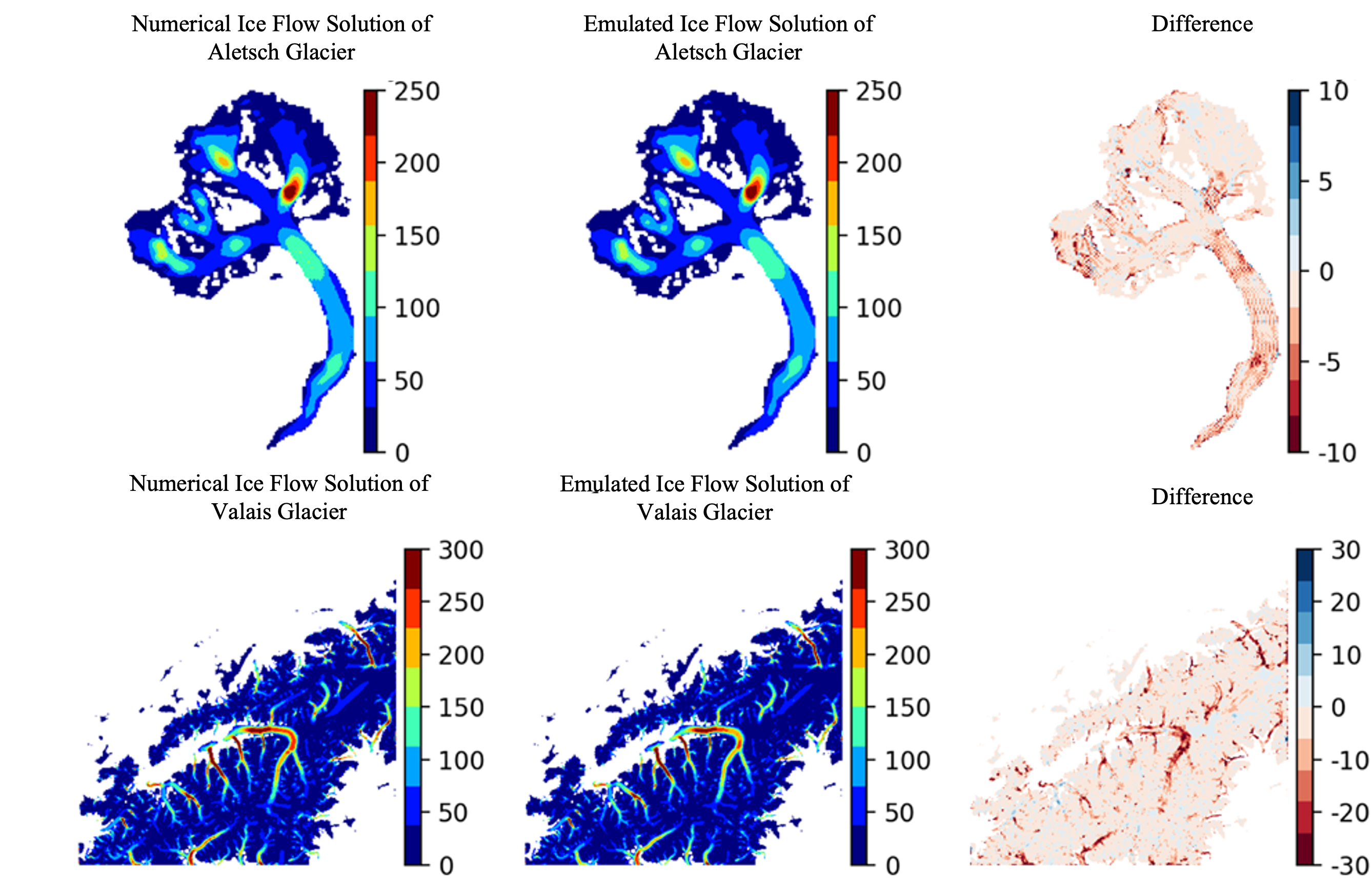}
  }
  \caption{Comparison between result from traditional solver and physics-informed emulator. Reproduced from original paper\cite{jouvet_cordonnier_2023}.}
  \Description{The figure shows a comparison between a traditional solver and a physics-informed emulator for generating the ice flow solution of Aletsch Glacier and Valais Glacier. Results show that the physics-informed emulator can provide high-accuracy results.}
  \label{emulatorresult}
\end{figure}

Jouvet et al. tested the model performance on snapshot solutions for Aletsch Glacier and Valais Glacier, shown in Fig.\ref{emulatorresult}. Calculating the difference between the physics-informed emulator and the traditional solver shows that the proposed emulator can get high-fidelity solutions, and the geometric features and thickness of the glacier can be accurately reproduced.

\section{Advantages of physics-informed machine learning on studying polar ice}\label{advantages}
Compared with pure physical or data-driven methods, the use of physics-informed machine learning on polar ice has a few advantages. In this section, we present a summary of how physics-informed machine learning can help researchers achieve better results in polar regions.

\subsection{Work with partially known models and fewer observational data}
The biggest advantage of physics-informed machine learning is its ability to work with partially known models and fewer observational data. Due to the extreme weather, studies built on on-site measurements from Antarctic stations only have a limited amount of data\cite{onsitestation}. Although there are reanalysis data that can provide global distribution for climate properties, like ice change or snow accumulation, the results still have high uncertainty in polar regions, especially in Antarctica\cite{reanalysis1,reanalysis2}. Additionally, even with the most advanced equipment and in the study region that is easy to access, noises still exist in the observational data. Moreover, the mechanisms of some physical processes are still debated and lack a widely accepted conclusion, like the sliding laws of the glacier basal drag. For this scenario, pure physical models or pure data-driven models may perform poorly, while physics-informed machine learning can combine all the available information together and find out the underlying pattern better. Combining the Bayesian method with physics-informed learning may also be helpful, especially for the task of uncertainty quantification\cite{Karniadakis2021}. 

\subsection{Provide better explanation and understanding of the whole algorithm}
Unlike those physical models where the intermediate result can also have physical interpretable meaning, pure data-driven machine learning algorithms are often seen as black box algorithms, where the inner mechanism of how the network can extract and learn features from the training data is unknown\cite{MakingtheBlackBoxMoreTransparentUnderstandingthePhysicalImplicationsofMachineLearning}. By incorporating machine learning with physical laws, this physical knowledge can provide beneficial insights into understanding the hidden mechanism of machine learning. Additionally, physics-informed machine learning can provide a better understanding of the whole framework of the proposed algorithm. In the application of polar ice, physics-informed machine learning may serve as a surrogate model for certain physical processes and provide an explanation about intermediate results, which can possibly be the pre-processed observational data or the residuals of physical equations. Some proposed algorithms can also be understood as multi-task learning that approximates the relationships between different ice properties and ensures the output is physically reasonable.

\subsection{Mesh-free and more efficient in computation}
In traditional numerical methods for PDE-related optimization, one of the most computationally intense steps is to set up a proper mesh grid and discretize the original equation. In physics-informed machine learning frameworks, its mesh-free approach makes it more efficient to solve PDE equations and easily handle problems of irregular boundaries. Additionally, while using physics-informed machine learning, instead of setting a specific solution domain, it is possible to generate on-demand solutions where the domain is chosen based on the task \cite{Cuomo2022}. For example, Teisberg et al. used physics-informed machine learning to generate ice thickness over certain regions. Under the framework of physics-informed machine learning, ice thickness can be considered as a function of spatial coordinates, and the result resolution of ice thickness can be decided later based on different needs, whereas the traditional interpolation method requires a pre-defined grid and a fixed result resolution. Lastly, the framework of physics-informed machine learning is scalable: it can be easily implemented using open-source packages (e.g., PyTorch, Tensorflow) and take advantage of GPU acceleration to speed up the calculation. However, traditional optimization methods, such as solvers for the shallow ice approximation model, cannot be easily implemented and paralleled on GPU.

\section{Challenges and Opportunity}\label{Comparision}
As we discuss in Section \ref{case}, the study of physics-informed machine learning on polar ice is only in the starting stage. Therefore, a few challenges and opportunities exist for new researchers who want to contribute to physics-informed machine learning methods on polar ice.

\subsection{Limitation and opportunity in research area}
One significant challenge is that current physics-informed machine learning research only focuses on land ice and rarely handles sea ice problems. Considering the strong seasonal trend of sea ice and its impact on global climate, it is scientifically crucial to build accurate models to predict the properties and evolution of sea ice. As discussed in Sections \ref{physics} and \ref{datadriven}, various physical models have been developed based on mass balance and momentum balance equations for sea ice, and network architectures have been actively applied to the sea ice regime. As physics-informed machine learning has shown noteworthy progress on land ice, integrating sea-ice-relevant physics and machine learning algorithms has a lot of potential for success. Even though sea ice has more complicated physics due to its floating movements, researchers may follow the path of how physics-informed machine learning is applied to land ice and then explore further opportunities to form unique methods for sea ice problems.

\subsection{Limitation and opportunity in the combination method of physics}
Most current physics-informed machine learning methods incorporate physics in the loss function, especially using the residual form of underlying physical equations. Although this straightforward and effective way does not require significant changes to network design, integrating physics in the loss function has a few potential problems. To integrate physics in the loss function, the physical laws need to be well-developed and have a commonly agreed form of equation. Typically, the physical equations should only have the network output as unknown variables, and all the needed derivatives should be calculated through automatic differentiation. Moreover, the weights for physical loss and data fitting loss need to be fine-tuned; different weights will affect the final results differently. Some weights may even cause problems as Iwasaki et al. \cite{Hardness} found that the PINN model has high prediction errors when the weight for data fitting loss is equal to the physical loss, and the model has an unwanted bimodal distribution of prediction error when the ratios of weights satisfy a certain condition. Their study shows the importance of choosing proper weights for physical loss terms and other loss terms. In general, adding physics to the loss function can be a good starting point when researchers want to introduce physical constraints to existing data-driven algorithms on sea ice. After proving that physical loss terms can produce better results, they can further consider other methods and form a hybrid method to combine physical laws with machine learning algorithms. Nevertheless, the idea of neural operator learning can also be applicable to modeling different physical processes and learning hidden physics laws. Physics-informed neural operator\cite{physics_informed_neural_operator}, especially the conservation law-encoded neural operator\cite{liu2023harnessing,liu2023ino}, can learn from data and ensure the relevant physical laws are satisfied. These studies can be extended to polar ice studies and provide better results for ice dynamics.

\subsection{Limitation and opportunity in network choice}
Current studies of physics-informed machine learning on polar ice have attempted only limited types of network architectures. Most studies used the MLP as their backbone architecture and incorporated physics in the loss function. This is a similar framework to the well-developed PINNs proposed by Raissi et al.\cite{PINN}. Although MLP works well in PINNs to approximate the known PDE functions, it may not be the best choice for other problems. In general, the capacity of a certain network depends on its complexity. Although MLP can have many hidden layers, it is generally considered to be a simpler neural network compared with more complex architectures, including CNN or RNN. Some researchers have already tried different frameworks, such as CNN \cite{Varshneyphysicslabels, kamangir, jouvet_cordonnier_2023} and DeepONet \cite{HE2023112428} and reached great success on the improvements of final predictions. As we discuss in Section \ref{datadriven}, numerous data-driven methods have chosen more deep and complex architectures to represent complex patterns or functions. Based on the current exploration of applying physics-informed machine learning to other network architectures besides MLP and CNN, it shows great potential that physical laws can be introduced to RNNs, GNNs, and GANs to enhance their results to be physically meaningful and consistent. 

\subsection{Limitation and opportunity in the choice of physics}
In current studies of physics-informed machine learning on polar ice, the choice of physics is limited. Table \ref{summary} shows that although the combination methods can be different, most land ice papers used known physical laws based on mass conservation and momentum conservation. These known physics can serve as regularization or assist in training strategy. However, although researchers have reached great success in providing physical explanations for different polar ice behaviors, there might be some physical laws that are totally unknown. In the future, the study  can be extended to the idea of neural inverse operators. Neural inverse operator\cite{molinaro_neural_inverse_operators,long2024invertible} is a kind of network architecture that can solve inverse-PDE problems and learn hidden physical law based on observational data. By applying these networks on polar ice, it is possible to better infer unknown physical quantities or reveal hidden physical processes.

\section{Conclusion and Future outlook}\label{conclusion}
In this paper, we review several physics and data-driven models for polar ice, including land and sea ice. Although these approaches produce reasonable results in monitoring land ice and sea ice, they have several limitations: the difficulty of representing every detail in the physical process when using physical models and the difficulty of collecting high-quality training datasets when using data-driven models. Physics-informed machine learning (PIML), a promising framework that combines physical laws with machine learning algorithms, can be an alternative choice to address the limitations of pure physics and data-driven methods. We discuss the current methodological development in integrating physics with machine learning algorithms. Then, we review and summarize recent progress on PIML for polar ice. We categorize them based on how physics is combined with data-driven approaches: combination through (1) loss function, (2) model architecture, and (3) training strategy. With the comprehensive comparison of PIML and other traditional methods for polar ice, we find that PIML has a few advantages: (1) its applicability to imperfect physical laws or fewer observational data (2) computational efficiency (3) totally mesh-free and (4) providing physics-meaningful explanations.

However, current studies on PIML on polar ice have a few challenges. In the future, researchers can consider the following aspects:
\begin{enumerate}
\item Apply the idea of PIML to sea ice problems. As we discussed, current studies are only on land ice. Although the mechanisms of land ice and sea ice can be quite different, similar mass balance and momentum balance equations can be applied to land ice and sea ice. As PIML has achieved success on land ice using the corresponding mass balance and momentum balance equations, it would be possible to extend similar ideas to sea ice problems.

\item Try different combination methods. In the future, besides combining physics with data-driven machine learning approaches in the loss functions, researchers may try more approaches on model architecture, training strategies, or forming some hybrid combination methods.

\item Try neural operator methods. Neural operator methods have shown great potential in modeling complex physical systems more accurately and efficiently, while physics-informed neural operator methods can improve the model's ability to satisfy underlying physical laws. In the future, researchers may try the idea of physics-informed neural operator on polar ice studies. 

\item Try different network architecture. Most PIML studies for polar ice use simple networks such as multi-layer perceptrons and convolutional neural networks. Leveraging recent developments in various data-driven network architectures, we suggest introducing physics to other advanced network architectures, such as graph neural networks, recurrent neural networks, and generative models. 

\item Use different physics. Most PIML studies for polar ice use known physical laws based on mass conservation and momentum conservation. In the future, researchers may consider the idea of inverse neural operator, where the network can learning hidden physics from observation data, better solve the PDE inverse problems and reveal unknown rules that modeling polar ice changes.

\end{enumerate}

We aim to provide a valuable summary for researchers who are new to this field or who want to fill the gap in the current study, as well as guide further research on physics-informed machine learning for polar ice study.

\begin{acks}
This work is supported by NSF BIGDATA (IIS-1838230, 2308649) and NSF Leadership Class Computing (OAC-2139536) awards.
\end{acks}

\bibliographystyle{ACM-Reference-Format}
\bibliography{Bib_Reference}


\begin{thebibliography}{147}


\ifx \showCODEN    \undefined \def \showCODEN     #1{\unskip}     \fi
\ifx \showDOI      \undefined \def \showDOI       #1{#1}\fi
\ifx \showISBNx    \undefined \def \showISBNx     #1{\unskip}     \fi
\ifx \showISBNxiii \undefined \def \showISBNxiii  #1{\unskip}     \fi
\ifx \showISSN     \undefined \def \showISSN      #1{\unskip}     \fi
\ifx \showLCCN     \undefined \def \showLCCN      #1{\unskip}     \fi
\ifx \shownote     \undefined \def \shownote      #1{#1}          \fi
\ifx \showarticletitle \undefined \def \showarticletitle #1{#1}   \fi
\ifx \showURL      \undefined \def \showURL       {\relax}        \fi
\providecommand\bibfield[2]{#2}
\providecommand\bibinfo[2]{#2}
\providecommand\natexlab[1]{#1}
\providecommand\showeprint[2][]{arXiv:#2}

\bibitem[Andersson et~al\mbox{.}(2021)]%
        {Andersson2021}
\bibfield{author}{\bibinfo{person}{Tom~R. Andersson}, \bibinfo{person}{J.~Scott Hosking}, \bibinfo{person}{María Pérez-Ortiz}, \bibinfo{person}{Brooks Paige}, \bibinfo{person}{Andrew Elliott}, \bibinfo{person}{Chris Russell}, \bibinfo{person}{Stephen Law}, \bibinfo{person}{Daniel~C. Jones}, \bibinfo{person}{Jeremy Wilkinson}, \bibinfo{person}{Tony Phillips}, \bibinfo{person}{James Byrne}, \bibinfo{person}{Steffen Tietsche}, \bibinfo{person}{Beena~Balan Sarojini}, \bibinfo{person}{Eduardo Blanchard-Wrigglesworth}, \bibinfo{person}{Yevgeny Aksenov}, \bibinfo{person}{Rod Downie}, {and} \bibinfo{person}{Emily Shuckburgh}.} \bibinfo{year}{2021}\natexlab{}.
\newblock \showarticletitle{Seasonal Arctic sea ice forecasting with probabilistic deep learning}.
\newblock \bibinfo{journal}{\emph{Nature Communications}} \bibinfo{volume}{12}, \bibinfo{number}{5124} (\bibinfo{year}{2021}).
\newblock


\bibitem[Argenti et~al\mbox{.}(2013)]%
        {UDWT1}
\bibfield{author}{\bibinfo{person}{Fabrizio Argenti}, \bibinfo{person}{Alessandro Lapini}, \bibinfo{person}{Tiziano Bianchi}, {and} \bibinfo{person}{Luciano Alparone}.} \bibinfo{year}{2013}\natexlab{}.
\newblock \showarticletitle{A Tutorial on Speckle Reduction in Synthetic Aperture Radar Images}.
\newblock \bibinfo{journal}{\emph{IEEE Geoscience and Remote Sensing Magazine}} \bibinfo{volume}{1}, \bibinfo{number}{3} (\bibinfo{year}{2013}), \bibinfo{pages}{6--35}.
\newblock
\urldef\tempurl%
\url{https://doi.org/10.1109/MGRS.2013.2277512}
\showDOI{\tempurl}


\bibitem[Baumhoer et~al\mbox{.}(2019)]%
        {Baumhoer2019}
\bibfield{author}{\bibinfo{person}{Celia~A. Baumhoer}, \bibinfo{person}{Andreas~J. Dietz}, \bibinfo{person}{C. Kneisel}, {and} \bibinfo{person}{C. Kuenzer}.} \bibinfo{year}{2019}\natexlab{}.
\newblock \showarticletitle{Automated Extraction of Antarctic Glacier and Ice Shelf Fronts from Sentinel-1 Imagery Using Deep Learning}.
\newblock \bibinfo{journal}{\emph{Remote Sensing}} \bibinfo{volume}{11}, \bibinfo{number}{21} (\bibinfo{year}{2019}).
\newblock
\showISSN{2072-4292}
\urldef\tempurl%
\url{https://doi.org/10.3390/rs11212529}
\showDOI{\tempurl}


\bibitem[Bell and Seroussi(2020)]%
        {Bell2020}
\bibfield{author}{\bibinfo{person}{Robin~E. Bell} {and} \bibinfo{person}{Helene Seroussi}.} \bibinfo{year}{2020}\natexlab{}.
\newblock \showarticletitle{History, mass loss, structure, and dynamic behavior of the Antarctic Ice Sheet}.
\newblock \bibinfo{journal}{\emph{Science}} \bibinfo{volume}{367}, \bibinfo{number}{6484} (\bibinfo{year}{2020}), \bibinfo{pages}{1321--1325}.
\newblock
\urldef\tempurl%
\url{https://doi.org/10.1126/science.aaz5489}
\showDOI{\tempurl}
\showeprint{https://www.science.org/doi/pdf/10.1126/science.aaz5489}


\bibitem[Blatter(1995a)]%
        {Blatter1995}
\bibfield{author}{\bibinfo{person}{Heinz Blatter}.} \bibinfo{year}{1995}\natexlab{a}.
\newblock \showarticletitle{Velocity and stress fields in grounded glaciers: a simple algorithm for including deviatoric stress gradients}.
\newblock \bibinfo{journal}{\emph{Journal of Glaciology}} \bibinfo{volume}{41}, \bibinfo{number}{138} (\bibinfo{year}{1995}), \bibinfo{pages}{333–344}.
\newblock
\urldef\tempurl%
\url{https://doi.org/10.3189/S002214300001621X}
\showDOI{\tempurl}


\bibitem[Blatter(1995b)]%
        {FOAmodel}
\bibfield{author}{\bibinfo{person}{Heinz Blatter}.} \bibinfo{year}{1995}\natexlab{b}.
\newblock \showarticletitle{Velocity and stress fields in grounded glaciers: a simple algorithm for including deviatoric stress gradients}.
\newblock \bibinfo{journal}{\emph{Journal of Glaciology}} \bibinfo{volume}{41}, \bibinfo{number}{138} (\bibinfo{year}{1995}), \bibinfo{pages}{333–344}.
\newblock
\urldef\tempurl%
\url{https://doi.org/10.3189/S002214300001621X}
\showDOI{\tempurl}


\bibitem[Bolibar et~al\mbox{.}(2020a)]%
        {Bolibar2020_reconstruction}
\bibfield{author}{\bibinfo{person}{J. Bolibar}, \bibinfo{person}{A. Rabatel}, \bibinfo{person}{I. Gouttevin}, {and} \bibinfo{person}{C. Galiez}.} \bibinfo{year}{2020}\natexlab{a}.
\newblock \showarticletitle{A deep learning reconstruction of mass balance series for all glaciers in the French Alps: 1967--2015}.
\newblock \bibinfo{journal}{\emph{Earth System Science Data}} \bibinfo{volume}{12}, \bibinfo{number}{3} (\bibinfo{year}{2020}), \bibinfo{pages}{1973--1983}.
\newblock
\urldef\tempurl%
\url{https://doi.org/10.5194/essd-12-1973-2020}
\showDOI{\tempurl}


\bibitem[Bolibar et~al\mbox{.}(2020b)]%
        {Bolibar2020}
\bibfield{author}{\bibinfo{person}{J. Bolibar}, \bibinfo{person}{A. Rabatel}, \bibinfo{person}{I. Gouttevin}, \bibinfo{person}{C. Galiez}, \bibinfo{person}{T. Condom}, {and} \bibinfo{person}{E. Sauquet}.} \bibinfo{year}{2020}\natexlab{b}.
\newblock \showarticletitle{Deep learning applied to glacier evolution modelling}.
\newblock \bibinfo{journal}{\emph{The Cryosphere}} \bibinfo{volume}{14}, \bibinfo{number}{2} (\bibinfo{year}{2020}), \bibinfo{pages}{565--584}.
\newblock
\urldef\tempurl%
\url{https://doi.org/10.5194/tc-14-565-2020}
\showDOI{\tempurl}


\bibitem[Brinkerhoff et~al\mbox{.}(2021)]%
        {brinkerhoff_aschwanden_fahnestock_2021}
\bibfield{author}{\bibinfo{person}{Douglas Brinkerhoff}, \bibinfo{person}{Andy Aschwanden}, {and} \bibinfo{person}{Mark Fahnestock}.} \bibinfo{year}{2021}\natexlab{}.
\newblock \showarticletitle{Constraining subglacial processes from surface velocity observations using surrogate-based Bayesian inference}.
\newblock \bibinfo{journal}{\emph{Journal of Glaciology}} \bibinfo{volume}{67}, \bibinfo{number}{263} (\bibinfo{year}{2021}), \bibinfo{pages}{385–403}.
\newblock
\urldef\tempurl%
\url{https://doi.org/10.1017/jog.2020.112}
\showDOI{\tempurl}


\bibitem[Cai et~al\mbox{.}(2023)]%
        {multibranch}
\bibfield{author}{\bibinfo{person}{Yiheng Cai}, \bibinfo{person}{Fuxing Wan}, \bibinfo{person}{Shinan Lang}, \bibinfo{person}{Xiangbin Cui}, {and} \bibinfo{person}{Zijun Yao}.} \bibinfo{year}{2023}\natexlab{}.
\newblock \showarticletitle{Multi-Branch Deep Neural Network for Bed Topography of Antarctica Super-Resolution: Reasonable Integration of Multiple Remote Sensing Data}.
\newblock \bibinfo{journal}{\emph{Remote Sensing}} \bibinfo{volume}{15}, \bibinfo{number}{5} (\bibinfo{year}{2023}).
\newblock
\showISSN{2072-4292}
\urldef\tempurl%
\url{https://doi.org/10.3390/rs15051359}
\showDOI{\tempurl}


\bibitem[Chen et~al\mbox{.}(2018)]%
        {Chen2018}
\bibfield{author}{\bibinfo{person}{Liang-Chieh Chen}, \bibinfo{person}{Yukun Zhu}, \bibinfo{person}{George Papandreou}, \bibinfo{person}{Florian Schroff}, {and} \bibinfo{person}{Hartwig Adam}.} \bibinfo{year}{2018}\natexlab{}.
\newblock \showarticletitle{Encoder-Decoder with Atrous Separable Convolution for Semantic Image Segmentation}. In \bibinfo{booktitle}{\emph{Computer Vision – ECCV 2018: 15th European Conference, Munich, Germany, September 8–14, 2018, Proceedings, Part VII}} (Munich, Germany). \bibinfo{publisher}{Springer-Verlag}, \bibinfo{address}{Berlin, Heidelberg}, \bibinfo{pages}{833–851}.
\newblock
\showISBNx{978-3-030-01233-5}
\urldef\tempurl%
\url{https://doi.org/10.1007/978-3-030-01234-2_49}
\showDOI{\tempurl}


\bibitem[Chi and Kim(2017)]%
        {Chi2017_LSTM}
\bibfield{author}{\bibinfo{person}{Junhwa Chi} {and} \bibinfo{person}{Hyun-choel Kim}.} \bibinfo{year}{2017}\natexlab{}.
\newblock \showarticletitle{Prediction of Arctic Sea Ice Concentration Using a Fully Data Driven Deep Neural Network}.
\newblock \bibinfo{journal}{\emph{Remote Sensing}} \bibinfo{volume}{9}, \bibinfo{number}{12} (\bibinfo{year}{2017}).
\newblock
\showISSN{2072-4292}
\urldef\tempurl%
\url{https://doi.org/10.3390/rs9121305}
\showDOI{\tempurl}


\bibitem[Choi et~al\mbox{.}(2019)]%
        {Choi2019_GRU}
\bibfield{author}{\bibinfo{person}{Minjoo Choi}, \bibinfo{person}{Liyanarachchi Waruna~Arampath De~Silva}, {and} \bibinfo{person}{Hajime Yamaguchi}.} \bibinfo{year}{2019}\natexlab{}.
\newblock \showarticletitle{Artificial Neural Network for the Short-Term Prediction of Arctic Sea Ice Concentration}.
\newblock \bibinfo{journal}{\emph{Remote Sensing}} \bibinfo{volume}{11}, \bibinfo{number}{9} (\bibinfo{year}{2019}).
\newblock
\showISSN{2072-4292}
\urldef\tempurl%
\url{https://doi.org/10.3390/rs11091071}
\showDOI{\tempurl}


\bibitem[Choi et~al\mbox{.}(2021)]%
        {Choi2021}
\bibfield{author}{\bibinfo{person}{Y. Choi}, \bibinfo{person}{M. Morlighem}, \bibinfo{person}{E. Rignot}, {and} \bibinfo{person}{Wood M.}} \bibinfo{year}{2021}\natexlab{}.
\newblock \showarticletitle{Ice dynamics will remain a primary driver of Greenland ice sheet mass loss over the next century}.
\newblock \bibinfo{journal}{\emph{Communications Earth \& Environment}} \bibinfo{volume}{2}, \bibinfo{number}{26} (\bibinfo{year}{2021}).
\newblock
\urldef\tempurl%
\url{https://doi.org/10.1038/s43247-021-00092-z}
\showDOI{\tempurl}


\bibitem[Clarke et~al\mbox{.}(2009)]%
        {Clarke_2009}
\bibfield{author}{\bibinfo{person}{Garry K.~C. Clarke}, \bibinfo{person}{Etienne Berthier}, \bibinfo{person}{Christian~G. Schoof}, {and} \bibinfo{person}{Alexander~H. Jarosch}.} \bibinfo{year}{2009}\natexlab{}.
\newblock \showarticletitle{Neural Networks Applied to Estimating Subglacial Topography and Glacier Volume}.
\newblock \bibinfo{journal}{\emph{Journal of Climate}} \bibinfo{volume}{22}, \bibinfo{number}{8} (\bibinfo{year}{2009}), \bibinfo{pages}{2146 -- 2160}.
\newblock
\urldef\tempurl%
\url{https://doi.org/10.1175/2008JCLI2572.1}
\showDOI{\tempurl}


\bibitem[Cox and Weeks(1974)]%
        {Sea_Ice_Salinity}
\bibfield{author}{\bibinfo{person}{G.~F.~N. Cox} {and} \bibinfo{person}{W.~F. Weeks}.} \bibinfo{year}{1974}\natexlab{}.
\newblock \showarticletitle{Salinity Variations in Sea Ice}.
\newblock \bibinfo{journal}{\emph{Journal of Glaciology}} \bibinfo{volume}{13}, \bibinfo{number}{67} (\bibinfo{year}{1974}), \bibinfo{pages}{109–120}.
\newblock
\urldef\tempurl%
\url{https://doi.org/10.3189/S0022143000023418}
\showDOI{\tempurl}


\bibitem[Cuomo et~al\mbox{.}(2022)]%
        {Cuomo2022}
\bibfield{author}{\bibinfo{person}{Salvatore Cuomo}, \bibinfo{person}{Vincenzo~Schiano Di~Cola}, \bibinfo{person}{Fabio Giampaolo}, \bibinfo{person}{Gianluigi Rozza}, \bibinfo{person}{Maziar Raissi}, {and} \bibinfo{person}{Francesco Piccialli}.} \bibinfo{year}{2022}\natexlab{}.
\newblock \showarticletitle{Scientific Machine Learning Through Physics--Informed Neural Networks: Where we are and What's Next}.
\newblock \bibinfo{journal}{\emph{Journal of Scientific Computing}} \bibinfo{volume}{92}, \bibinfo{number}{3} (\bibinfo{date}{26 Jul} \bibinfo{year}{2022}), \bibinfo{pages}{88}.
\newblock
\showISSN{1573-7691}
\urldef\tempurl%
\url{https://doi.org/10.1007/s10915-022-01939-z}
\showDOI{\tempurl}


\bibitem[Curry et~al\mbox{.}(1995)]%
        {Judith1995_albedo_feedback}
\bibfield{author}{\bibinfo{person}{Judith~A. Curry}, \bibinfo{person}{Julie~L. Schramm}, {and} \bibinfo{person}{Elizabeth~E. Ebert}.} \bibinfo{year}{1995}\natexlab{}.
\newblock \showarticletitle{Sea Ice-Albedo Climate Feedback Mechanism}.
\newblock \bibinfo{journal}{\emph{Journal of Climate}} \bibinfo{volume}{8}, \bibinfo{number}{2} (\bibinfo{year}{1995}), \bibinfo{pages}{240 -- 247}.
\newblock
\urldef\tempurl%
\url{https://doi.org/10.1175/1520-0442(1995)008<0240:SIACFM>2.0.CO;2}
\showDOI{\tempurl}


\bibitem[Desai and Strachan(2021)]%
        {Desai2021}
\bibfield{author}{\bibinfo{person}{Saaketh Desai} {and} \bibinfo{person}{Alejandro Strachan}.} \bibinfo{year}{2021}\natexlab{}.
\newblock \showarticletitle{Parsimonious neural networks learn interpretable physical laws}.
\newblock \bibinfo{journal}{\emph{Scientific Reports}} \bibinfo{volume}{11}, \bibinfo{number}{1} (\bibinfo{date}{17 Jun} \bibinfo{year}{2021}), \bibinfo{pages}{12761}.
\newblock
\showISSN{2045-2322}
\urldef\tempurl%
\url{https://doi.org/10.1038/s41598-021-92278-w}
\showDOI{\tempurl}


\bibitem[Dias~dos Santos et~al\mbox{.}(2022)]%
        {MOLHO}
\bibfield{author}{\bibinfo{person}{T. Dias~dos Santos}, \bibinfo{person}{M. Morlighem}, {and} \bibinfo{person}{D. Brinkerhoff}.} \bibinfo{year}{2022}\natexlab{}.
\newblock \showarticletitle{A new vertically integrated MOno-Layer Higher-Order (MOLHO) ice flow model}.
\newblock \bibinfo{journal}{\emph{The Cryosphere}} \bibinfo{volume}{16}, \bibinfo{number}{1} (\bibinfo{year}{2022}), \bibinfo{pages}{179--195}.
\newblock
\urldef\tempurl%
\url{https://doi.org/10.5194/tc-16-179-2022}
\showDOI{\tempurl}


\bibitem[Diebold et~al\mbox{.}(2023)]%
        {Francis2023}
\bibfield{author}{\bibinfo{person}{Francis~X. Diebold}, \bibinfo{person}{Glenn~D. Rudebusch}, \bibinfo{person}{Maximilian Göbel}, \bibinfo{person}{Philippe {Goulet Coulombe}}, {and} \bibinfo{person}{Boyuan Zhang}.} \bibinfo{year}{2023}\natexlab{}.
\newblock \showarticletitle{When will Arctic sea ice disappear? Projections of area, extent, thickness, and volume}.
\newblock \bibinfo{journal}{\emph{Journal of Econometrics}} \bibinfo{volume}{236}, \bibinfo{number}{2} (\bibinfo{year}{2023}), \bibinfo{pages}{105479}.
\newblock
\showISSN{0304-4076}
\urldef\tempurl%
\url{https://doi.org/10.1016/j.jeconom.2023.105479}
\showDOI{\tempurl}


\bibitem[Dong et~al\mbox{.}(2022)]%
        {Dong2022}
\bibfield{author}{\bibinfo{person}{Sheng Dong}, \bibinfo{person}{Xueyuan Tang}, \bibinfo{person}{Jingxue Guo}, \bibinfo{person}{Lei Fu}, \bibinfo{person}{Xiaofei Chen}, {and} \bibinfo{person}{Bo Sun}.} \bibinfo{year}{2022}\natexlab{}.
\newblock \showarticletitle{EisNet: Extracting Bedrock and Internal Layers From Radiostratigraphy of Ice Sheets With Machine Learning}.
\newblock \bibinfo{journal}{\emph{IEEE Transactions on Geoscience and Remote Sensing}}  \bibinfo{volume}{60} (\bibinfo{year}{2022}), \bibinfo{pages}{1--12}.
\newblock
\urldef\tempurl%
\url{https://doi.org/10.1109/TGRS.2021.3136648}
\showDOI{\tempurl}


\bibitem[Eayrs et~al\mbox{.}(2021)]%
        {eayrs2021}
\bibfield{author}{\bibinfo{person}{Clare Eayrs}, \bibinfo{person}{Xichen Li}, \bibinfo{person}{Marilyn~N Raphael}, {and} \bibinfo{person}{David~M Holland}.} \bibinfo{year}{2021}\natexlab{}.
\newblock \showarticletitle{Rapid decline in Antarctic sea ice in recent years hints at future change}.
\newblock \bibinfo{journal}{\emph{Nature Geoscience}} \bibinfo{volume}{14}, \bibinfo{number}{7} (\bibinfo{year}{2021}), \bibinfo{pages}{460--464}.
\newblock


\bibitem[Faroughi et~al\mbox{.}(2024)]%
        {survey_subdomain_fluid}
\bibfield{author}{\bibinfo{person}{Salah~A. Faroughi}, \bibinfo{person}{Nikhil~M. Pawar}, \bibinfo{person}{Célio Fernandes}, \bibinfo{person}{Maziar Raissi}, \bibinfo{person}{Subasish Das}, \bibinfo{person}{Nima~K. Kalantari}, {and} \bibinfo{person}{Seyed~Kourosh Mahjour}.} \bibinfo{year}{2024}\natexlab{}.
\newblock \showarticletitle{{Physics-Guided, Physics-Informed, and Physics-Encoded Neural Networks and Operators in Scientific Computing: Fluid and Solid Mechanics}}.
\newblock \bibinfo{journal}{\emph{Journal of Computing and Information Science in Engineering}} (\bibinfo{date}{01} \bibinfo{year}{2024}), \bibinfo{pages}{1--45}.
\newblock
\showISSN{1530-9827}
\urldef\tempurl%
\url{https://doi.org/10.1115/1.4064449}
\showDOI{\tempurl}
\showeprint{https://asmedigitalcollection.asme.org/computingengineering/article-pdf/doi/10.1115/1.4064449/7227573/jcise-23-1207.pdf}


\bibitem[Fathi et~al\mbox{.}(2020)]%
        {FATHI2020105729}
\bibfield{author}{\bibinfo{person}{Mojtaba~F. Fathi}, \bibinfo{person}{Isaac Perez-Raya}, \bibinfo{person}{Ahmadreza Baghaie}, \bibinfo{person}{Philipp Berg}, \bibinfo{person}{Gabor Janiga}, \bibinfo{person}{Amirhossein Arzani}, {and} \bibinfo{person}{Roshan~M. D’Souza}.} \bibinfo{year}{2020}\natexlab{}.
\newblock \showarticletitle{Super-resolution and denoising of 4D-Flow MRI using physics-Informed deep neural nets}.
\newblock \bibinfo{journal}{\emph{Computer Methods and Programs in Biomedicine}}  \bibinfo{volume}{197} (\bibinfo{year}{2020}), \bibinfo{pages}{105729}.
\newblock
\showISSN{0169-2607}
\urldef\tempurl%
\url{https://doi.org/10.1016/j.cmpb.2020.105729}
\showDOI{\tempurl}


\bibitem[Feng et~al\mbox{.}(2023)]%
        {Feng2023_ConvLSTM}
\bibfield{author}{\bibinfo{person}{Juanjuan Feng}, \bibinfo{person}{Jia Li}, \bibinfo{person}{Wenjie Zhong}, \bibinfo{person}{Junhui Wu}, \bibinfo{person}{Zhiqiang Li}, \bibinfo{person}{Lingshuai Kong}, {and} \bibinfo{person}{Lei Guo}.} \bibinfo{year}{2023}\natexlab{}.
\newblock \showarticletitle{Daily-Scale Prediction of Arctic Sea Ice Concentration Based on Recurrent Neural Network Models}.
\newblock \bibinfo{journal}{\emph{Journal of Marine Science and Engineering}} \bibinfo{volume}{11}, \bibinfo{number}{12} (\bibinfo{year}{2023}).
\newblock
\showISSN{2077-1312}
\urldef\tempurl%
\url{https://doi.org/10.3390/jmse11122319}
\showDOI{\tempurl}


\bibitem[Forsberg et~al\mbox{.}(2017)]%
        {Forsberg2017}
\bibfield{author}{\bibinfo{person}{Rene Forsberg}, \bibinfo{person}{Louise S{\o}rensen}, {and} \bibinfo{person}{Sebastian Simonsen}.} \bibinfo{year}{2017}\natexlab{}.
\newblock \bibinfo{booktitle}{\emph{Greenland and Antarctica Ice Sheet Mass Changes and Effects on Global Sea Level}}.
\newblock \bibinfo{publisher}{Springer International Publishing}, \bibinfo{address}{Cham}, \bibinfo{pages}{91--106}.
\newblock
\showISBNx{978-3-319-56490-6}
\urldef\tempurl%
\url{https://doi.org/10.1007/978-3-319-56490-6_5}
\showDOI{\tempurl}


\bibitem[Fretwell et~al\mbox{.}(2013)]%
        {Bedmap2}
\bibfield{author}{\bibinfo{person}{P. Fretwell}, \bibinfo{person}{H.~D. Pritchard}, \bibinfo{person}{D.~G. Vaughan}, \bibinfo{person}{J.~L. Bamber}, \bibinfo{person}{N.~E. Barrand}, \bibinfo{person}{R. Bell}, \bibinfo{person}{C. Bianchi}, \bibinfo{person}{R.~G. Bingham}, \bibinfo{person}{D.~D. Blankenship}, \bibinfo{person}{G. Casassa}, {and} \bibinfo{person}{et al.}} \bibinfo{year}{2013}\natexlab{}.
\newblock \showarticletitle{Bedmap2: improved ice bed, surface and thickness datasets for Antarctica}.
\newblock \bibinfo{journal}{\emph{The Cryosphere}} \bibinfo{volume}{7}, \bibinfo{number}{1} (\bibinfo{year}{2013}), \bibinfo{pages}{375--393}.
\newblock
\urldef\tempurl%
\url{https://doi.org/10.5194/tc-7-375-2013}
\showDOI{\tempurl}


\bibitem[Fritzner et~al\mbox{.}(2020)]%
        {Fritzner2020}
\bibfield{author}{\bibinfo{person}{Sindre Fritzner}, \bibinfo{person}{Rune Graversen}, {and} \bibinfo{person}{Kai~H. Christensen}.} \bibinfo{year}{2020}\natexlab{}.
\newblock \showarticletitle{Assessment of High-Resolution Dynamical and Machine Learning Models for Prediction of Sea Ice Concentration in a Regional Application}.
\newblock \bibinfo{journal}{\emph{Journal of Geophysical Research: Oceans}} \bibinfo{volume}{125}, \bibinfo{number}{11} (\bibinfo{year}{2020}), \bibinfo{pages}{e2020JC016277}.
\newblock
\urldef\tempurl%
\url{https://doi.org/10.1029/2020JC016277}
\showDOI{\tempurl}
\showeprint{https://agupubs.onlinelibrary.wiley.com/doi/pdf/10.1029/2020JC016277}
\newblock
\shownote{e2020JC016277 10.1029/2020JC016277}.


\bibitem[F{\"u}rst et~al\mbox{.}(2016)]%
        {control1}
\bibfield{author}{\bibinfo{person}{Johannes~Jakob F{\"u}rst}, \bibinfo{person}{Ga{\"e}l Durand}, \bibinfo{person}{Fabien Gillet-Chaulet}, \bibinfo{person}{Laure Tavard}, \bibinfo{person}{Melanie Rankl}, \bibinfo{person}{Matthias Braun}, {and} \bibinfo{person}{Olivier Gagliardini}.} \bibinfo{year}{2016}\natexlab{}.
\newblock \showarticletitle{The safety band of Antarctic ice shelves}.
\newblock \bibinfo{journal}{\emph{Nature Climate Change}} \bibinfo{volume}{6}, \bibinfo{number}{5} (\bibinfo{date}{01 May} \bibinfo{year}{2016}), \bibinfo{pages}{479--482}.
\newblock
\showISSN{1758-6798}
\urldef\tempurl%
\url{https://doi.org/10.1038/nclimate2912}
\showDOI{\tempurl}


\bibitem[Glen and Perutz(1955)]%
        {Glen1955}
\bibfield{author}{\bibinfo{person}{J.~W. Glen} {and} \bibinfo{person}{Max~Ferdinand Perutz}.} \bibinfo{year}{1955}\natexlab{}.
\newblock \showarticletitle{The creep of polycrystalline ice}.
\newblock \bibinfo{journal}{\emph{Proceedings of the Royal Society of London. Series A. Mathematical and Physical Sciences}} \bibinfo{volume}{228}, \bibinfo{number}{1175} (\bibinfo{year}{1955}), \bibinfo{pages}{519--538}.
\newblock
\urldef\tempurl%
\url{https://doi.org/10.1098/rspa.1955.0066}
\showDOI{\tempurl}
\showeprint{https://royalsocietypublishing.org/doi/pdf/10.1098/rspa.1955.0066}


\bibitem[Goswami et~al\mbox{.}(2022)]%
        {goswami2022physicsinformed}
\bibfield{author}{\bibinfo{person}{Somdatta Goswami}, \bibinfo{person}{Aniruddha Bora}, \bibinfo{person}{Yue Yu}, {and} \bibinfo{person}{George~Em Karniadakis}.} \bibinfo{year}{2022}\natexlab{}.
\newblock \bibinfo{title}{Physics-Informed Deep Neural Operator Networks}.
\newblock
\newblock
\showeprint[arxiv]{2207.05748}~[cs.LG]


\bibitem[Greve and Blatter(2009)]%
        {Greve_Blatter_Glen}
\bibfield{author}{\bibinfo{person}{Ralf Greve} {and} \bibinfo{person}{Heinz Blatter}.} \bibinfo{year}{2009}\natexlab{}.
\newblock \bibinfo{booktitle}{\emph{Dynamics of Ice Sheets and Glaciers}}.
\newblock \bibinfo{publisher}{Springer}.
\newblock
\urldef\tempurl%
\url{https://doi.org/10.1007/978-3-642-03415-2}
\showDOI{\tempurl}


\bibitem[Grigoryev et~al\mbox{.}(2022)]%
        {Grigoryev2022}
\bibfield{author}{\bibinfo{person}{Timofey Grigoryev}, \bibinfo{person}{Polina Verezemskaya}, \bibinfo{person}{Mikhail Krinitskiy}, \bibinfo{person}{Nikita Anikin}, \bibinfo{person}{Alexander Gavrikov}, \bibinfo{person}{Ilya Trofimov}, \bibinfo{person}{Nikita Balabin}, \bibinfo{person}{Aleksei Shpilman}, \bibinfo{person}{Andrei Eremchenko}, \bibinfo{person}{Sergey Gulev}, \bibinfo{person}{Evgeny Burnaev}, {and} \bibinfo{person}{Vladimir Vanovskiy}.} \bibinfo{year}{2022}\natexlab{}.
\newblock \showarticletitle{Data-Driven Short-Term Daily Operational Sea Ice Regional Forecasting}.
\newblock \bibinfo{journal}{\emph{Remote Sensing}} \bibinfo{volume}{14}, \bibinfo{number}{22} (\bibinfo{year}{2022}).
\newblock
\showISSN{2072-4292}
\urldef\tempurl%
\url{https://doi.org/10.3390/rs14225837}
\showDOI{\tempurl}


\bibitem[He et~al\mbox{.}(2023)]%
        {HE2023112428}
\bibfield{author}{\bibinfo{person}{QiZhi He}, \bibinfo{person}{Mauro Perego}, \bibinfo{person}{Amanda~A. Howard}, \bibinfo{person}{George~Em Karniadakis}, {and} \bibinfo{person}{Panos Stinis}.} \bibinfo{year}{2023}\natexlab{}.
\newblock \showarticletitle{A hybrid deep neural operator/finite element method for ice-sheet modeling}.
\newblock \bibinfo{journal}{\emph{J. Comput. Phys.}}  \bibinfo{volume}{492} (\bibinfo{year}{2023}), \bibinfo{pages}{112428}.
\newblock
\showISSN{0021-9991}
\urldef\tempurl%
\url{https://doi.org/10.1016/j.jcp.2023.112428}
\showDOI{\tempurl}


\bibitem[Herron and Langway(1980)]%
        {herron_langway_1980}
\bibfield{author}{\bibinfo{person}{Michael~M. Herron} {and} \bibinfo{person}{Chester~C. Langway}.} \bibinfo{year}{1980}\natexlab{}.
\newblock \showarticletitle{Firn Densification: An Empirical Model}.
\newblock \bibinfo{journal}{\emph{Journal of Glaciology}} \bibinfo{volume}{25}, \bibinfo{number}{93} (\bibinfo{year}{1980}), \bibinfo{pages}{373–385}.
\newblock
\urldef\tempurl%
\url{https://doi.org/10.3189/S0022143000015239}
\showDOI{\tempurl}


\bibitem[Hersbach et~al\mbox{.}(2020)]%
        {reanalysis1}
\bibfield{author}{\bibinfo{person}{Hans Hersbach}, \bibinfo{person}{Bill Bell}, \bibinfo{person}{Paul Berrisford}, \bibinfo{person}{Shoji Hirahara}, \bibinfo{person}{András Horányi}, \bibinfo{person}{Joaquín Muñoz-Sabater}, \bibinfo{person}{Julien Nicolas}, \bibinfo{person}{Carole Peubey}, \bibinfo{person}{Raluca Radu}, \bibinfo{person}{Dinand Schepers}, {and} \bibinfo{person}{et al.}} \bibinfo{year}{2020}\natexlab{}.
\newblock \showarticletitle{The ERA5 global reanalysis}.
\newblock \bibinfo{journal}{\emph{Quarterly Journal of the Royal Meteorological Society}} \bibinfo{volume}{146}, \bibinfo{number}{730} (\bibinfo{year}{2020}), \bibinfo{pages}{1999--2049}.
\newblock
\urldef\tempurl%
\url{https://doi.org/10.1002/qj.3803}
\showDOI{\tempurl}
\showeprint{https://rmets.onlinelibrary.wiley.com/doi/pdf/10.1002/qj.3803}


\bibitem[Hibler(1979)]%
        {Hibler1979}
\bibfield{author}{\bibinfo{person}{W.~D. Hibler}.} \bibinfo{year}{1979}\natexlab{}.
\newblock \showarticletitle{A Dynamic Thermodynamic Sea Ice Model}.
\newblock \bibinfo{journal}{\emph{Journal of Physical Oceanography}} \bibinfo{volume}{9}, \bibinfo{number}{4} (\bibinfo{year}{1979}), \bibinfo{pages}{815 -- 846}.
\newblock
\urldef\tempurl%
\url{https://doi.org/10.1175/1520-0485(1979)009<0815:ADTSIM>2.0.CO;2}
\showDOI{\tempurl}


\bibitem[Hindmarsh(2004)]%
        {Hindmarsh2004}
\bibfield{author}{\bibinfo{person}{R.~C.~A. Hindmarsh}.} \bibinfo{year}{2004}\natexlab{}.
\newblock \showarticletitle{A numerical comparison of approximations to the Stokes equations used in ice sheet and glacier modeling}.
\newblock \bibinfo{journal}{\emph{Journal of Geophysical Research: Earth Surface}} \bibinfo{volume}{109}, \bibinfo{number}{F1} (\bibinfo{year}{2004}).
\newblock
\urldef\tempurl%
\url{https://doi.org/10.1029/2003JF000065}
\showDOI{\tempurl}
\showeprint{https://agupubs.onlinelibrary.wiley.com/doi/pdf/10.1029/2003JF000065}


\bibitem[Holland and Kwok(2012)]%
        {Holland2012}
\bibfield{author}{\bibinfo{person}{Paul.~R. Holland} {and} \bibinfo{person}{Ron Kwok}.} \bibinfo{year}{2012}\natexlab{}.
\newblock \showarticletitle{Wind-driven trends in Antarctic sea-ice drift}.
\newblock \bibinfo{journal}{\emph{Nature Geoscience}}  \bibinfo{volume}{5} (\bibinfo{year}{2012}), \bibinfo{pages}{872--875}.
\newblock
\urldef\tempurl%
\url{https://doi.org/10.1038/ngeo1627}
\showDOI{\tempurl}


\bibitem[Hu et~al\mbox{.}(2021)]%
        {Hu2021_surface_melting}
\bibfield{author}{\bibinfo{person}{Z. Hu}, \bibinfo{person}{P. Kuipers~Munneke}, \bibinfo{person}{S. Lhermitte}, \bibinfo{person}{M. Izeboud}, {and} \bibinfo{person}{M. van~den Broeke}.} \bibinfo{year}{2021}\natexlab{}.
\newblock \showarticletitle{Improving surface melt estimation over the Antarctic Ice Sheet using deep learning: a proof of concept over the Larsen Ice Shelf}.
\newblock \bibinfo{journal}{\emph{The Cryosphere}} \bibinfo{volume}{15}, \bibinfo{number}{12} (\bibinfo{year}{2021}), \bibinfo{pages}{5639--5658}.
\newblock
\urldef\tempurl%
\url{https://doi.org/10.5194/tc-15-5639-2021}
\showDOI{\tempurl}


\bibitem[Huang et~al\mbox{.}(2017)]%
        {waveletcnn}
\bibfield{author}{\bibinfo{person}{Huaibo Huang}, \bibinfo{person}{Ran He}, \bibinfo{person}{Zhenan Sun}, {and} \bibinfo{person}{Tieniu Tan}.} \bibinfo{year}{2017}\natexlab{}.
\newblock \showarticletitle{Wavelet-SRNet: A Wavelet-Based CNN for Multi-scale Face Super Resolution}. In \bibinfo{booktitle}{\emph{2017 IEEE International Conference on Computer Vision (ICCV)}}. \bibinfo{pages}{1698--1706}.
\newblock
\urldef\tempurl%
\url{https://doi.org/10.1109/ICCV.2017.187}
\showDOI{\tempurl}


\bibitem[Huang et~al\mbox{.}(2021)]%
        {Huang2021_GNN}
\bibfield{author}{\bibinfo{person}{Yiyi Huang}, \bibinfo{person}{MatthÃ¤us Kleindessner}, \bibinfo{person}{Alexey Munishkin}, \bibinfo{person}{Debvrat Varshney}, \bibinfo{person}{Pei Guo}, {and} \bibinfo{person}{Jianwu Wang}.} \bibinfo{year}{2021}\natexlab{}.
\newblock \showarticletitle{Benchmarking of Data-Driven Causality Discovery Approaches in the Interactions of Arctic Sea Ice and Atmosphere}.
\newblock \bibinfo{journal}{\emph{Frontiers in Big Data}}  \bibinfo{volume}{4} (\bibinfo{year}{2021}).
\newblock
\showISSN{2624-909X}
\urldef\tempurl%
\url{https://doi.org/10.3389/fdata.2021.642182}
\showDOI{\tempurl}


\bibitem[Hutter(1983)]%
        {Hutter1983}
\bibfield{author}{\bibinfo{person}{K. Hutter}.} \bibinfo{year}{1983}\natexlab{}.
\newblock \bibinfo{booktitle}{\emph{Theoretical Glaciology: Material Science of Ice and the Mechanics of Glaciers and Ice Sheets}}.
\newblock \bibinfo{publisher}{Springer}.
\newblock
\showISBNx{9789027714732}
\showLCCN{lc83004479}
\urldef\tempurl%
\url{https://doi.org/10.1007/978-94-015-1167-4}
\showDOI{\tempurl}


\bibitem[Ibikunle et~al\mbox{.}(2020)]%
        {ibikunle_2020}
\bibfield{author}{\bibinfo{person}{Oluwanisola Ibikunle}, \bibinfo{person}{John Paden}, \bibinfo{person}{Maryam Rahnemoonfar}, \bibinfo{person}{David Crandall}, {and} \bibinfo{person}{Masoud Yari}.} \bibinfo{year}{2020}\natexlab{}.
\newblock \showarticletitle{Snow Radar Layer Tracking Using Iterative Neural Network Approach}. In \bibinfo{booktitle}{\emph{IGARSS 2020 - 2020 IEEE International Geoscience and Remote Sensing Symposium}}. \bibinfo{pages}{2960--2963}.
\newblock
\urldef\tempurl%
\url{https://doi.org/10.1109/IGARSS39084.2020.9323957}
\showDOI{\tempurl}


\bibitem[Ibikunle et~al\mbox{.}(2023)]%
        {ibikunle_snow_radar_echogram_2023}
\bibfield{author}{\bibinfo{person}{Oluwanisola Ibikunle}, \bibinfo{person}{Hara~Madhav Talasila}, \bibinfo{person}{Debvrat Varshney}, \bibinfo{person}{John~D. Paden}, \bibinfo{person}{Jilu Li}, {and} \bibinfo{person}{Maryam Rahnemoonfar}.} \bibinfo{year}{2023}\natexlab{}.
\newblock \showarticletitle{Snow Radar Echogram Layer Tracker: Deep Neural Networks for radar data from NASA Operation IceBridge}. In \bibinfo{booktitle}{\emph{2023 IEEE Radar Conference (RadarConf23)}}. \bibinfo{pages}{1--6}.
\newblock
\urldef\tempurl%
\url{https://doi.org/10.1109/RadarConf2351548.2023.10149734}
\showDOI{\tempurl}


\bibitem[Iwasaki and Lai(2023)]%
        {Hardness}
\bibfield{author}{\bibinfo{person}{Yunona Iwasaki} {and} \bibinfo{person}{Ching-Yao Lai}.} \bibinfo{year}{2023}\natexlab{}.
\newblock \showarticletitle{One-dimensional ice shelf hardness inversion: Clustering behavior and collocation resampling in physics-informed neural networks}.
\newblock \bibinfo{journal}{\emph{J. Comput. Phys.}}  \bibinfo{volume}{492} (\bibinfo{year}{2023}), \bibinfo{pages}{112435}.
\newblock
\showISSN{0021-9991}
\urldef\tempurl%
\url{https://doi.org/10.1016/j.jcp.2023.112435}
\showDOI{\tempurl}


\bibitem[Jouvet and Cordonnier(2023)]%
        {jouvet_cordonnier_2023}
\bibfield{author}{\bibinfo{person}{Guillaume Jouvet} {and} \bibinfo{person}{Guillaume Cordonnier}.} \bibinfo{year}{2023}\natexlab{}.
\newblock \showarticletitle{Ice-flow model emulator based on physics-informed deep learning}.
\newblock \bibinfo{journal}{\emph{Journal of Glaciology}} (\bibinfo{year}{2023}), \bibinfo{pages}{1–15}.
\newblock
\urldef\tempurl%
\url{https://doi.org/10.1017/jog.2023.73}
\showDOI{\tempurl}


\bibitem[Jouvet et~al\mbox{.}(2022)]%
        {jouvet_cordonnier_kim_lüthi_vieli_aschwanden_2022}
\bibfield{author}{\bibinfo{person}{Guillaume Jouvet}, \bibinfo{person}{Guillaume Cordonnier}, \bibinfo{person}{Byungsoo Kim}, \bibinfo{person}{Martin Lüthi}, \bibinfo{person}{Andreas Vieli}, {and} \bibinfo{person}{Andy Aschwanden}.} \bibinfo{year}{2022}\natexlab{}.
\newblock \showarticletitle{Deep learning speeds up ice flow modelling by several orders of magnitude}.
\newblock \bibinfo{journal}{\emph{Journal of Glaciology}} \bibinfo{volume}{68}, \bibinfo{number}{270} (\bibinfo{year}{2022}), \bibinfo{pages}{651–664}.
\newblock
\urldef\tempurl%
\url{https://doi.org/10.1017/jog.2021.120}
\showDOI{\tempurl}


\bibitem[Kamangir et~al\mbox{.}(2018)]%
        {kamangir}
\bibfield{author}{\bibinfo{person}{Hamid Kamangir}, \bibinfo{person}{Maryam Rahnemoonfar}, \bibinfo{person}{Dugan Dobbs}, \bibinfo{person}{John Paden}, {and} \bibinfo{person}{Geoffrey Fox}.} \bibinfo{year}{2018}\natexlab{}.
\newblock \showarticletitle{Deep Hybrid Wavelet Network for Ice Boundary Detection in Radra Imagery}. In \bibinfo{booktitle}{\emph{IGARSS 2018 - 2018 IEEE International Geoscience and Remote Sensing Symposium}}. \bibinfo{pages}{3449--3452}.
\newblock
\urldef\tempurl%
\url{https://doi.org/10.1109/IGARSS.2018.8518617}
\showDOI{\tempurl}


\bibitem[Karniadakis et~al\mbox{.}(2021)]%
        {Karniadakis2021}
\bibfield{author}{\bibinfo{person}{George~Em Karniadakis}, \bibinfo{person}{Ioannis~G. Kevrekidis}, \bibinfo{person}{Lu Lu}, \bibinfo{person}{Paris Perdikaris}, \bibinfo{person}{Sifan Wang}, {and} \bibinfo{person}{Liu Yang}.} \bibinfo{year}{2021}\natexlab{}.
\newblock \showarticletitle{Physics-informed machine learning}.
\newblock \bibinfo{journal}{\emph{Nature Reviews Physics}} \bibinfo{volume}{3}, \bibinfo{number}{6} (\bibinfo{date}{01 Jun} \bibinfo{year}{2021}), \bibinfo{pages}{422--440}.
\newblock
\showISSN{2522-5820}
\urldef\tempurl%
\url{https://doi.org/10.1038/s42254-021-00314-5}
\showDOI{\tempurl}


\bibitem[Kashinath et~al\mbox{.}(2021)]%
        {survey_climate}
\bibfield{author}{\bibinfo{person}{K. Kashinath}, \bibinfo{person}{M. Mustafa}, \bibinfo{person}{A. Albert}, \bibinfo{person}{J-L. Wu}, \bibinfo{person}{C. Jiang}, \bibinfo{person}{S. Esmaeilzadeh}, \bibinfo{person}{K. Azizzadenesheli}, \bibinfo{person}{R. Wang}, \bibinfo{person}{A. Chattopadhyay}, \bibinfo{person}{A. Singh}, \bibinfo{person}{A. Manepalli}, {and} \bibinfo{person}{et al.}} \bibinfo{year}{2021}\natexlab{}.
\newblock \showarticletitle{Physics-informed machine learning: case studies for weather and climate modelling}.
\newblock \bibinfo{journal}{\emph{Philosophical Transactions of the Royal Society A: Mathematical, Physical and Engineering Sciences}} \bibinfo{volume}{379}, \bibinfo{number}{2194} (\bibinfo{year}{2021}), \bibinfo{pages}{20200093}.
\newblock
\urldef\tempurl%
\url{https://doi.org/10.1098/rsta.2020.0093}
\showDOI{\tempurl}
\showeprint{https://royalsocietypublishing.org/doi/pdf/10.1098/rsta.2020.0093}


\bibitem[Keegan et~al\mbox{.}(2014)]%
        {Land_Ice_Interaction}
\bibfield{author}{\bibinfo{person}{Kaitlin~M. Keegan}, \bibinfo{person}{Mary~R. Albert}, \bibinfo{person}{Joseph~R. McConnell}, {and} \bibinfo{person}{Ian Baker}.} \bibinfo{year}{2014}\natexlab{}.
\newblock \showarticletitle{Climate change and forest fires synergistically drive widespread melt events of the Greenland Ice Sheet}.
\newblock \bibinfo{journal}{\emph{Proceedings of the National Academy of Sciences}} \bibinfo{volume}{111}, \bibinfo{number}{22} (\bibinfo{year}{2014}), \bibinfo{pages}{7964--7967}.
\newblock
\urldef\tempurl%
\url{https://doi.org/10.1073/pnas.1405397111}
\showDOI{\tempurl}
\showeprint{https://www.pnas.org/doi/pdf/10.1073/pnas.1405397111}


\bibitem[Kharazmi et~al\mbox{.}(2019)]%
        {variationalpinn}
\bibfield{author}{\bibinfo{person}{Ehsan Kharazmi}, \bibinfo{person}{Zhongqiang Zhang}, {and} \bibinfo{person}{George~Em Karniadakis}.} \bibinfo{year}{2019}\natexlab{}.
\newblock \showarticletitle{Variational Physics-Informed Neural Networks For Solving Partial Differential Equations}.
\newblock \bibinfo{journal}{\emph{CoRR}}  \bibinfo{volume}{abs/1912.00873} (\bibinfo{year}{2019}).
\newblock
\showeprint[arXiv]{1912.00873}
\urldef\tempurl%
\url{http://arxiv.org/abs/1912.00873}
\showURL{%
\tempurl}


\bibitem[Kim et~al\mbox{.}(2019)]%
        {Kim2019_ANN}
\bibfield{author}{\bibinfo{person}{Jiwon Kim}, \bibinfo{person}{Kwangjin Kim}, \bibinfo{person}{Jaeil Cho}, \bibinfo{person}{Yong~Q. Kang}, \bibinfo{person}{Hong-Joo Yoon}, {and} \bibinfo{person}{Yang-Won Lee}.} \bibinfo{year}{2019}\natexlab{}.
\newblock \showarticletitle{Satellite-Based Prediction of Arctic Sea Ice Concentration Using a Deep Neural Network with Multi-Model Ensemble}.
\newblock \bibinfo{journal}{\emph{Remote Sensing}} \bibinfo{volume}{11}, \bibinfo{number}{1} (\bibinfo{year}{2019}).
\newblock
\showISSN{2072-4292}
\urldef\tempurl%
\url{https://doi.org/10.3390/rs11010019}
\showDOI{\tempurl}


\bibitem[Kim et~al\mbox{.}(2023)]%
        {Kim2023_GAN}
\bibfield{author}{\bibinfo{person}{Mingyu Kim}, \bibinfo{person}{Jaekyeong Lee}, \bibinfo{person}{Leechan Choi}, {and} \bibinfo{person}{Minjoo Choi}.} \bibinfo{year}{2023}\natexlab{}.
\newblock \showarticletitle{PolarGAN: Creating realistic Arctic sea ice concentration images with user-defined geometric preferences}.
\newblock \bibinfo{journal}{\emph{Engineering Applications of Artificial Intelligence}}  \bibinfo{volume}{126} (\bibinfo{year}{2023}), \bibinfo{pages}{106920}.
\newblock
\showISSN{0952-1976}
\urldef\tempurl%
\url{https://doi.org/10.1016/j.engappai.2023.106920}
\showDOI{\tempurl}


\bibitem[Kim et~al\mbox{.}(2021)]%
        {Kim2021}
\bibfield{author}{\bibinfo{person}{Sung~Wook Kim}, \bibinfo{person}{Iljeok Kim}, \bibinfo{person}{Jonghwan Lee}, {and} \bibinfo{person}{Seungchul Lee}.} \bibinfo{year}{2021}\natexlab{}.
\newblock \showarticletitle{Knowledge Integration into deep learning in dynamical systems: an overview and taxonomy}.
\newblock \bibinfo{journal}{\emph{Journal of Mechanical Science and Technology}} \bibinfo{volume}{35}, \bibinfo{number}{4} (\bibinfo{date}{01 Apr} \bibinfo{year}{2021}), \bibinfo{pages}{1331--1342}.
\newblock
\showISSN{1976-3824}
\urldef\tempurl%
\url{https://doi.org/10.1007/s12206-021-0342-5}
\showDOI{\tempurl}


\bibitem[Kim et~al\mbox{.}(2020)]%
        {Kim2020_SIC_CNN}
\bibfield{author}{\bibinfo{person}{Y.~J. Kim}, \bibinfo{person}{H.-C. Kim}, \bibinfo{person}{D. Han}, \bibinfo{person}{S. Lee}, {and} \bibinfo{person}{J. Im}.} \bibinfo{year}{2020}\natexlab{}.
\newblock \showarticletitle{Prediction of monthly Arctic sea ice concentrations using satellite and reanalysis data based on convolutional neural networks}.
\newblock \bibinfo{journal}{\emph{The Cryosphere}} \bibinfo{volume}{14}, \bibinfo{number}{3} (\bibinfo{year}{2020}), \bibinfo{pages}{1083--1104}.
\newblock
\urldef\tempurl%
\url{https://doi.org/10.5194/tc-14-1083-2020}
\showDOI{\tempurl}


\bibitem[Koo and Rahnemoonfar(2023)]%
        {koo2023multitask}
\bibfield{author}{\bibinfo{person}{Younghyun Koo} {and} \bibinfo{person}{Maryam Rahnemoonfar}.} \bibinfo{year}{2023}\natexlab{}.
\newblock \bibinfo{title}{Multi-task Deep Convolutional Network to Predict Sea Ice Concentration and Drift in the Arctic Ocean}.
\newblock
\newblock
\showeprint[arxiv]{2311.00167}~[cs.LG]


\bibitem[Kwok(2018)]%
        {Kwok2018}
\bibfield{author}{\bibinfo{person}{R Kwok}.} \bibinfo{year}{2018}\natexlab{}.
\newblock \showarticletitle{Arctic sea ice thickness, volume, and multiyear ice coverage: losses and coupled variability (1958-2018)}.
\newblock \bibinfo{journal}{\emph{Environmental Research Letters}} \bibinfo{volume}{13}, \bibinfo{number}{10} (\bibinfo{date}{oct} \bibinfo{year}{2018}), \bibinfo{pages}{105005}.
\newblock
\urldef\tempurl%
\url{https://doi.org/10.1088/1748-9326/aae3ec}
\showDOI{\tempurl}


\bibitem[Larour et~al\mbox{.}(2012)]%
        {Larour2012}
\bibfield{author}{\bibinfo{person}{E. Larour}, \bibinfo{person}{H. Seroussi}, \bibinfo{person}{M. Morlighem}, {and} \bibinfo{person}{E. Rignot}.} \bibinfo{year}{2012}\natexlab{}.
\newblock \showarticletitle{Continental scale, high order, high spatial resolution, ice sheet modeling using the Ice Sheet System Model (ISSM)}.
\newblock \bibinfo{journal}{\emph{Journal of Geophysical Research: Earth Surface}} \bibinfo{volume}{117}, \bibinfo{number}{F1} (\bibinfo{year}{2012}).
\newblock
\urldef\tempurl%
\url{https://doi.org/10.1029/2011JF002140}
\showDOI{\tempurl}
\showeprint{https://agupubs.onlinelibrary.wiley.com/doi/pdf/10.1029/2011JF002140}


\bibitem[Lazzara et~al\mbox{.}(2012)]%
        {onsitestation}
\bibfield{author}{\bibinfo{person}{Matthew~A. Lazzara}, \bibinfo{person}{George~A. Weidner}, \bibinfo{person}{Linda~M. Keller}, \bibinfo{person}{Jonathan~E. Thom}, {and} \bibinfo{person}{John~J. Cassano}.} \bibinfo{year}{2012}\natexlab{}.
\newblock \showarticletitle{Antarctic Automatic Weather Station Program: 30 Years of Polar Observation}.
\newblock \bibinfo{journal}{\emph{Bulletin of the American Meteorological Society}} \bibinfo{volume}{93}, \bibinfo{number}{10} (\bibinfo{year}{2012}), \bibinfo{pages}{1519 -- 1537}.
\newblock
\urldef\tempurl%
\url{https://doi.org/10.1175/BAMS-D-11-00015.1}
\showDOI{\tempurl}


\bibitem[Leong and Horgan(2020)]%
        {Leong2020_DeepBedMap}
\bibfield{author}{\bibinfo{person}{W.~J. Leong} {and} \bibinfo{person}{H.~J. Horgan}.} \bibinfo{year}{2020}\natexlab{}.
\newblock \showarticletitle{DeepBedMap: a deep neural network for resolving the bed topography of Antarctica}.
\newblock \bibinfo{journal}{\emph{The Cryosphere}} \bibinfo{volume}{14}, \bibinfo{number}{11} (\bibinfo{year}{2020}), \bibinfo{pages}{3687--3705}.
\newblock
\urldef\tempurl%
\url{https://doi.org/10.5194/tc-14-3687-2020}
\showDOI{\tempurl}


\bibitem[Li et~al\mbox{.}(2024)]%
        {physics_informed_neural_operator}
\bibfield{author}{\bibinfo{person}{Zongyi Li}, \bibinfo{person}{Hongkai Zheng}, \bibinfo{person}{Nikola Kovachki}, \bibinfo{person}{David Jin}, \bibinfo{person}{Haoxuan Chen}, \bibinfo{person}{Burigede Liu}, \bibinfo{person}{Kamyar Azizzadenesheli}, {and} \bibinfo{person}{Anima Anandkumar}.} \bibinfo{year}{2024}\natexlab{}.
\newblock \showarticletitle{Physics-Informed Neural Operator for Learning Partial Differential Equations}.
\newblock \bibinfo{journal}{\emph{ACM / IMS J. Data Sci.}} (\bibinfo{date}{feb} \bibinfo{year}{2024}).
\newblock
\urldef\tempurl%
\url{https://doi.org/10.1145/3648506}
\showDOI{\tempurl}
\newblock
\shownote{Just Accepted}.


\bibitem[Liang et~al\mbox{.}(2023)]%
        {Liang2023}
\bibfield{author}{\bibinfo{person}{Zeyu Liang}, \bibinfo{person}{Qing Ji}, \bibinfo{person}{Xiaoping Pang}, \bibinfo{person}{Pei Fan}, \bibinfo{person}{Xuedong Yao}, \bibinfo{person}{Yizhuo Chen}, \bibinfo{person}{Ying Chen}, {and} \bibinfo{person}{Zhongnan Yan}.} \bibinfo{year}{2023}\natexlab{}.
\newblock \showarticletitle{Estimation of Daily Arctic Winter Sea Ice Thickness from Thermodynamic Parameters Using a Self-Attention Convolutional Neural Network}.
\newblock \bibinfo{journal}{\emph{Remote Sensing}} \bibinfo{volume}{15}, \bibinfo{number}{7} (\bibinfo{year}{2023}).
\newblock
\showISSN{2072-4292}
\urldef\tempurl%
\url{https://doi.org/10.3390/rs15071887}
\showDOI{\tempurl}


\bibitem[Lin et~al\mbox{.}(2023)]%
        {Lin2023_GNN}
\bibfield{author}{\bibinfo{person}{Yucheng Lin}, \bibinfo{person}{Pippa~L. Whitehouse}, \bibinfo{person}{Andrew~P. Valentine}, {and} \bibinfo{person}{Sarah~A. Woodroffe}.} \bibinfo{year}{2023}\natexlab{}.
\newblock \showarticletitle{GEORGIA: A Graph Neural Network Based EmulatOR for Glacial Isostatic Adjustment}.
\newblock \bibinfo{journal}{\emph{Geophysical Research Letters}} \bibinfo{volume}{50}, \bibinfo{number}{18} (\bibinfo{year}{2023}), \bibinfo{pages}{e2023GL103672}.
\newblock
\urldef\tempurl%
\url{https://doi.org/10.1029/2023GL103672}
\showDOI{\tempurl}
\showeprint{https://agupubs.onlinelibrary.wiley.com/doi/pdf/10.1029/2023GL103672}
\newblock
\shownote{e2023GL103672 2023GL103672}.


\bibitem[Liu et~al\mbox{.}(2023a)]%
        {liu2023harnessing}
\bibfield{author}{\bibinfo{person}{Ning Liu}, \bibinfo{person}{Yiming Fan}, \bibinfo{person}{Xianyi Zeng}, \bibinfo{person}{Milan Klöwer}, {and} \bibinfo{person}{Yue Yu}.} \bibinfo{year}{2023}\natexlab{a}.
\newblock \bibinfo{title}{Harnessing the Power of Neural Operators with Automatically Encoded Conservation Laws}.
\newblock
\newblock
\showeprint[arxiv]{2312.11176}~[cs.LG]


\bibitem[Liu et~al\mbox{.}(2023b)]%
        {liu2023ino}
\bibfield{author}{\bibinfo{person}{Ning Liu}, \bibinfo{person}{Yue Yu}, \bibinfo{person}{Huaiqian You}, {and} \bibinfo{person}{Neeraj Tatikola}.} \bibinfo{year}{2023}\natexlab{b}.
\newblock \bibinfo{title}{INO: Invariant Neural Operators for Learning Complex Physical Systems with Momentum Conservation}.
\newblock
\newblock
\showeprint[arxiv]{2212.14365}~[cs.LG]


\bibitem[Liu et~al\mbox{.}(2021b)]%
        {Liu2021_LSTM_SIC}
\bibfield{author}{\bibinfo{person}{Quanhong Liu}, \bibinfo{person}{Ren Zhang}, \bibinfo{person}{Yangjun Wang}, \bibinfo{person}{Hengqian Yan}, {and} \bibinfo{person}{Mei Hong}.} \bibinfo{year}{2021}\natexlab{b}.
\newblock \showarticletitle{Daily Prediction of the Arctic Sea Ice Concentration Using Reanalysis Data Based on a Convolutional LSTM Network}.
\newblock \bibinfo{journal}{\emph{Journal of Marine Science and Engineering}} \bibinfo{volume}{9}, \bibinfo{number}{3} (\bibinfo{year}{2021}).
\newblock
\showISSN{2077-1312}
\urldef\tempurl%
\url{https://doi.org/10.3390/jmse9030330}
\showDOI{\tempurl}


\bibitem[Liu et~al\mbox{.}(2021a)]%
        {Liu_seaice_ConvLSTM}
\bibfield{author}{\bibinfo{person}{Yang Liu}, \bibinfo{person}{Laurens Bogaardt}, \bibinfo{person}{Jisk Attema}, {and} \bibinfo{person}{Wilco Hazeleger}.} \bibinfo{year}{2021}\natexlab{a}.
\newblock \showarticletitle{Extended-Range Arctic Sea Ice Forecast with Convolutional Long Short-Term Memory Networks}.
\newblock \bibinfo{journal}{\emph{Monthly Weather Review}} \bibinfo{volume}{149}, \bibinfo{number}{6} (\bibinfo{year}{2021}), \bibinfo{pages}{1673 -- 1693}.
\newblock
\urldef\tempurl%
\url{https://doi.org/10.1175/MWR-D-20-0113.1}
\showDOI{\tempurl}


\bibitem[Loebel et~al\mbox{.}(2022)]%
        {Loebel2022}
\bibfield{author}{\bibinfo{person}{Erik Loebel}, \bibinfo{person}{Mirko Scheinert}, \bibinfo{person}{Martin Horwath}, \bibinfo{person}{Konrad Heidler}, \bibinfo{person}{Julia Christmann}, \bibinfo{person}{Long~Duc Phan}, \bibinfo{person}{Angelika Humbert}, {and} \bibinfo{person}{Xiao~Xiang Zhu}.} \bibinfo{year}{2022}\natexlab{}.
\newblock \showarticletitle{Extracting Glacier Calving Fronts by Deep Learning: The Benefit of Multispectral, Topographic, and Textural Input Features}.
\newblock \bibinfo{journal}{\emph{IEEE Transactions on Geoscience and Remote Sensing}}  \bibinfo{volume}{60} (\bibinfo{year}{2022}), \bibinfo{pages}{1--12}.
\newblock
\urldef\tempurl%
\url{https://doi.org/10.1109/TGRS.2022.3208454}
\showDOI{\tempurl}


\bibitem[Long and Zhe(2024)]%
        {long2024invertible}
\bibfield{author}{\bibinfo{person}{Da Long} {and} \bibinfo{person}{Shandian Zhe}.} \bibinfo{year}{2024}\natexlab{}.
\newblock \bibinfo{title}{Invertible Fourier Neural Operators for Tackling Both Forward and Inverse Problems}.
\newblock
\newblock
\showeprint[arxiv]{2402.11722}~[cs.LG]


\bibitem[Lu et~al\mbox{.}(2021)]%
        {Lu2021}
\bibfield{author}{\bibinfo{person}{Lu Lu}, \bibinfo{person}{Pengzhan Jin}, \bibinfo{person}{Guofei Pang}, \bibinfo{person}{Zhongqiang Zhang}, {and} \bibinfo{person}{George~Em Karniadakis}.} \bibinfo{year}{2021}\natexlab{}.
\newblock \showarticletitle{Learning nonlinear operators via DeepONet based on the universal approximation theorem of operators}.
\newblock \bibinfo{journal}{\emph{Nature Machine Intelligence}} \bibinfo{volume}{3}, \bibinfo{number}{3} (\bibinfo{date}{01 Mar} \bibinfo{year}{2021}), \bibinfo{pages}{218--229}.
\newblock
\showISSN{2522-5839}
\urldef\tempurl%
\url{https://doi.org/10.1038/s42256-021-00302-5}
\showDOI{\tempurl}


\bibitem[MacAyeal(1989)]%
        {SSA1}
\bibfield{author}{\bibinfo{person}{Douglas~R. MacAyeal}.} \bibinfo{year}{1989}\natexlab{}.
\newblock \showarticletitle{Large-scale ice flow over a viscous basal sediment: Theory and application to ice stream B, Antarctica}.
\newblock \bibinfo{journal}{\emph{Journal of Geophysical Research: Solid Earth}} \bibinfo{volume}{94}, \bibinfo{number}{B4} (\bibinfo{year}{1989}), \bibinfo{pages}{4071--4087}.
\newblock
\urldef\tempurl%
\url{https://doi.org/10.1029/JB094iB04p04071}
\showDOI{\tempurl}
\showeprint{https://agupubs.onlinelibrary.wiley.com/doi/pdf/10.1029/JB094iB04p04071}


\bibitem[MacAyeal(1993)]%
        {control2}
\bibfield{author}{\bibinfo{person}{Douglas~R. MacAyeal}.} \bibinfo{year}{1993}\natexlab{}.
\newblock \showarticletitle{A tutorial on the use of control methods in ice-sheet modeling}.
\newblock \bibinfo{journal}{\emph{Journal of Glaciology}} \bibinfo{volume}{39}, \bibinfo{number}{131} (\bibinfo{year}{1993}), \bibinfo{pages}{91–98}.
\newblock
\urldef\tempurl%
\url{https://doi.org/10.3189/S0022143000015744}
\showDOI{\tempurl}


\bibitem[Mallat(1989)]%
        {mallat1989}
\bibfield{author}{\bibinfo{person}{S.G. Mallat}.} \bibinfo{year}{1989}\natexlab{}.
\newblock \showarticletitle{A theory for multiresolution signal decomposition: the wavelet representation}.
\newblock \bibinfo{journal}{\emph{IEEE Transactions on Pattern Analysis and Machine Intelligence}} \bibinfo{volume}{11}, \bibinfo{number}{7} (\bibinfo{year}{1989}), \bibinfo{pages}{674--693}.
\newblock
\urldef\tempurl%
\url{https://doi.org/10.1109/34.192463}
\showDOI{\tempurl}


\bibitem[Markidis(2021)]%
        {basicpinn}
\bibfield{author}{\bibinfo{person}{Stefano Markidis}.} \bibinfo{year}{2021}\natexlab{}.
\newblock \showarticletitle{The Old and the New: Can Physics-Informed Deep-Learning Replace Traditional Linear Solvers?}
\newblock \bibinfo{journal}{\emph{Frontiers in Big Data}}  \bibinfo{volume}{4} (\bibinfo{year}{2021}).
\newblock
\showISSN{2624-909X}
\urldef\tempurl%
\url{https://doi.org/10.3389/fdata.2021.669097}
\showDOI{\tempurl}


\bibitem[Martin and Fowlkes(2001)]%
        {HEDbenchmark}
\bibfield{author}{\bibinfo{person}{David Martin} {and} \bibinfo{person}{Charless Fowlkes}.} \bibinfo{year}{2001}\natexlab{}.
\newblock \bibinfo{title}{The Berkeley Segmentation Database and Benchmark}.
\newblock \bibinfo{howpublished}{Computer Science Department, Berkeley University}.
\newblock
\urldef\tempurl%
\url{http://www.eecs.berkeley.edu/Research/Projects/CS/vision/bsds}
\showURL{%
\tempurl}


\bibitem[McClenny and Braga-Neto(2023)]%
        {MCCLENNY2023111722selfadaptive}
\bibfield{author}{\bibinfo{person}{Levi~D. McClenny} {and} \bibinfo{person}{Ulisses~M. Braga-Neto}.} \bibinfo{year}{2023}\natexlab{}.
\newblock \showarticletitle{Self-adaptive physics-informed neural networks}.
\newblock \bibinfo{journal}{\emph{J. Comput. Phys.}}  \bibinfo{volume}{474} (\bibinfo{year}{2023}), \bibinfo{pages}{111722}.
\newblock
\showISSN{0021-9991}
\urldef\tempurl%
\url{https://doi.org/10.1016/j.jcp.2022.111722}
\showDOI{\tempurl}


\bibitem[McGovern et~al\mbox{.}(2019)]%
        {MakingtheBlackBoxMoreTransparentUnderstandingthePhysicalImplicationsofMachineLearning}
\bibfield{author}{\bibinfo{person}{Amy McGovern}, \bibinfo{person}{Ryan Lagerquist}, \bibinfo{person}{David~John Gagne}, \bibinfo{person}{G.~Eli Jergensen}, \bibinfo{person}{Kimberly~L. Elmore}, \bibinfo{person}{Cameron~R. Homeyer}, {and} \bibinfo{person}{Travis Smith}.} \bibinfo{year}{2019}\natexlab{}.
\newblock \showarticletitle{Making the Black Box More Transparent: Understanding the Physical Implications of Machine Learning}.
\newblock \bibinfo{journal}{\emph{Bulletin of the American Meteorological Society}} \bibinfo{volume}{100}, \bibinfo{number}{11} (\bibinfo{year}{2019}), \bibinfo{pages}{2175 -- 2199}.
\newblock
\urldef\tempurl%
\url{https://doi.org/10.1175/BAMS-D-18-0195.1}
\showDOI{\tempurl}


\bibitem[Mitchell(2007)]%
        {needforbias}
\bibfield{author}{\bibinfo{person}{Tom~Michael Mitchell}.} \bibinfo{year}{2007}\natexlab{}.
\newblock \showarticletitle{The Need for Biases in Learning Generalizations}.
\newblock
\urldef\tempurl%
\url{https://api.semanticscholar.org/CorpusID:3237155}
\showURL{%
\tempurl}


\bibitem[Mohajerani et~al\mbox{.}(2019)]%
        {Mohajerani2019}
\bibfield{author}{\bibinfo{person}{Yara Mohajerani}, \bibinfo{person}{Michael Wood}, \bibinfo{person}{Isabella Velicogna}, {and} \bibinfo{person}{Eric Rignot}.} \bibinfo{year}{2019}\natexlab{}.
\newblock \showarticletitle{Detection of Glacier Calving Margins with Convolutional Neural Networks: A Case Study}.
\newblock \bibinfo{journal}{\emph{Remote Sensing}} \bibinfo{volume}{11}, \bibinfo{number}{1} (\bibinfo{year}{2019}).
\newblock
\showISSN{2072-4292}
\urldef\tempurl%
\url{https://doi.org/10.3390/rs11010074}
\showDOI{\tempurl}


\bibitem[Molinaro et~al\mbox{.}(2023)]%
        {molinaro_neural_inverse_operators}
\bibfield{author}{\bibinfo{person}{Roberto Molinaro}, \bibinfo{person}{Yunan Yang}, \bibinfo{person}{Björn Engquist}, {and} \bibinfo{person}{Siddhartha Mishra}.} \bibinfo{year}{2023}\natexlab{}.
\newblock \bibinfo{title}{Neural Inverse Operators for Solving PDE Inverse Problems}.
\newblock
\newblock
\showeprint[arxiv]{2301.11167}~[cs.LG]


\bibitem[Morland(1987)]%
        {SSA2}
\bibfield{author}{\bibinfo{person}{L.~W. Morland}.} \bibinfo{year}{1987}\natexlab{}.
\newblock \showarticletitle{Unconfined Ice-Shelf Flow}. In \bibinfo{booktitle}{\emph{Dynamics of the West Antarctic Ice Sheet}}, \bibfield{editor}{\bibinfo{person}{C.~J. Van~der Veen} {and} \bibinfo{person}{J.~Oerlemans}} (Eds.). \bibinfo{publisher}{Springer Netherlands}, \bibinfo{address}{Dordrecht}, \bibinfo{pages}{99--116}.
\newblock
\showISBNx{978-94-009-3745-1}


\bibitem[Morlighem(2022)]%
        {Morlighem}
\bibfield{author}{\bibinfo{person}{M. Morlighem}.} \bibinfo{year}{2022}\natexlab{}.
\newblock \bibinfo{title}{MEaSUREs BedMachine Antarctica, Version 3}.
\newblock
\newblock
\urldef\tempurl%
\url{https://doi.org/10.5067/FPSU0V1MWUB6}
\showDOI{\tempurl}


\bibitem[Morlighem et~al\mbox{.}(2020)]%
        {morlighem2020_bedmachine_Ant}
\bibfield{author}{\bibinfo{person}{Mathieu Morlighem}, \bibinfo{person}{Eric Rignot}, \bibinfo{person}{Tobias Binder}, \bibinfo{person}{Donald Blankenship}, \bibinfo{person}{Reinhard Drews}, \bibinfo{person}{Graeme Eagles}, \bibinfo{person}{Olaf Eisen}, \bibinfo{person}{Fausto Ferraccioli}, \bibinfo{person}{Ren{\'e} Forsberg}, \bibinfo{person}{Peter Fretwell}, {et~al\mbox{.}}} \bibinfo{year}{2020}\natexlab{}.
\newblock \showarticletitle{Deep glacial troughs and stabilizing ridges unveiled beneath the margins of the Antarctic ice sheet}.
\newblock \bibinfo{journal}{\emph{Nature geoscience}} \bibinfo{volume}{13}, \bibinfo{number}{2} (\bibinfo{year}{2020}), \bibinfo{pages}{132--137}.
\newblock


\bibitem[Morlighem et~al\mbox{.}(2014)]%
        {Morlighem2014}
\bibfield{author}{\bibinfo{person}{M. Morlighem}, \bibinfo{person}{E. Rignot}, \bibinfo{person}{J. Mouginot}, \bibinfo{person}{H. Seroussi}, {and} \bibinfo{person}{E. Larour}.} \bibinfo{year}{2014}\natexlab{}.
\newblock \showarticletitle{Deeply incised submarine glacial valleys beneath the Greenland ice sheet}.
\newblock \bibinfo{journal}{\emph{Nature Geoscience}}  \bibinfo{volume}{7} (\bibinfo{year}{2014}).
\newblock
\urldef\tempurl%
\url{https://doi.org/10.1038/ngeo2167}
\showDOI{\tempurl}


\bibitem[Morlighem et~al\mbox{.}(2013)]%
        {Morlighem2013}
\bibfield{author}{\bibinfo{person}{M. Morlighem}, \bibinfo{person}{E. Rignot}, \bibinfo{person}{J. Mouginot}, \bibinfo{person}{X. Wu}, \bibinfo{person}{H. Seroussi}, \bibinfo{person}{E. Larour}, {and} \bibinfo{person}{J. Paden}.} \bibinfo{year}{2013}\natexlab{}.
\newblock \showarticletitle{High-resolution bed topography mapping of Russell Glacier, Greenland, inferred from Operation IceBridge data}.
\newblock \bibinfo{journal}{\emph{Journal of Glaciology}} \bibinfo{volume}{59}, \bibinfo{number}{218} (\bibinfo{year}{2013}), \bibinfo{pages}{1015–1023}.
\newblock
\urldef\tempurl%
\url{https://doi.org/10.3189/2013JoG12J235}
\showDOI{\tempurl}


\bibitem[Morlighem et~al\mbox{.}(2011)]%
        {massconservation}
\bibfield{author}{\bibinfo{person}{M. Morlighem}, \bibinfo{person}{E. Rignot}, \bibinfo{person}{H. Seroussi}, \bibinfo{person}{E. Larour}, \bibinfo{person}{H. Ben~Dhia}, {and} \bibinfo{person}{D. Aubry}.} \bibinfo{year}{2011}\natexlab{}.
\newblock \showarticletitle{A mass conservation approach for mapping glacier ice thickness}.
\newblock \bibinfo{journal}{\emph{Geophysical Research Letters}} \bibinfo{volume}{38}, \bibinfo{number}{19} (\bibinfo{year}{2011}).
\newblock
\urldef\tempurl%
\url{https://doi.org/10.1029/2011GL048659}
\showDOI{\tempurl}
\showeprint{https://agupubs.onlinelibrary.wiley.com/doi/pdf/10.1029/2011GL048659}


\bibitem[Morlighem et~al\mbox{.}(2017)]%
        {Morlinghem2017_bedmachine}
\bibfield{author}{\bibinfo{person}{M. Morlighem}, \bibinfo{person}{C.~N. Williams}, \bibinfo{person}{E. Rignot}, \bibinfo{person}{L. An}, \bibinfo{person}{J.~E. Arndt}, \bibinfo{person}{J.~L. Bamber}, \bibinfo{person}{G. Catania}, \bibinfo{person}{N. Chauché}, \bibinfo{person}{J.~A. Dowdeswell}, \bibinfo{person}{B. Dorschel}, {and} \bibinfo{person}{et al.}} \bibinfo{year}{2017}\natexlab{}.
\newblock \showarticletitle{BedMachine v3: Complete Bed Topography and Ocean Bathymetry Mapping of Greenland From Multibeam Echo Sounding Combined With Mass Conservation}.
\newblock \bibinfo{journal}{\emph{Geophysical Research Letters}} \bibinfo{volume}{44}, \bibinfo{number}{21} (\bibinfo{year}{2017}), \bibinfo{pages}{11,051--11,061}.
\newblock
\urldef\tempurl%
\url{https://doi.org/10.1002/2017GL074954}
\showDOI{\tempurl}
\showeprint{https://agupubs.onlinelibrary.wiley.com/doi/pdf/10.1002/2017GL074954}


\bibitem[Mouginot et~al\mbox{.}(2019)]%
        {Mouginot2019}
\bibfield{author}{\bibinfo{person}{Jérémie Mouginot}, \bibinfo{person}{Eric Rignot}, \bibinfo{person}{Anders~A. Bjørk}, \bibinfo{person}{Michiel van~den Broeke}, \bibinfo{person}{Romain Millan}, \bibinfo{person}{Mathieu Morlighem}, \bibinfo{person}{Brice Noël}, \bibinfo{person}{Bernd Scheuchl}, {and} \bibinfo{person}{Michael Wood}.} \bibinfo{year}{2019}\natexlab{}.
\newblock \showarticletitle{Forty-six years of Greenland Ice Sheet mass balance from 1972 to 2018}.
\newblock \bibinfo{journal}{\emph{Proceedings of the National Academy of Sciences}} \bibinfo{volume}{116}, \bibinfo{number}{19} (\bibinfo{year}{2019}), \bibinfo{pages}{9239--9244}.
\newblock
\urldef\tempurl%
\url{https://doi.org/10.1073/pnas.1904242116}
\showDOI{\tempurl}
\showeprint{https://www.pnas.org/doi/pdf/10.1073/pnas.1904242116}


\bibitem[Notz(2012)]%
        {Sea_Ice_Topics}
\bibfield{author}{\bibinfo{person}{Dirk Notz}.} \bibinfo{year}{2012}\natexlab{}.
\newblock \showarticletitle{Challenges in simulating sea ice in Earth System Models}.
\newblock \bibinfo{journal}{\emph{WIREs Climate Change}} \bibinfo{volume}{3}, \bibinfo{number}{6} (\bibinfo{year}{2012}), \bibinfo{pages}{509--526}.
\newblock
\urldef\tempurl%
\url{https://doi.org/10.1002/wcc.189}
\showDOI{\tempurl}
\showeprint{https://wires.onlinelibrary.wiley.com/doi/pdf/10.1002/wcc.189}


\bibitem[Notz and Stroeve(2016)]%
        {Notz2016}
\bibfield{author}{\bibinfo{person}{Dirk Notz} {and} \bibinfo{person}{Julienne Stroeve}.} \bibinfo{year}{2016}\natexlab{}.
\newblock \showarticletitle{Observed Arctic sea-ice loss directly follows anthropogenic CO2 emission}.
\newblock \bibinfo{journal}{\emph{Science}} \bibinfo{volume}{354}, \bibinfo{number}{6313} (\bibinfo{year}{2016}), \bibinfo{pages}{747--750}.
\newblock
\urldef\tempurl%
\url{https://doi.org/10.1126/science.aag2345}
\showDOI{\tempurl}
\showeprint{https://www.science.org/doi/pdf/10.1126/science.aag2345}


\bibitem[Otosaka et~al\mbox{.}(2023)]%
        {Otosaka2023}
\bibfield{author}{\bibinfo{person}{I.~N. Otosaka}, \bibinfo{person}{A. Shepherd}, \bibinfo{person}{E.~R. Ivins}, \bibinfo{person}{N.-J. Schlegel}, \bibinfo{person}{C. Amory}, \bibinfo{person}{M.~R. van~den Broeke}, \bibinfo{person}{M. Horwath}, \bibinfo{person}{I. Joughin}, \bibinfo{person}{M.~D. King}, \bibinfo{person}{G. Krinner}, {and} \bibinfo{person}{et al.}} \bibinfo{year}{2023}\natexlab{}.
\newblock \showarticletitle{Mass balance of the Greenland and Antarctic ice sheets from 1992 to 2020}.
\newblock \bibinfo{journal}{\emph{Earth System Science Data}} \bibinfo{volume}{15}, \bibinfo{number}{4} (\bibinfo{year}{2023}), \bibinfo{pages}{1597--1616}.
\newblock
\urldef\tempurl%
\url{https://doi.org/10.5194/essd-15-1597-2023}
\showDOI{\tempurl}


\bibitem[Pateras et~al\mbox{.}(2023)]%
        {taxonomic_survey}
\bibfield{author}{\bibinfo{person}{Joseph Pateras}, \bibinfo{person}{Pratip Rana}, {and} \bibinfo{person}{Preetam Ghosh}.} \bibinfo{year}{2023}\natexlab{}.
\newblock \showarticletitle{A Taxonomic Survey of Physics-Informed Machine Learning}.
\newblock \bibinfo{journal}{\emph{Applied Sciences}} \bibinfo{volume}{13}, \bibinfo{number}{12} (\bibinfo{year}{2023}).
\newblock
\showISSN{2076-3417}
\urldef\tempurl%
\url{https://doi.org/10.3390/app13126892}
\showDOI{\tempurl}


\bibitem[Pattyn(1996)]%
        {Pattyn1996}
\bibfield{author}{\bibinfo{person}{Frank Pattyn}.} \bibinfo{year}{1996}\natexlab{}.
\newblock \showarticletitle{Numerical modelling of a fast-flowing outlet glacier: experiments with different basal conditions}.
\newblock \bibinfo{journal}{\emph{Annals of Glaciology}}  \bibinfo{volume}{23} (\bibinfo{year}{1996}), \bibinfo{pages}{237–246}.
\newblock
\urldef\tempurl%
\url{https://doi.org/10.3189/S0260305500013495}
\showDOI{\tempurl}


\bibitem[Perovich et~al\mbox{.}(2002)]%
        {Sea_Ice_Evolution}
\bibfield{author}{\bibinfo{person}{D.~K. Perovich}, \bibinfo{person}{T.~C. Grenfell}, \bibinfo{person}{B. Light}, {and} \bibinfo{person}{P.~V. Hobbs}.} \bibinfo{year}{2002}\natexlab{}.
\newblock \showarticletitle{Seasonal evolution of the albedo of multiyear Arctic sea ice}.
\newblock \bibinfo{journal}{\emph{Journal of Geophysical Research: Oceans}} \bibinfo{volume}{107}, \bibinfo{number}{C10} (\bibinfo{year}{2002}), \bibinfo{pages}{SHE 20--1--SHE 20--13}.
\newblock
\urldef\tempurl%
\url{https://doi.org/10.1029/2000JC000438}
\showDOI{\tempurl}
\showeprint{https://agupubs.onlinelibrary.wiley.com/doi/pdf/10.1029/2000JC000438}


\bibitem[Petrou and Tian(2017)]%
        {Petrou_2017_RNN_seaicemotion}
\bibfield{author}{\bibinfo{person}{Zisis~I. Petrou} {and} \bibinfo{person}{YingLi Tian}.} \bibinfo{year}{2017}\natexlab{}.
\newblock \showarticletitle{Prediction of sea ice motion with recurrent neural networks}. In \bibinfo{booktitle}{\emph{2017 IEEE International Geoscience and Remote Sensing Symposium (IGARSS)}}. \bibinfo{pages}{5422--5425}.
\newblock
\urldef\tempurl%
\url{https://doi.org/10.1109/IGARSS.2017.8128230}
\showDOI{\tempurl}


\bibitem[Rahnemoonfar et~al\mbox{.}(2019)]%
        {Rahnemoonfar2019_GAN}
\bibfield{author}{\bibinfo{person}{Maryam Rahnemoonfar}, \bibinfo{person}{Jimmy Johnson}, {and} \bibinfo{person}{John Paden}.} \bibinfo{year}{2019}\natexlab{}.
\newblock \showarticletitle{AI Radar Sensor: Creating Radar Depth Sounder Images Based on Generative Adversarial Network}.
\newblock \bibinfo{journal}{\emph{Sensors}} \bibinfo{volume}{19}, \bibinfo{number}{24} (\bibinfo{year}{2019}).
\newblock
\showISSN{1424-8220}
\urldef\tempurl%
\url{https://doi.org/10.3390/s19245479}
\showDOI{\tempurl}


\bibitem[Rahnemoonfar and Koo(2024)]%
        {rahnemoonfar2024graph}
\bibfield{author}{\bibinfo{person}{Maryam Rahnemoonfar} {and} \bibinfo{person}{Younghyun Koo}.} \bibinfo{year}{2024}\natexlab{}.
\newblock \bibinfo{title}{Graph Neural Networks as Fast and High-fidelity Emulators for Finite-Element Ice Sheet Modeling}.
\newblock
\newblock
\showeprint[arxiv]{2402.05291}~[cs.LG]


\bibitem[Rahnemoonfar et~al\mbox{.}(2020)]%
        {Rahnemoonfar2020_GAN}
\bibfield{author}{\bibinfo{person}{Maryam Rahnemoonfar}, \bibinfo{person}{Masoud Yari}, {and} \bibinfo{person}{John Paden}.} \bibinfo{year}{2020}\natexlab{}.
\newblock \showarticletitle{Radar Sensor Simulation with Generative Adversarial Network}. In \bibinfo{booktitle}{\emph{IGARSS 2020 - 2020 IEEE International Geoscience and Remote Sensing Symposium}}. \bibinfo{pages}{7001--7004}.
\newblock
\urldef\tempurl%
\url{https://doi.org/10.1109/IGARSS39084.2020.9323676}
\showDOI{\tempurl}


\bibitem[Rahnemoonfar et~al\mbox{.}(2021)]%
        {rahnemoonfar_yari_paden_koenig_ibikunle_2021}
\bibfield{author}{\bibinfo{person}{Maryam Rahnemoonfar}, \bibinfo{person}{Masoud Yari}, \bibinfo{person}{John Paden}, \bibinfo{person}{Lora Koenig}, {and} \bibinfo{person}{Oluwanisola Ibikunle}.} \bibinfo{year}{2021}\natexlab{}.
\newblock \showarticletitle{Deep multi-scale learning for automatic tracking of internal layers of ice in radar data}.
\newblock \bibinfo{journal}{\emph{Journal of Glaciology}} \bibinfo{volume}{67}, \bibinfo{number}{261} (\bibinfo{year}{2021}), \bibinfo{pages}{39–48}.
\newblock
\urldef\tempurl%
\url{https://doi.org/10.1017/jog.2020.80}
\showDOI{\tempurl}


\bibitem[Raissi et~al\mbox{.}(2019)]%
        {PINN}
\bibfield{author}{\bibinfo{person}{M. Raissi}, \bibinfo{person}{P. Perdikaris}, {and} \bibinfo{person}{G.E. Karniadakis}.} \bibinfo{year}{2019}\natexlab{}.
\newblock \showarticletitle{Physics-informed neural networks: A deep learning framework for solving forward and inverse problems involving nonlinear partial differential equations}.
\newblock \bibinfo{journal}{\emph{J. Comput. Phys.}}  \bibinfo{volume}{378} (\bibinfo{year}{2019}), \bibinfo{pages}{686--707}.
\newblock
\showISSN{0021-9991}
\urldef\tempurl%
\url{https://doi.org/10.1016/j.jcp.2018.10.045}
\showDOI{\tempurl}


\bibitem[Ren et~al\mbox{.}(2022)]%
        {Ren2022_SIC_CNN}
\bibfield{author}{\bibinfo{person}{Yibin Ren}, \bibinfo{person}{Xiaofeng Li}, {and} \bibinfo{person}{Wenhao Zhang}.} \bibinfo{year}{2022}\natexlab{}.
\newblock \showarticletitle{A Data-Driven Deep Learning Model for Weekly Sea Ice Concentration Prediction of the Pan-Arctic During the Melting Season}.
\newblock \bibinfo{journal}{\emph{IEEE Transactions on Geoscience and Remote Sensing}}  \bibinfo{volume}{60} (\bibinfo{year}{2022}), \bibinfo{pages}{1--19}.
\newblock
\urldef\tempurl%
\url{https://doi.org/10.1109/TGRS.2022.3177600}
\showDOI{\tempurl}


\bibitem[Riel and Minchew(2023)]%
        {riel_minchew_2023_variational}
\bibfield{author}{\bibinfo{person}{Bryan Riel} {and} \bibinfo{person}{Brent Minchew}.} \bibinfo{year}{2023}\natexlab{}.
\newblock \showarticletitle{Variational inference of ice shelf rheology with physics-informed machine learning}.
\newblock \bibinfo{journal}{\emph{Journal of Glaciology}} \bibinfo{volume}{69}, \bibinfo{number}{277} (\bibinfo{year}{2023}), \bibinfo{pages}{1167–1186}.
\newblock
\urldef\tempurl%
\url{https://doi.org/10.1017/jog.2023.8}
\showDOI{\tempurl}


\bibitem[Riel et~al\mbox{.}(2021)]%
        {slip}
\bibfield{author}{\bibinfo{person}{B. Riel}, \bibinfo{person}{B. Minchew}, {and} \bibinfo{person}{T. Bischoff}.} \bibinfo{year}{2021}\natexlab{}.
\newblock \showarticletitle{Data-Driven Inference of the Mechanics of Slip Along Glacier Beds Using Physics-Informed Neural Networks: Case Study on Rutford Ice Stream, Antarctica}.
\newblock \bibinfo{journal}{\emph{Journal of Advances in Modeling Earth Systems}} \bibinfo{volume}{13}, \bibinfo{number}{11} (\bibinfo{year}{2021}), \bibinfo{pages}{e2021MS002621}.
\newblock
\urldef\tempurl%
\url{https://doi.org/10.1029/2021MS002621}
\showDOI{\tempurl}
\showeprint{https://agupubs.onlinelibrary.wiley.com/doi/pdf/10.1029/2021MS002621}
\newblock
\shownote{e2021MS002621 2021MS002621}.


\bibitem[Rignot et~al\mbox{.}(2011)]%
        {Rignot2011}
\bibfield{author}{\bibinfo{person}{E. Rignot}, \bibinfo{person}{I. Velicogna}, \bibinfo{person}{M.~R. van~den Broeke}, \bibinfo{person}{A. Monaghan}, {and} \bibinfo{person}{J.~T.~M. Lenaerts}.} \bibinfo{year}{2011}\natexlab{}.
\newblock \showarticletitle{Acceleration of the contribution of the Greenland and Antarctic ice sheets to sea level rise}.
\newblock \bibinfo{journal}{\emph{Geophysical Research Letters}} \bibinfo{volume}{38}, \bibinfo{number}{5} (\bibinfo{year}{2011}).
\newblock
\urldef\tempurl%
\url{https://doi.org/10.1029/2011GL046583}
\showDOI{\tempurl}
\showeprint{https://agupubs.onlinelibrary.wiley.com/doi/pdf/10.1029/2011GL046583}


\bibitem[Rodwell and Palmer(2007)]%
        {reanalysis2}
\bibfield{author}{\bibinfo{person}{M.~J. Rodwell} {and} \bibinfo{person}{T.~N. Palmer}.} \bibinfo{year}{2007}\natexlab{}.
\newblock \showarticletitle{Using numerical weather prediction to assess climate models}.
\newblock \bibinfo{journal}{\emph{Quarterly Journal of the Royal Meteorological Society}} \bibinfo{volume}{133}, \bibinfo{number}{622} (\bibinfo{year}{2007}), \bibinfo{pages}{129--146}.
\newblock
\urldef\tempurl%
\url{https://doi.org/10.1002/qj.23}
\showDOI{\tempurl}
\showeprint{https://rmets.onlinelibrary.wiley.com/doi/pdf/10.1002/qj.23}


\bibitem[Ronneberger et~al\mbox{.}(2015)]%
        {ronneberger2015unet}
\bibfield{author}{\bibinfo{person}{Olaf Ronneberger}, \bibinfo{person}{Philipp Fischer}, {and} \bibinfo{person}{Thomas Brox}.} \bibinfo{year}{2015}\natexlab{}.
\newblock \bibinfo{title}{U-Net: Convolutional Networks for Biomedical Image Segmentation}.
\newblock
\newblock
\showeprint[arxiv]{1505.04597}~[cs.CV]


\bibitem[Schannwell et~al\mbox{.}(2020)]%
        {Land_Ice_Bed_Properties}
\bibfield{author}{\bibinfo{person}{C. Schannwell}, \bibinfo{person}{R. Drews}, \bibinfo{person}{T.~A. Ehlers}, \bibinfo{person}{O. Eisen}, \bibinfo{person}{C. Mayer}, \bibinfo{person}{M. Malinen}, \bibinfo{person}{E.~C. Smith}, {and} \bibinfo{person}{H. Eisermann}.} \bibinfo{year}{2020}\natexlab{}.
\newblock \showarticletitle{Quantifying the effect of ocean bed properties on ice sheet geometry over 40\,000 years with a full-Stokes model}.
\newblock \bibinfo{journal}{\emph{The Cryosphere}} \bibinfo{volume}{14}, \bibinfo{number}{11} (\bibinfo{year}{2020}), \bibinfo{pages}{3917--3934}.
\newblock
\urldef\tempurl%
\url{https://doi.org/10.5194/tc-14-3917-2020}
\showDOI{\tempurl}


\bibitem[Schweiger et~al\mbox{.}(2011)]%
        {Schweiger2011}
\bibfield{author}{\bibinfo{person}{Axel Schweiger}, \bibinfo{person}{Ron Lindsay}, \bibinfo{person}{Jinlun Zhang}, \bibinfo{person}{Mike Steele}, \bibinfo{person}{Harry Stern}, {and} \bibinfo{person}{Ron Kwok}.} \bibinfo{year}{2011}\natexlab{}.
\newblock \showarticletitle{Uncertainty in modeled Arctic sea ice volume}.
\newblock \bibinfo{journal}{\emph{Journal of Geophysical Research: Oceans}} \bibinfo{volume}{116}, \bibinfo{number}{C8} (\bibinfo{year}{2011}).
\newblock
\urldef\tempurl%
\url{https://doi.org/10.1029/2011JC007084}
\showDOI{\tempurl}
\showeprint{https://agupubs.onlinelibrary.wiley.com/doi/pdf/10.1029/2011JC007084}


\bibitem[Sellevold and Vizcaino(2021)]%
        {Sellevold2021}
\bibfield{author}{\bibinfo{person}{Raymond Sellevold} {and} \bibinfo{person}{Miren Vizcaino}.} \bibinfo{year}{2021}\natexlab{}.
\newblock \showarticletitle{First Application of Artificial Neural Networks to Estimate 21st Century Greenland Ice Sheet Surface Melt}.
\newblock \bibinfo{journal}{\emph{Geophysical Research Letters}} \bibinfo{volume}{48}, \bibinfo{number}{16} (\bibinfo{year}{2021}), \bibinfo{pages}{e2021GL092449}.
\newblock
\urldef\tempurl%
\url{https://doi.org/10.1029/2021GL092449}
\showDOI{\tempurl}
\showeprint{https://agupubs.onlinelibrary.wiley.com/doi/pdf/10.1029/2021GL092449}
\newblock
\shownote{e2021GL092449 2021GL092449}.


\bibitem[Seyyedi et~al\mbox{.}(2023)]%
        {survey_integrated_models}
\bibfield{author}{\bibinfo{person}{Azra Seyyedi}, \bibinfo{person}{Mahdi Bohlouli}, {and} \bibinfo{person}{Seyedehsan~Nedaaee Oskoee}.} \bibinfo{year}{2023}\natexlab{}.
\newblock \showarticletitle{Machine Learning and Physics: A Survey of Integrated Models}.
\newblock \bibinfo{journal}{\emph{ACM Comput. Surv.}} \bibinfo{volume}{56}, \bibinfo{number}{5}, Article \bibinfo{articleno}{115} (\bibinfo{date}{nov} \bibinfo{year}{2023}), \bibinfo{numpages}{33}~pages.
\newblock
\showISSN{0360-0300}
\urldef\tempurl%
\url{https://doi.org/10.1145/3611383}
\showDOI{\tempurl}


\bibitem[Shensa(1992)]%
        {UDWT2}
\bibfield{author}{\bibinfo{person}{M.J. Shensa}.} \bibinfo{year}{1992}\natexlab{}.
\newblock \showarticletitle{The discrete wavelet transform: wedding the a trous and Mallat algorithms}.
\newblock \bibinfo{journal}{\emph{IEEE Transactions on Signal Processing}} \bibinfo{volume}{40}, \bibinfo{number}{10} (\bibinfo{year}{1992}), \bibinfo{pages}{2464--2482}.
\newblock
\urldef\tempurl%
\url{https://doi.org/10.1109/78.157290}
\showDOI{\tempurl}


\bibitem[Shepherd et~al\mbox{.}(2020)]%
        {icerise}
\bibfield{author}{\bibinfo{person}{Andrew Shepherd}, \bibinfo{person}{Erik Ivins}, \bibinfo{person}{Eric Rignot}, \bibinfo{person}{Ben Smith}, \bibinfo{person}{Michiel van~den Broeke}, \bibinfo{person}{Isabella Velicogna}, \bibinfo{person}{Pippa Whitehouse}, \bibinfo{person}{Kate Briggs}, \bibinfo{person}{Ian Joughin}, \bibinfo{person}{Gerhard Krinner}, {and} \bibinfo{person}{et al.}} \bibinfo{year}{2020}\natexlab{}.
\newblock \showarticletitle{Mass balance of the Greenland Ice Sheet from 1992 to 2018}.
\newblock \bibinfo{journal}{\emph{Nature}} \bibinfo{volume}{579}, \bibinfo{number}{7798} (\bibinfo{date}{01 Mar} \bibinfo{year}{2020}), \bibinfo{pages}{233--239}.
\newblock
\showISSN{1476-4687}
\urldef\tempurl%
\url{https://doi.org/10.1038/s41586-019-1855-2}
\showDOI{\tempurl}


\bibitem[Tang et~al\mbox{.}(2016)]%
        {Land_Ice_Internal_Layers}
\bibfield{author}{\bibinfo{person}{Xue-Yuan Tang}, \bibinfo{person}{Jing-Xue Guo}, \bibinfo{person}{Bo Sun}, \bibinfo{person}{Tian-Tian Wang}, {and} \bibinfo{person}{Xiang-Bin Cui}.} \bibinfo{year}{2016}\natexlab{}.
\newblock \showarticletitle{Ice thickness, internal layers, and surface and subglacial topography in the vicinity of Chinese Antarctic Taishan station in Princess Elizabeth Land, East Antarctica}.
\newblock \bibinfo{journal}{\emph{Applied Geophysics}} \bibinfo{volume}{13}, \bibinfo{number}{1} (\bibinfo{date}{01 Mar} \bibinfo{year}{2016}), \bibinfo{pages}{203--208}.
\newblock
\showISSN{1993-0658}
\urldef\tempurl%
\url{https://doi.org/10.1007/s11770-016-0540-6}
\showDOI{\tempurl}


\bibitem[Teisberg et~al\mbox{.}(2021)]%
        {teisberg}
\bibfield{author}{\bibinfo{person}{Thomas~O. Teisberg}, \bibinfo{person}{Dustin~M. Schroeder}, {and} \bibinfo{person}{Emma~J. MacKie}.} \bibinfo{year}{2021}\natexlab{}.
\newblock \showarticletitle{A Machine Learning Approach to Mass-Conserving Ice Thickness Interpolation}. In \bibinfo{booktitle}{\emph{2021 IEEE International Geoscience and Remote Sensing Symposium IGARSS}}. \bibinfo{pages}{8664--8667}.
\newblock
\urldef\tempurl%
\url{https://doi.org/10.1109/IGARSS47720.2021.9555002}
\showDOI{\tempurl}


\bibitem[Van Der~Veen and Payne(2004)]%
        {Land_Ice_Dynamics}
\bibfield{author}{\bibinfo{person}{Cornelis~J. Van Der~Veen} {and} \bibinfo{person}{Antony~J. Payne}.} \bibinfo{year}{2004}\natexlab{}.
\newblock \bibinfo{booktitle}{\emph{Modelling land-ice dynamics}}.
\newblock \bibinfo{publisher}{Cambridge University Press}, \bibinfo{pages}{169–226}.
\newblock
\urldef\tempurl%
\url{https://doi.org/10.1017/CBO9780511535659.008}
\showDOI{\tempurl}


\bibitem[Van~Katwyk et~al\mbox{.}(2023)]%
        {Katwyk2023}
\bibfield{author}{\bibinfo{person}{Peter Van~Katwyk}, \bibinfo{person}{Baylor Fox-Kemper}, \bibinfo{person}{Hélène Seroussi}, \bibinfo{person}{Sophie Nowicki}, {and} \bibinfo{person}{Karianne~J. Bergen}.} \bibinfo{year}{2023}\natexlab{}.
\newblock \showarticletitle{A Variational LSTM Emulator of Sea Level Contribution From the Antarctic Ice Sheet}.
\newblock \bibinfo{journal}{\emph{Journal of Advances in Modeling Earth Systems}} \bibinfo{volume}{15}, \bibinfo{number}{12} (\bibinfo{year}{2023}), \bibinfo{pages}{e2023MS003899}.
\newblock
\urldef\tempurl%
\url{https://doi.org/10.1029/2023MS003899}
\showDOI{\tempurl}
\showeprint{https://agupubs.onlinelibrary.wiley.com/doi/pdf/10.1029/2023MS003899}
\newblock
\shownote{e2023MS003899 2023MS003899}.


\bibitem[Varshney et~al\mbox{.}(2022)]%
        {Varshneyphysicslabels}
\bibfield{author}{\bibinfo{person}{Debvrat Varshney}, \bibinfo{person}{Oluwanisola Ibikunle}, \bibinfo{person}{John Paden}, {and} \bibinfo{person}{Maryam Rahnemoonfar}.} \bibinfo{year}{2022}\natexlab{}.
\newblock \showarticletitle{Learning Snow Layer Thickness Through Physics Defined Labels}. In \bibinfo{booktitle}{\emph{IGARSS 2022 - 2022 IEEE International Geoscience and Remote Sensing Symposium}}. \bibinfo{pages}{1233--1236}.
\newblock
\urldef\tempurl%
\url{https://doi.org/10.1109/IGARSS46834.2022.9884370}
\showDOI{\tempurl}


\bibitem[Varshney et~al\mbox{.}(2020)]%
        {varshney2020_deep_ice_layer}
\bibfield{author}{\bibinfo{person}{D. Varshney}, \bibinfo{person}{M. Rahnemoonfar}, \bibinfo{person}{M. Yari}, {and} \bibinfo{person}{J. Paden}.} \bibinfo{year}{2020}\natexlab{}.
\newblock \showarticletitle{Deep Ice Layer Tracking and Thickness Estimation using Fully Convolutional Networks}. In \bibinfo{booktitle}{\emph{2020 IEEE International Conference on Big Data (Big Data)}}. \bibinfo{publisher}{IEEE Computer Society}, \bibinfo{address}{Los Alamitos, CA, USA}, \bibinfo{pages}{3943--3952}.
\newblock
\urldef\tempurl%
\url{https://doi.org/10.1109/BigData50022.2020.9378070}
\showDOI{\tempurl}


\bibitem[Varshney et~al\mbox{.}(2021a)]%
        {varshney_2021_regression_networks}
\bibfield{author}{\bibinfo{person}{Debvrat Varshney}, \bibinfo{person}{Maryam Rahnemoonfar}, \bibinfo{person}{Masoud Yari}, {and} \bibinfo{person}{John Paden}.} \bibinfo{year}{2021}\natexlab{a}.
\newblock \showarticletitle{Regression Networks for Calculating Englacial Layer Thickness}. In \bibinfo{booktitle}{\emph{2021 IEEE International Geoscience and Remote Sensing Symposium IGARSS}}. \bibinfo{pages}{2393--2396}.
\newblock
\urldef\tempurl%
\url{https://doi.org/10.1109/IGARSS47720.2021.9553596}
\showDOI{\tempurl}


\bibitem[Varshney et~al\mbox{.}(2021b)]%
        {Varshney2021}
\bibfield{author}{\bibinfo{person}{Debvrat Varshney}, \bibinfo{person}{Maryam Rahnemoonfar}, \bibinfo{person}{Masoud Yari}, \bibinfo{person}{John Paden}, \bibinfo{person}{Oluwanisola Ibikunle}, {and} \bibinfo{person}{Jilu Li}.} \bibinfo{year}{2021}\natexlab{b}.
\newblock \showarticletitle{Deep Learning on Airborne Radar Echograms for Tracing Snow Accumulation Layers of the Greenland Ice Sheet}.
\newblock \bibinfo{journal}{\emph{Remote Sensing}} \bibinfo{volume}{13}, \bibinfo{number}{14} (\bibinfo{year}{2021}).
\newblock
\showISSN{2072-4292}
\urldef\tempurl%
\url{https://doi.org/10.3390/rs13142707}
\showDOI{\tempurl}


\bibitem[Varshney et~al\mbox{.}(2021c)]%
        {varshney2021refining}
\bibfield{author}{\bibinfo{person}{Debvrat Varshney}, \bibinfo{person}{Masoud Yari}, \bibinfo{person}{Tashnim Chowdhury}, {and} \bibinfo{person}{Maryam Rahnemoonfar}.} \bibinfo{year}{2021}\natexlab{c}.
\newblock \showarticletitle{Refining Ice Layer Tracking through Wavelet combined Neural Networks}. In \bibinfo{booktitle}{\emph{ICML 2021 Workshop on Tackling Climate Change with Machine Learning}}.
\newblock
\urldef\tempurl%
\url{https://www.climatechange.ai/papers/icml2021/49}
\showURL{%
\tempurl}


\bibitem[Varshney et~al\mbox{.}(2023)]%
        {varshney2023skipwavenet}
\bibfield{author}{\bibinfo{person}{Debvrat Varshney}, \bibinfo{person}{Masoud Yari}, \bibinfo{person}{Oluwanisola Ibikunle}, \bibinfo{person}{Jilu Li}, \bibinfo{person}{John Paden}, {and} \bibinfo{person}{Maryam Rahnemoonfar}.} \bibinfo{year}{2023}\natexlab{}.
\newblock \bibinfo{title}{Skip-WaveNet: A Wavelet based Multi-scale Architecture to Trace Firn Layers in Radar Echograms}.
\newblock
\newblock
\showeprint[arxiv]{2310.19574}~[cs.CV]


\bibitem[Vieli and Payne(2003)]%
        {control3}
\bibfield{author}{\bibinfo{person}{Andreas Vieli} {and} \bibinfo{person}{Antony~J. Payne}.} \bibinfo{year}{2003}\natexlab{}.
\newblock \showarticletitle{Application of control methods for modelling the flow of Pine Island Glacier, West Antarctica}.
\newblock \bibinfo{journal}{\emph{Annals of Glaciology}}  \bibinfo{volume}{36} (\bibinfo{year}{2003}), \bibinfo{pages}{197–204}.
\newblock
\urldef\tempurl%
\url{https://doi.org/10.3189/172756403781816338}
\showDOI{\tempurl}


\bibitem[Wang et~al\mbox{.}(2018)]%
        {wang2018_ESRGAN}
\bibfield{author}{\bibinfo{person}{Xintao Wang}, \bibinfo{person}{Ke Yu}, \bibinfo{person}{Shixiang Wu}, \bibinfo{person}{Jinjin Gu}, \bibinfo{person}{Yihao Liu}, \bibinfo{person}{Chao Dong}, \bibinfo{person}{Yu Qiao}, {and} \bibinfo{person}{Chen Change~Loy}.} \bibinfo{year}{2018}\natexlab{}.
\newblock \showarticletitle{Esrgan: Enhanced super-resolution generative adversarial networks}. In \bibinfo{booktitle}{\emph{Proceedings of the European conference on computer vision (ECCV) workshops}}. \bibinfo{pages}{0--0}.
\newblock


\bibitem[Wang et~al\mbox{.}(2022)]%
        {rheologyiceshelf}
\bibfield{author}{\bibinfo{person}{Yongji Wang}, \bibinfo{person}{Ching-Yao Lai}, {and} \bibinfo{person}{Charlie Cowen-Breen}.} \bibinfo{year}{2022}\natexlab{}.
\newblock \bibinfo{title}{Discovering the rheology of Antarctic Ice Shelves via physics-informed deep learning}.
\newblock
\newblock
\urldef\tempurl%
\url{https://doi.org/10.21203/rs.3.rs-2135795/v1}
\showDOI{\tempurl}


\bibitem[Wang et~al\mbox{.}(2021)]%
        {icelayers2021icme}
\bibfield{author}{\bibinfo{person}{Yuchen Wang}, \bibinfo{person}{Mingze Xu}, \bibinfo{person}{John Paden}, \bibinfo{person}{Lara Koenig}, \bibinfo{person}{Geoffrey~C. Fox}, {and} \bibinfo{person}{David~J. Crandall}.} \bibinfo{year}{2021}\natexlab{}.
\newblock \showarticletitle{Deep Tiered Image Segmentation for Detecting Internal Ice Layers in Radar Imagery}. In \bibinfo{booktitle}{\emph{IEEE International Conference on Multimedia and Expo (ICME)}}.
\newblock


\bibitem[Wu et~al\mbox{.}(2023)]%
        {survey_subdomain_chemical_engineering}
\bibfield{author}{\bibinfo{person}{Zhiyong Wu}, \bibinfo{person}{Huan Wang}, \bibinfo{person}{Chang He}, \bibinfo{person}{Bingjian Zhang}, \bibinfo{person}{Tao Xu}, {and} \bibinfo{person}{Qinglin Chen}.} \bibinfo{year}{2023}\natexlab{}.
\newblock \showarticletitle{The Application of Physics-Informed Machine Learning in Multiphysics Modeling in Chemical Engineering}.
\newblock \bibinfo{journal}{\emph{Industrial {\&} Engineering Chemistry Research}} \bibinfo{volume}{62}, \bibinfo{number}{44} (\bibinfo{date}{08 Nov} \bibinfo{year}{2023}), \bibinfo{pages}{18178--18204}.
\newblock
\showISSN{0888-5885}
\urldef\tempurl%
\url{https://doi.org/10.1021/acs.iecr.3c02383}
\showDOI{\tempurl}


\bibitem[Xie and Tu(2015)]%
        {HEDedge}
\bibfield{author}{\bibinfo{person}{Saining Xie} {and} \bibinfo{person}{Zhuowen Tu}.} \bibinfo{year}{2015}\natexlab{}.
\newblock \showarticletitle{Holistically-Nested Edge Detection}. In \bibinfo{booktitle}{\emph{2015 IEEE International Conference on Computer Vision (ICCV)}}. \bibinfo{pages}{1395--1403}.
\newblock
\urldef\tempurl%
\url{https://doi.org/10.1109/ICCV.2015.164}
\showDOI{\tempurl}


\bibitem[Xu et~al\mbox{.}(2023)]%
        {survey_subdomain_system}
\bibfield{author}{\bibinfo{person}{Yanwen Xu}, \bibinfo{person}{Sara Kohtz}, \bibinfo{person}{Jessica Boakye}, \bibinfo{person}{Paolo Gardoni}, {and} \bibinfo{person}{Pingfeng Wang}.} \bibinfo{year}{2023}\natexlab{}.
\newblock \showarticletitle{Physics-informed machine learning for reliability and systems safety applications: State of the art and challenges}.
\newblock \bibinfo{journal}{\emph{Reliability Engineering \& System Safety}}  \bibinfo{volume}{230} (\bibinfo{year}{2023}), \bibinfo{pages}{108900}.
\newblock
\showISSN{0951-8320}
\urldef\tempurl%
\url{https://doi.org/10.1016/j.ress.2022.108900}
\showDOI{\tempurl}


\bibitem[Yari et~al\mbox{.}(2021)]%
        {yari_airborne_snow_radar}
\bibfield{author}{\bibinfo{person}{Masoud Yari}, \bibinfo{person}{Oluwanisola Ibikunle}, \bibinfo{person}{Debvrat Varshney}, \bibinfo{person}{Tashnim Chowdhury}, \bibinfo{person}{Argho Sarkar}, \bibinfo{person}{John Paden}, \bibinfo{person}{Jilu Li}, {and} \bibinfo{person}{Maryam Rahnemoonfar}.} \bibinfo{year}{2021}\natexlab{}.
\newblock \showarticletitle{Airborne Snow Radar Data Simulation With Deep Learning and Physics-Driven Methods}.
\newblock \bibinfo{journal}{\emph{IEEE Journal of Selected Topics in Applied Earth Observations and Remote Sensing}}  \bibinfo{volume}{14} (\bibinfo{year}{2021}), \bibinfo{pages}{12035--12047}.
\newblock
\urldef\tempurl%
\url{https://doi.org/10.1109/JSTARS.2021.3126547}
\showDOI{\tempurl}


\bibitem[Yari et~al\mbox{.}(2020)]%
        {yari_2020_multi_scale}
\bibfield{author}{\bibinfo{person}{Masoud Yari}, \bibinfo{person}{Maryam Rahnemoonfar}, {and} \bibinfo{person}{John Paden}.} \bibinfo{year}{2020}\natexlab{}.
\newblock \showarticletitle{Multi-Scale and Temporal Transfer Learning for Automatic Tracking of Internal Ice Layers}. In \bibinfo{booktitle}{\emph{IGARSS 2020 - 2020 IEEE International Geoscience and Remote Sensing Symposium}}. \bibinfo{pages}{6934--6937}.
\newblock
\urldef\tempurl%
\url{https://doi.org/10.1109/IGARSS39084.2020.9323758}
\showDOI{\tempurl}


\bibitem[Yari et~al\mbox{.}(2019)]%
        {yari_2019_smart_tracking}
\bibfield{author}{\bibinfo{person}{Masoud Yari}, \bibinfo{person}{Maryam Rahnemoonfar}, \bibinfo{person}{John Paden}, \bibinfo{person}{Ibikunle Oluwanisola}, \bibinfo{person}{Lora Koenig}, {and} \bibinfo{person}{Lynn Montgomery}.} \bibinfo{year}{2019}\natexlab{}.
\newblock \showarticletitle{Smart Tracking of Internal Layers of Ice in Radar Data via Multi-Scale Learning}. In \bibinfo{booktitle}{\emph{2019 IEEE International Conference on Big Data (Big Data)}}. \bibinfo{pages}{5462--5468}.
\newblock
\urldef\tempurl%
\url{https://doi.org/10.1109/BigData47090.2019.9006083}
\showDOI{\tempurl}


\bibitem[Zalatan and Rahnemoonfar(2023a)]%
        {Zalatan_igarss}
\bibfield{author}{\bibinfo{person}{Benjamin Zalatan} {and} \bibinfo{person}{Maryam Rahnemoonfar}.} \bibinfo{year}{2023}\natexlab{a}.
\newblock \showarticletitle{Prediction of Annual Snow Accumulation Using a Recurrent Graph Convolutional Approach}. In \bibinfo{booktitle}{\emph{IGARSS 2023 - 2023 IEEE International Geoscience and Remote Sensing Symposium}}. \bibinfo{pages}{5344--5347}.
\newblock
\urldef\tempurl%
\url{https://doi.org/10.1109/IGARSS52108.2023.10283236}
\showDOI{\tempurl}


\bibitem[Zalatan and Rahnemoonfar(2023b)]%
        {zalatan_icip}
\bibfield{author}{\bibinfo{person}{Benjamin Zalatan} {and} \bibinfo{person}{Maryam Rahnemoonfar}.} \bibinfo{year}{2023}\natexlab{b}.
\newblock \showarticletitle{Prediction of Deep Ice Layer Thickness Using Adaptive Recurrent Graph Neural Networks}. In \bibinfo{booktitle}{\emph{2023 IEEE International Conference on Image Processing (ICIP)}}. \bibinfo{pages}{2835--2839}.
\newblock
\urldef\tempurl%
\url{https://doi.org/10.1109/ICIP49359.2023.10222391}
\showDOI{\tempurl}


\bibitem[Zalatan and Rahnemoonfar(2023c)]%
        {Zalatan2023}
\bibfield{author}{\bibinfo{person}{Benjamin Zalatan} {and} \bibinfo{person}{Maryam Rahnemoonfar}.} \bibinfo{year}{2023}\natexlab{c}.
\newblock \showarticletitle{Recurrent Graph Convolutional Networks for Spatiotemporal Prediction of Snow Accumulation Using Airborne Radar}. In \bibinfo{booktitle}{\emph{2023 IEEE Radar Conference (RadarConf23)}}. \bibinfo{pages}{1--6}.
\newblock
\urldef\tempurl%
\url{https://doi.org/10.1109/RadarConf2351548.2023.10149562}
\showDOI{\tempurl}


\bibitem[Zhang et~al\mbox{.}(2019)]%
        {Zhang2019}
\bibfield{author}{\bibinfo{person}{E. Zhang}, \bibinfo{person}{L. Liu}, {and} \bibinfo{person}{L. Huang}.} \bibinfo{year}{2019}\natexlab{}.
\newblock \showarticletitle{Automatically delineating the calving front of Jakobshavn Isbr{\ae} from multitemporal TerraSAR-X images: a deep learning approach}.
\newblock \bibinfo{journal}{\emph{The Cryosphere}} \bibinfo{volume}{13}, \bibinfo{number}{6} (\bibinfo{year}{2019}), \bibinfo{pages}{1729--1741}.
\newblock
\urldef\tempurl%
\url{https://doi.org/10.5194/tc-13-1729-2019}
\showDOI{\tempurl}


\bibitem[Zhang et~al\mbox{.}(2021)]%
        {Zhang2021}
\bibfield{author}{\bibinfo{person}{Enze Zhang}, \bibinfo{person}{Lin Liu}, \bibinfo{person}{Lingcao Huang}, {and} \bibinfo{person}{Ka~Shing Ng}.} \bibinfo{year}{2021}\natexlab{}.
\newblock \showarticletitle{An automated, generalized, deep-learning-based method for delineating the calving fronts of Greenland glaciers from multi-sensor remote sensing imagery}.
\newblock \bibinfo{journal}{\emph{Remote Sensing of Environment}}  \bibinfo{volume}{254} (\bibinfo{year}{2021}), \bibinfo{pages}{112265}.
\newblock
\showISSN{0034-4257}
\urldef\tempurl%
\url{https://doi.org/10.1016/j.rse.2020.112265}
\showDOI{\tempurl}


\bibitem[Zhang et~al\mbox{.}(2000)]%
        {Sea_Ice_Dynamics}
\bibfield{author}{\bibinfo{person}{Jinlun Zhang}, \bibinfo{person}{Drew Rothrock}, {and} \bibinfo{person}{Michael Steele}.} \bibinfo{year}{2000}\natexlab{}.
\newblock \showarticletitle{Recent Changes in Arctic Sea Ice: The Interplay between Ice Dynamics and Thermodynamics}.
\newblock \bibinfo{journal}{\emph{Journal of Climate}} \bibinfo{volume}{13}, \bibinfo{number}{17} (\bibinfo{year}{2000}), \bibinfo{pages}{3099 -- 3114}.
\newblock
\urldef\tempurl%
\url{https://doi.org/10.1175/1520-0442(2000)013<3099:RCIASI>2.0.CO;2}
\showDOI{\tempurl}


\bibitem[Zhang et~al\mbox{.}(2023)]%
        {Zhang2023}
\bibfield{author}{\bibinfo{person}{Yali Zhang}, \bibinfo{person}{Lifeng Zhang}, \bibinfo{person}{Yi He}, \bibinfo{person}{Sheng Yao}, \bibinfo{person}{Wang Yang}, \bibinfo{person}{Shengpeng Cao}, {and} \bibinfo{person}{Qiang Sun}.} \bibinfo{year}{2023}\natexlab{}.
\newblock \showarticletitle{Analysis of the future trends of typical mountain glacier movements along the Sichuan-Tibet Railway based on ConvGRU network}.
\newblock \bibinfo{journal}{\emph{International Journal of Digital Earth}} \bibinfo{volume}{16}, \bibinfo{number}{1} (\bibinfo{year}{2023}), \bibinfo{pages}{762--780}.
\newblock
\urldef\tempurl%
\url{https://doi.org/10.1080/17538947.2022.2152884}
\showDOI{\tempurl}
\showeprint{https://doi.org/10.1080/17538947.2022.2152884}


\bibitem[Zhao et~al\mbox{.}(2017)]%
        {zhao2017PSP}
\bibfield{author}{\bibinfo{person}{Hengshuang Zhao}, \bibinfo{person}{Jianping Shi}, \bibinfo{person}{Xiaojuan Qi}, \bibinfo{person}{Xiaogang Wang}, {and} \bibinfo{person}{Jiaya Jia}.} \bibinfo{year}{2017}\natexlab{}.
\newblock \showarticletitle{Pyramid scene parsing network}. In \bibinfo{booktitle}{\emph{Proceedings of the IEEE conference on computer vision and pattern recognition}}. \bibinfo{pages}{2881--2890}.
\newblock


\bibitem[Zheng et~al\mbox{.}(2022)]%
        {Zheng2022}
\bibfield{author}{\bibinfo{person}{Qingyu Zheng}, \bibinfo{person}{Wei Li}, \bibinfo{person}{Qi Shao}, \bibinfo{person}{Guijun Han}, {and} \bibinfo{person}{Xuan Wang}.} \bibinfo{year}{2022}\natexlab{}.
\newblock \showarticletitle{A Mid- and Long-Term Arctic Sea Ice Concentration Prediction Model Based on Deep Learning Technology}.
\newblock \bibinfo{journal}{\emph{Remote Sensing}} \bibinfo{volume}{14}, \bibinfo{number}{12} (\bibinfo{year}{2022}).
\newblock
\showISSN{2072-4292}
\urldef\tempurl%
\url{https://doi.org/10.3390/rs14122889}
\showDOI{\tempurl}


\bibitem[Zhu et~al\mbox{.}(2017)]%
        {Zhu2017_cycleGAN}
\bibfield{author}{\bibinfo{person}{Jun-Yan Zhu}, \bibinfo{person}{Taesung Park}, \bibinfo{person}{Phillip Isola}, {and} \bibinfo{person}{Alexei~A. Efros}.} \bibinfo{year}{2017}\natexlab{}.
\newblock \showarticletitle{Unpaired Image-to-Image Translation Using Cycle-Consistent Adversarial Networks}. In \bibinfo{booktitle}{\emph{2017 IEEE International Conference on Computer Vision (ICCV)}}. \bibinfo{pages}{2242--2251}.
\newblock
\urldef\tempurl%
\url{https://doi.org/10.1109/ICCV.2017.244}
\showDOI{\tempurl}


\bibitem[Zideh et~al\mbox{.}(2024)]%
        {survey_subdomain_data_anomaly}
\bibfield{author}{\bibinfo{person}{Mehdi~Jabbari Zideh}, \bibinfo{person}{Paroma Chatterjee}, {and} \bibinfo{person}{Anurag~K. Srivastava}.} \bibinfo{year}{2024}\natexlab{}.
\newblock \showarticletitle{Physics-Informed Machine Learning for Data Anomaly Detection, Classification, Localization, and Mitigation: A Review, Challenges, and Path Forward}.
\newblock \bibinfo{journal}{\emph{IEEE Access}}  \bibinfo{volume}{12} (\bibinfo{year}{2024}), \bibinfo{pages}{4597--4617}.
\newblock
\urldef\tempurl%
\url{https://doi.org/10.1109/ACCESS.2023.3347989}
\showDOI{\tempurl}


\bibitem[Zwally et~al\mbox{.}(2011)]%
        {zwally2011}
\bibfield{author}{\bibinfo{person}{H.~Jay Zwally}, \bibinfo{person}{Jun Li}, \bibinfo{person}{Anita~C. Brenner}, \bibinfo{person}{Matthew Beckley}, \bibinfo{person}{Helen~G. Cornejo}, \bibinfo{person}{John DiMarzio}, \bibinfo{person}{Mario~B. Giovinetto}, \bibinfo{person}{Thomas~A. Neumann}, \bibinfo{person}{John Robbins}, \bibinfo{person}{Jack~L. Saba}, {and} \bibinfo{person}{et al.}} \bibinfo{year}{2011}\natexlab{}.
\newblock \showarticletitle{Greenland ice sheet mass balance: distribution of increased mass loss with climate warming; 2003–07 versus 1992–2002}.
\newblock \bibinfo{journal}{\emph{Journal of Glaciology}} \bibinfo{volume}{57}, \bibinfo{number}{201} (\bibinfo{year}{2011}), \bibinfo{pages}{88–102}.
\newblock
\urldef\tempurl%
\url{https://doi.org/10.3189/002214311795306682}
\showDOI{\tempurl}


\end{thebibliography}

\end{document}